\documentclass{article}

\usepackage{PRIMEarxiv}

\usepackage[utf8]{inputenc} 
\usepackage[T1]{fontenc}    
\usepackage{hyperref}       
\usepackage{url}            
\usepackage{booktabs}       
\usepackage{amsfonts}       
\usepackage{nicefrac}       
\usepackage{microtype}      
\usepackage{lipsum}
\usepackage{fancyhdr}       
\usepackage{graphicx}       
\graphicspath{{media/}}     
\usepackage{gensymb}
\usepackage{enumerate}
\newcommand{\todo}[1]{{\color{black}#1}}
\newcommand{\tlts}[1]{{\color{black}#1}}

\usepackage{multirow}
\usepackage{amsthm}
\usepackage{subfig}
\usepackage{multirow}
\usepackage{color,soul}
\usepackage{graphicx}
\usepackage{subfig}
\usepackage{xcolor}
\usepackage{tabulary}
\usepackage{tabularx}
\usepackage{array}
\usepackage{enumitem}
\usepackage{comment}

\newcommand{\pglp}[1]{{\color{red}#1}}
\newcommand{\crj}[1]{{\color{cyan}#1}}

\newcommand{\etal}{\textit{et~al.~}}
\newcommand{\ie}{\textit{i.e.,~}}
\newcommand{\wrt}{\textit{w.r.t.~}}
\newcommand{\eg}{\textit{e.g.~}}
\newcolumntype{x}[1]{>{\centering\arraybackslash\hspace{0pt}}p{#1}}

\title{3D Scene Geometry Estimation from 360$^\circ$ Imagery: \\A Survey}

\author{
Thiago L. T. da Silveira, Paulo G. L. Pinto, Jeffri E. M. Llerena, Cláudio R. Jung \\
Institute of Informatics, Federal University of Rio Grande do Sul, Brazil \\
  \texttt{\{tltsilveira, paulo.pinto, jeffri.mllerena, crjung\}@inf.ufrgs.br}
}

\begin{document}
\maketitle

\begin{abstract}
This paper provides a comprehensive survey on pioneer and state-of-the-art 3D scene geometry estimation methodologies based on single, two, or multiple images captured under the
omnidirectional  optics.
%
We first revisit the basic concepts of the spherical camera model, and review the most common acquisition technologies and representation formats suitable for \tlts{omnidirectional (also called 360$^\circ$, spherical or panoramic)} images and videos.
We then survey monocular layout and depth inference approaches, highlighting the recent advances in learning-based solutions suited for spherical data.
The classical stereo matching is then revised 
\tlts{on the spherical domain}, where methodologies for detecting and describing sparse and dense features become crucial.
The stereo matching concepts are then
extrapolated for multiple view camera setups, categorizing them among light fields, multi-view stereo, and structure from motion (or visual simultaneous localization and mapping).
We also compile and discuss 
commonly adopted datasets and  figures of merit indicated for each purpose 
 \tlts{and list recent results for completeness}.
We conclude 
this paper by 
pointing out current and future trends.
 \end{abstract}


\section{Introduction}

The world is three-dimensional (3D). As such, recovering 3D information about real-world objects allows the exploration of many relevant applications, including 
self-driving cars~\cite{Wang:ACCV:2018, Schonbein:IROS:2014}, 
robot navigation~\cite{Moreau:SITIS:2012,Fernandez-Labrador:LRA:2020},
virtual tourism~\cite{Fangi:ISPRS:2018,Kim:JCVIU:2015}, 
infrastructure inspection~\cite{Pathak:IST:2016,Pathak:JCMSI:2017}, archaeological~\cite{Pagani:VAST:2011,Gava:VISAPP:2015} and architectural modeling~\cite{Fangi:ISPRS:2018,Giovanni:ECCV:2016}, 
city planning~\cite{Kim:IVMSPW:2013,Kuhn:GCPR:2017}, and 3D cinema~\cite{Azevedo:TCSVT:2019,Thatte:VCIP:2017}. 

Many sensors can be used to obtain 3D data from real objects, such as light detection and ranging 
\cite{Kim:IJCV:2013}, structured light~\cite{Silberman:ICCV:2011}, and time of flight 
\cite{Lukierski:ECMR:2015}. 
There is a plethora of approaches for inferring 3D information from plain color images/videos. 
The widespread accessibility and low-cost of consumer cameras is a strong motivation for the continued research efforts
devoted to image-based 3D scene reconstruction methods~\cite{Kim:ICCV:2009}. 
In theory, 3D information can only be inferred from two or more
captures of the scene, as in typical multi-view stereo 
\cite{Seitz:CVPR:2006} or structure from motion
 \cite{Ozyesil:AN:2017} approaches. 
However, recent approaches are exploring machine learning
to perform \emph{single-image}  depth inference~\cite{Hamzah:Sensors:2016,Kumari:IJEM:2016,Scharstein:SMBV:2001}.
Most methods developed so far rely on traditional perspective/pinhole-based cameras, which have a narrow field of view (FoV) and 
thus might 
require thousands of captures to model large scenes~\cite{Agarwal:CACM:2011, Pollefeys:IJCV:2008}.  

%

Unlike pinhole cameras, \emph{truly omnidirectional} cameras capture the light intensities from the entire surrounding 
environment ($360^\circ\times180^\circ$) at once~ \cite{Im:ECCV:2016, Huang:VR:2017}. 
%
 Thus,
from the application point of view, only two captures might be necessary for recovering 3D information from the whole scenario (except for occlusions and disocclusions) based on triangulation-like algorithms~\cite{Oliveira:ICASSP:2019} or even a single capture might be needed when using modern deep learning approaches~\cite{Tateno:ECCV:2018}.
%
Although catadioptric sensors date from the early 1970s~\cite{Rees:Patent:1970}, the recent release of affordable devices for capturing and visualizing 
\emph{full FoV} media besides the novel augmented, mixed and virtual (AR/MR/VR) applications are inciting both the scientific community and the industry to invest efforts in novel technological solutions.
Popular consumer-grade cameras include the 
Samsung Gear 360\footnote{\url{https://www.samsung.com/global/galaxy/gear-360/}}
and the 
Ricoh Theta\footnote{\url{https://theta360.com/en/}},
whereas examples of head-mounted displays 
are the 
Samsung Gear VR\footnote{\url{https://www.samsung.com/global/galaxy/gear-vr/}}
and the
Oculus Rift\footnote{\url{https://www.oculus.com/rift/}}.
%
It is worth mentioning that the existing nomenclature for 360 images is vast, 
and the terms 
omnidirectional \cite{Won:ICRA:2019, Wegner:ICIP:2018},
spherical \cite{Silveira:CVPR:2019, Eder:3DV:2019},
full panorama \cite{Pagani:ICCV:2011, Im:ECCV:2016}, and 
360\degree \cite{Zioulis:3DV:2019, Huang:VR:2017} images/videos/cameras are often used as synonyms.

360\degree images 
are
intrinsically defined on the  sphere surface and 
present three-degrees of freedom (3-DoF) natively, 
associated with the rotation angles pitch, yaw, and roll.
%
This allows, for instance, 
\tlts{head-mounted displays}
users to look at any direction of the captured view by performing rotational head movements.  
However, restricting the movements only to rotations may break the immersion sense when the observer's head translates and the rendered scene mismatches, causing discomfort and sickness~\cite{Thatte:VCIP:2017}.
Although most 
\tlts{head-mounted displays}
support 6-DoF tracking~\cite{Huang:VR:2017}, the vast majority of real capture content is monoscopic~\cite{Im:ECCV:2016}, so that full immersion is mostly experienced on computer-generated imagery~\cite{Luo:TVCG:2018}.
Depth perception is accomplished by presenting a stereoscopic image pair restricted to the user eye's viewport, which may reduce the discomfort~\cite{Serrano:TVCG:2019}. 
For allowing \emph{small} head translational movements, both color (often called texture) and depth (or disparity) information must be provided for that specific virtual camera (and user) location.
This still limited application scenario is called 3-DoF+~\cite{Jeong:Sensors:2018,Silveira:VR:2019}, but approximates the full 6-DoF-enabled navigation, where colored 3D information is required for every possible location within the scene~\cite{Wien:JETCAS:2019}.

Estimating 3D scene geometry from 360\degree imagery allows to address all the classical problems mentioned in the first paragraph of this paper and is also pivotal for providing full immersive experiences to 
\tlts{head-mounted displays}
users in novel AR/MR/VR-based applications.
Despite the benefits of presenting full FoV, both the capture sensors (CCD, CMOS) and the representation formats (equirectangular images or cube-maps) for spherical images involve a regular rectangular 2D grid. As such, omnidirectional 
media 
present heavy 
\tlts{non-affine}
distortions that are inherent to the spherical camera model~\cite{Eder:CVPR:2020}.
Therefore, most of the image processing and computer vision algorithms developed in the past -- which were designed to work with perspective images -- tend to fail when directly applied to omnidirectional media, regardless of their objective~\cite{Cruz-Mota:IJCV:2012,Su:NIPS:2017,Tateno:ECCV:2018,Azevedo:TCSVT:2019}.  
In this context, there is a growing number of works trying to bridge this gap by employing geometry- and learning-based approaches for recovering 3D information from one, two, or more full panoramas.
\tlts{How these applications are explored in immersive media is discussed throughout the text and summarized in the conclusions (Section~\ref{sec:conclusions}).}

This survey paper aims to briefly review the spherical imaging model (along with common acquisition pipelines and representation formats), and then focus on a comprehensive analysis of  representative methods that tackle the 3D geometry estimation problem using one or more 360\degree images.
A secondary goal is to compile meaningful publicly available datasets up to date, as well as the most widely accepted evaluation metrics for comparing the existing techniques.

\subsection{Comparison with Existing Surveys}


The survey by Gledhill and colleagues~\cite{Gledhill:CAG:2003} revises
pinhole-based approaches applied to image sequences taken by rotating cameras for composing \emph{incomplete} panoramic images via stitching. 
The authors only acknowledge the existence of pairs of opposite-located fisheye lenses and complex catadioptric systems.
%
The revised 3D reconstruction techniques 
are limited to approaches for triangulation of features matched in two or more perspective images, \ie traditional (multi-view) stereo matching, and the use of panoramas is mentioned as a trending topic. 
%

Pay\'a \etal~\cite{Paya:Sensors:2017} survey techniques for visual localization and mapping problems in the particular context of mobile robotics.
They refer to ``omnidirectional'' imagery, but the reviewed works rely on catadioptric systems that do not cover the full vertical FoV and often represent the captured information by cylindrical projection. 
Furthermore, most of the revisited methods
rely on a preprocessing step for ``undistort'' the images captured by the catadioptric systems since they cannot handle deformations.

Pintore and colleagues~\cite{pintore:EG:2020} 
review recent approaches for automatic 3D reconstruction of structured indoor scenarios  focusing on the diversity of input modalities (color, depth or multimodal).
%
They mention the advantages of working with full-FoV panoramas and the benefits of achieving registered depth information, highlighting the growing number of data-driven methods that tackle color images only and pointing out a very special niche of learning-based approaches that can infer the 3D scene layout from indoor panoramas.
Similarly, the survey paper from Zou and colleagues~\cite{zou:ijcv:2021} also concentrates on describing 3D layout recovery approaches. 
Related works addressed in~\cite{pintore:EG:2020, zou:ijcv:2021} are also revisited in our survey paper. 

Kang~\etal~\cite{Kang:ISPRS:2020} review recent approaches that use either pinhole-based cameras or active sensing for 3D reconstruction of indoor environments. 
The authors compile the most commonly used datasets for assessing methods dealing with one, two, or multiple views.
The authors divide their analysis into approaches based on geometric and topological aspects or semantics for estimating the 3D scene geometry.
However, only a few methods for depth or layout estimation based on panoramic images are revisited in their work, disregarding the particularities of the spherical camera model. In fact, the authors do not explicitly recognize as a trend the use of the emerging 360\degree image-based approaches for reconstructing the 3D geometry of scenes.

To the best of our knowledge, the current survey paper is the first to compile a comprehensive review of state-of-the-art 
approaches for 3D scene geometry recovery relying on one, two or multiple panoramas, focusing on either indoor or outdoor scenarios. This paper does not intend to review methods suited for perspective imagery. 
For that aim, the reader is redirected to relevant survey papers such as~\cite{Khan:Sensors:2020,Bhoi:arXiv:2019} for monocular depth inference,~\cite{Hamzah:Sensors:2016,Kumari:IJEM:2016,Scharstein:SMBV:2001} for stereo matching,~\cite{Johannsen:CVPR:2017} for  light fields-based depth estimation,~\cite{Seitz:CVPR:2006, furukawa:book:2015} for multi-view stereo, and~\cite{Ozyesil:AN:2017,Fuentes-Pacheco:AIR:2012,Saputra:CSUR:2018} for structure from motion or visual simultaneous localization and mapping.


\begin{figure}
 \centering
  \vspace*{.25cm}
 \includegraphics[width=.9\linewidth]{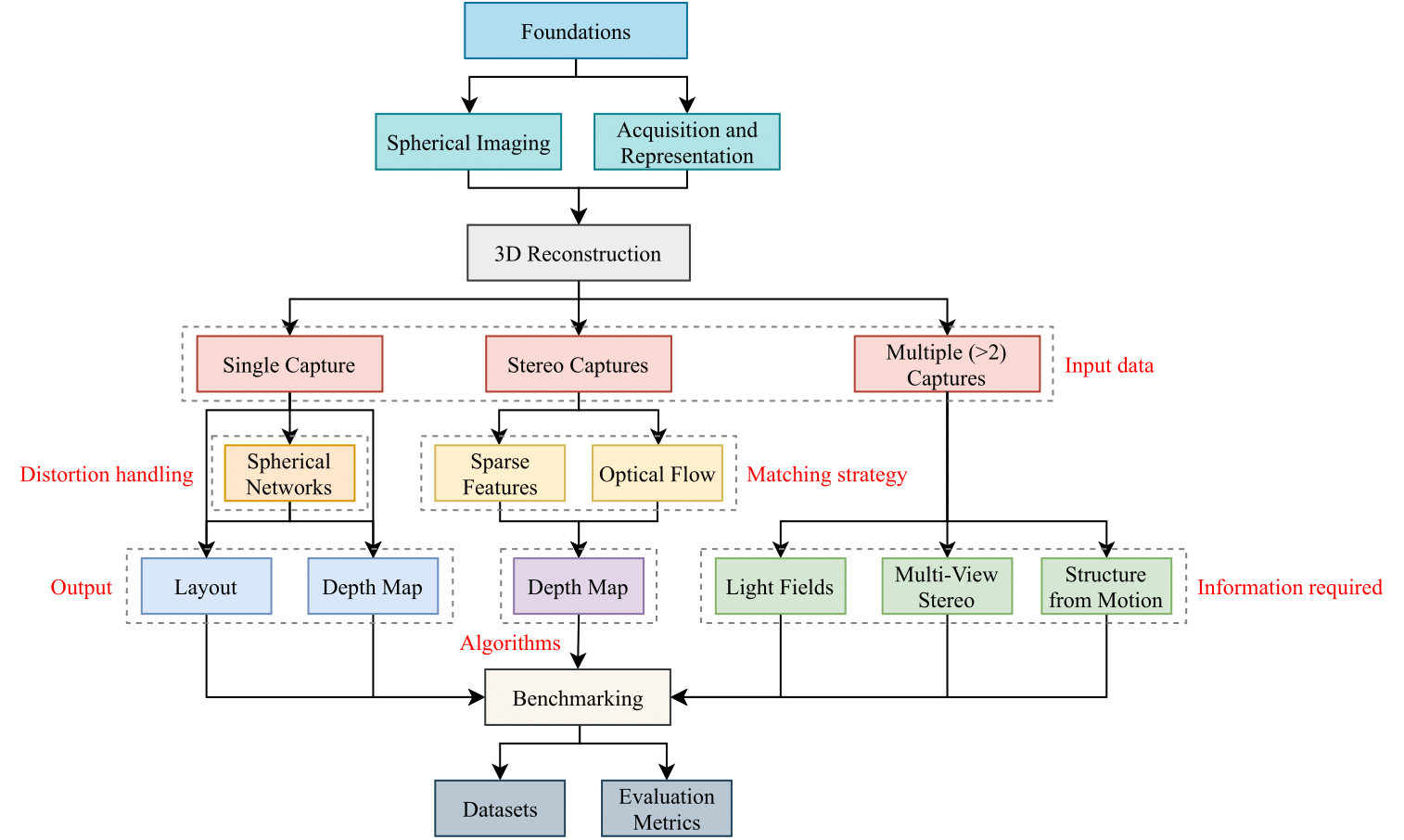}
 \caption{Paper Organization Diagram}
 \label{fig:organization}
 \vspace{-0.5cm}
\end{figure}
\subsection{Organization of this Paper}

The rest of this survey paper is organized as follows. 
Section~\ref{sec:review} \tlts{firstly} revisits the basic concepts of spherical imaging, the common acquisition pipelines, and the image representation formats often employed in the literature.
\tlts{Sections~\ref{sec:monocular},~\ref{sec:stereo}, and~\ref{sec:multiview} 
discuss techniques for recovering 3D geometry from scenarios and are organized according to the complexity of the traditional image capturing setup. More precisely,}
%
approaches for 3D geometry prediction from a single 360\degree image are examined throughout Section~\ref{sec:monocular}. 
More precisely, Section~\ref{subsec:layout} discusses the methods targeting 3D layout inference, whereas Section~\ref{subsec:monodepth}  reviews techniques aiming for pixel-wise depth estimates. Since most methods that deal with a single panorama rely on neural networks, we also present in Section~\ref{subsec:spherenn} a discussion on how to adapt planar networks to the spherical domain.
Methods that rely on pairs of spherical images for 3D scene reconstruction are reviewed along  Section~\ref{sec:stereo}. Since most of these approach require sparse or dense correspondences, we also revisit existing approaches for sparse keypoint detection and macthing and optical flow estimation, both considering the spherical domain, in Sections~\ref{subsec:sparsefeatures}  and~\ref{subsec:densefeatures}, respectively.
Section~\ref{sec:multiview} focuses on techniques designed for recovering 3D information from multiple omnidirectional captures of the environment. 
Multi-view-based approaches are grouped according to the primary camera setups: light fields, multi-view stereo, and structure from motion, and are separately addressed in Sections~\ref{subsec:lightfields},~\ref{subsec:mvs}~and~\ref{subsec:sfm}, respectively.
Section~\ref{sec:benchmark} compiles a list of publicly available datasets, 
commonly \tlts{used} assessment metrics and protocols for different scenarios and 3D representations
\tlts{as well as selected state-of-the-art results.} 
Finally, 
Section~\ref{sec:conclusions} concludes this survey paper, pointing out some of the recent trends towards 360\degree media depth-awareness and its integration to fully immersive VR/AR/MR experiences. Fig.~\ref{fig:organization} provides a schematic illustration about the content organization adopted in this survey article.

\section{Foundations, Acquisition and Representation of Spherical Images} \label{sec:review}

This section revisits the basic concepts of the spherical camera model, the commonly involved acquisition tools and devices, and the most meaningful representation formats for spherical media. 
%
%
All the information gathered throughout this section paves the way for further discussing the single-view-, stereo-, and multi-view-based approaches that estimate
3D geometry from panoramas.

\subsection{Spherical imaging} \label{ssec:sphimage}

Capturing the light intensities from the whole (full-FoV) surrounding 3D environment (world) is accomplished by spherical projection~\cite{Li:VR:2005}, which leads us to the concept of spherical imaging~\cite{Li:TITS:2008,Fujiki:MIRAGE:2007}. 
A spherical camera is modeled as an \emph{unit sphere}, having no intrinsic parameters, and, thus, being fully represented by its extrinsics~\cite{Guan:TIP:2017, Krolla:BMVC:2014}.

\begin{figure}
 \centering
 \vspace*{.125cm}
 \includegraphics[scale=0.75]{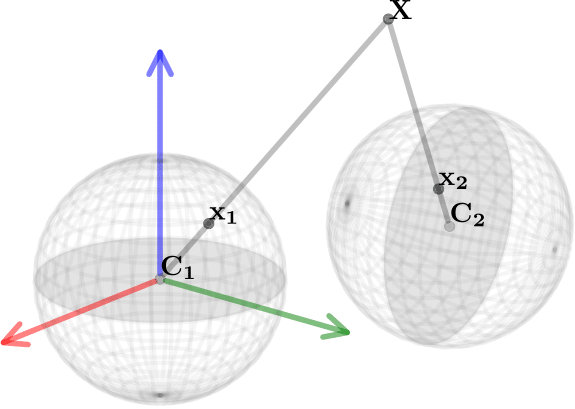}
 \caption{
    Projection of a world point onto two spherical cameras with different extrinsic parameters. 
 }
 \label{fig:epipolar}
 \vspace{-0.25cm}
\end{figure}

For the mathematical characterization of the spherical imaging, let us consider a world point $\mathbf{X} \in \mathbb{R}^3$ and a spherical camera centered at $\mathbf{C}\in \mathbb{R}^3$, both positions w.r.t. a preset world coordinate system. 
%
The camera extrinsics $[\mathbf{R}|\mathbf{t}]$ are explained as $\mathbf{R} \in SO(3)$ being a rotation matrix and $\mathbf{t} = -\mathbf{R}\mathbf{C} \in \mathbb{R}^3$ being a translation vector relative to 
\tlts{that world coordinate system}.
%
The projection of a world point $\mathbf{X}$ onto a spherical camera parameterized by $[\mathbf{R}|\mathbf{t}]$ results on a point $\mathbf{x} \in S^2 \subset \mathbb{R}^3$~\cite{Akihiko:WOVCNNC:2005}:
\begin{equation}
\mathbf{x} = \frac{\mathbf{R}\mathbf{X}+\mathbf{t}}{\|\mathbf{R}\mathbf{X}+\mathbf{t}\|_2}.
\label{eq:spherical}
\end{equation}
Fig.~\ref{fig:epipolar} 
illustrates a world point $\mathbf{X}$ 
projected onto two spherical cameras. 
One of them is centered at the origin of and aligned to the 
\tlts{world coordinate system}, \ie its extrinsic parameters are [$\mathbf{R}_1=\mathbf{I}|\mathbf{t}_1=-\mathbf{R}_1\mathbf{C}_1=\mathbf{0}$].
The other is not at the origin and might be rotated,
presenting extrinsic parameters given by
[$\mathbf{R}_2|\mathbf{t}_2=-\mathbf{R}_2\mathbf{C}_2\neq\mathbf{0}$]. %
The projections $\mathbf{x}_1$ and $\mathbf{x}_2$ represent local image coordinates (\wrt each camera), and do not contain explicit information about the camera poses \wrt 
\tlts{the preset world coordinate system}.

\tlts{From a geometric point of view,}
recovering the relative motion between two cameras
can \tlts{be} explore\tlts{d by} the coplanarity property\tlts{,}
\tlts{which associates }correspondent image points and cameras centers~\cite{Akihiko:WOVCNNC:2005}.
The epipolar geometry describes the \emph{geometrical} relation between a pair of central projection cameras capturing a static scene~\cite{Hartley:Book:2003}.
Since an omnidirectional camera is a central projection camera, so as pinhole-based ones are, this definition also applies to the spherical camera model~\cite{Gava:VISAPP:2015}. 
\tlts{Knowing the relative camera pose for two captures allows recovering the actual 3D points by geometrical reasoning.}

Let us consider 
$\mathbf{x_1}$ and $\mathbf{x_2}$
as projections of $\mathbf{X}$ onto two spherical cameras, as depicted in Fig.~\ref{fig:epipolar}. 
We can consider that the first camera is \emph{canonical}, \ie it is located at the origin of the 
\tlts{world coordinate system}
and aligned to it.
Since spherical cameras have no intrinsic parameters -- \ie the Essential $\mathbf{E}$ and the Fundamental matrices $\mathbf{F}$ are the same --
the projections $\mathbf{x_1}$ and $\mathbf{x_2}$ are related according to
\cite{Li:VR:2005,Pagani:VAST:2011,Akihiko:WOVCNNC:2005,Pathak:ICCAS:2016}
\begin{equation}
\label{eq:epipolarconstr}
\mathbf{x}_2^\top\mathbf{E}\mathbf{x}_1 = 0, \quad \mathbf{E}=[\mathbf{t}]_{\times}\mathbf{R} ,
\end{equation}
where $[\mathbf{t}]_\times$ is the skew-symmetric matrix corresponding to the cross-product with 
$\mathbf{t}$~\cite{Guan:TIP:2017, Silveira:CVPR:2019}. 
Since $\mathbf{E}$ is defined up to an unknown scale~\cite{Yang:ECCV:2014}, only five from the six 
DoFs
involving both the cameras can be resolved by exploring the epipolar constraint.

There are many strategies for estimating
$\mathbf{E}$ from pairs of correspondences~\cite{Hartley:TPAMI:1997, Nister:TPAMI:2004, Yang:ECCV:2014}.
Several approaches~\cite{Pagani:VAST:2011,Pathak:IST:2016,Silveira:VR:2019} use the traditional eight-point algorithm (8-PA)~\cite{Longuet-Higgins:RCV:1987,Hartley:TPAMI:1997} to infer the Essential matrix, and some of them refine the estimates using non-linear optimization~\cite{Guan:TIP:2017, Huang:VR:2017, Pagani:VAST:2011}.
Obtaining the relative poses $[\mathbf{R}|\mathbf{t}]$ and the underlying 3D world points $\mathbf{X}^k$ from a set of $k$ \emph{noiseless} correspondence points $\mathbf{x}_1^k$ and $\mathbf{x}_2^k$ and the recovered Essential matrix $\mathbf{E}$ is straightforward~\cite{Abdel-Aziz:PERS:1971,Hartley:Book:2003}. 
It~is~worth mentioning that dense correspondences on the sphere are hardly ambiguous in terms of epipolar geometry~\cite{brodsky:ijcv:1998} and that even the 8-PA might produce accurate estimates for the Essential matrix if the
correspondence points 
\tlts{well distributed}
on the sphere~\cite{Silveira:CVPR:2019}.

\tlts{As the reader will see in Section~\ref{sec:monocular}, many works try 
to estimate the distance connecting the camera center to the actual 3D point (\ie its depth) from single color panorama, mostly relying on paired color plus depth datasets to train a deep-learning model.
Unlike approaches based on geometric aspects relating two or more cameras, learning-based methods might not rely on keypoint correspondences and triangulation.
Different approaches for single-view 3D scene geometry recovery explore particular aspects of the spherical projection, and are further discussed in Section~\ref{sec:monocular}. }


\subsection{Acquisition and Representation}

Although the spherical camera model provides a mathematical formulation for omnidirectional vision, 
\tlts{there is no such \emph{single-sensor device}}
for capturing all the scene information at once. 
The signals are still captured on a planar silicon surface that might result in a heavily distorted output image~\cite{Adarve:LRA:2017}.

\begin{figure}
    \centering
    \subfloat[Catadioptric camera]{\includegraphics[scale=1]{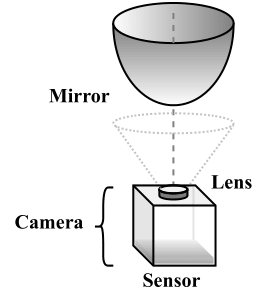}}\qquad
    \subfloat[Polydioptric camera rig]{\includegraphics[scale=1]{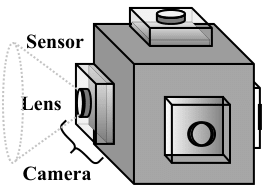}}\qquad
    \subfloat[Dual fisheye camera]{\includegraphics[scale=1]{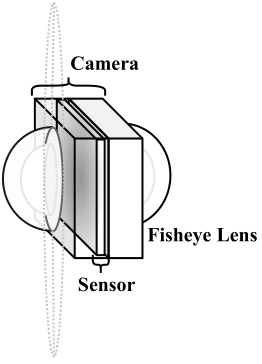}}
    \caption{\tlts{Popular omnidirectional capture setups. Dotted cylindroids represent the FoV of each lens.}}
    \label{fig:cameras}
\end{figure}

Catadioptric cameras\tlts{, illustrated in Fig.~\ref{fig:cameras}(a),} 
associate a perspective camera with convex-shaped  (conic, spherical, parabolic, or hyperbolic) mirrors~\cite{Kim:ICCV:2009,Im:ECCV:2016}, \tlts{and}
introduced the idea of capturing the full horizontal FoV~\cite{Nayar:CVPR:1997}.
However, due to its sensor/mirror self-occlusions, these cameras are incapable of capturing the full vertical FoV~\cite{Huang:Wiley:2008}, \tlts{and the images are commonly represented through cylindrical projections~\cite{Kim:IJCV:2013,Cruz-Mota:IJCV:2012}. A} 
compromise solution 
\tlts{for increasing the represented FoV}
relies on catadioptric camera rigs~\cite{Labutov:ICRA:2011}.
%
It is possible to warp the scene captures and represent the (incomplete) image on the sphere if the intrinsic parameters of these cameras are known~\cite{Akihiko:WOVCNNC:2005}.
Although most classical approaches rely on bulky and costly equipment, recent works like~\cite{Aggarwal:CVPR:2016} focus on portable mobile-aided solutions.
\tlts{Catadioptric cameras cannot capture the full-FoV, have fragile mirror components and are thus rare in recent scientific articles or industry applications.}

A polydioptric imaging setup 
is a collection of \tlts{an arbitrary number of regular} cameras organized in a rig that might capture the whole surrounding scenario,  
\tlts{as illustrated in Fig. \ref{fig:cameras}(b).}
These camera rigs are often associated with a software-based solution for image stitching (mosaicking) that composes a panoramic image from the multiple captures~\cite{Im:ECCV:2016, Won:ICRA:2019}.
Compact camera rigs are often coupled to a rotating base for capturing stereoscopic panoramas with small vertical FoV or employ multiple fish-eye lenses for wider-FoV vision~\cite{Anderson:TGA:2016, Kim:JCVIU:2015, Moreau:SITIS:2012}.
Professional devices can produce 
high-quality panoramic images but are expensive and demand a time-consuming post-processing step for image stitching~\cite{Fangi:ISPRS:2018}.
Prominent examples include the Panono camera\footnote{\url{https://www.panono.com/}}, the Vuze camera\footnote{\url{https://vuze.camera/camera/vuze-camera}}, the Facebook Surround360\footnote{\url{https://github.com/facebook/Surround360}}, and the GoPro Odyssey\footnote{\url{https://gopro.com/en/us/news/here-is-odyssey}}.

More recently, many manufacturers introduced cheap and portable devices equipped with two opposite located wide-FoV fish-eye lenses~\cite{Im:ECCV:2016, Su:NIPS:2017, Shan:ACCESS:2018}, \tlts{ and a schematic illustration is shown in Fig. \ref{fig:cameras}(c).}
Each sensor captures a hemispherical image with a diagonal FoV slightly larger than 180$^\circ$, producing overlapping regions that are suitable for two-image full-FoV stitching~\cite{Deng:WCICA:2008,Lo:ICIP:2018}.
These affordable cameras greatly democratized and simplified the acquisition of real-world 360\degree content~\cite{Jung:VR:2019}.
The facilitated access to full-FoV imagery boosted the AR/VR/MR industry and research on related fields.
Examples of those capturing devices are the Ricoh Theta,
the  
Samsung Gear 360,
\tlts{the Insta360 cameras\footnote{\tlts{\url{https://www.insta360.com/}}},}
the LG 360 CAM\footnote{\url{https://www.lg.com/us/mobile-accessories/lg-LGR105AVRZTS-360-cam}}, and the Nikon Keymission 360$^\circ$\footnote{\url{https://www.nikonusa.com/en/nikon-products/product-archive/action-camera/keymission-360.html}}.


\tlts{In summary, catadioptric cameras have a historical value but are no longer widely considered for omnidirectional imaging. Polydioptric camera rigs are expensive and bulky, but can produce high-resolution panoramas. Dual-fisheye sensors are cheaper and more compact, but with limited resolution. Hence, the choice for a given capture device depends on the target application. In fact, most papers (particularly those that require single panoramas) do not mention the capture process itself, but instead focus on the \textit{representation} of the spherical images, as discussed next. }

Because of the wide diversity of technologies employed for acquiring omnidirectional content,  most computational methods prefer to work with a canonical format representation of the sphere.
The equirectangular projection 
is known as the standard mapping of the sphere to the plane~\cite{Eder:3DV:2019, Coors:ECCV:2018} and it is widely employed among both camera vendors and researchers~\cite{Su:NIPS:2017}, allowing easy pixel mapping from plane to sphere (and vice-versa). 
In the formulation presented in Section~\ref{ssec:sphimage}, every imaged  point $\mathbf{x}$ lies on the surface of a unit sphere. 
Therefore, $\mathbf{x}$ can be rewritten in terms of spherical coordinates $(r=1,\theta,\phi)$ as~\cite{Akihiko:WOVCNNC:2005} 
\begin{equation}
\label{eq:sphericaltocartesian}
\mathbf{x} = 
[\cos\theta\sin\phi\,\, \sin\theta\sin\phi\,\, \cos\phi]^\top,
\end{equation}
where $\theta \in [0,2\pi)$ and $\phi \in [0,\pi)$. 
Since there is information associated to every position $(\theta,\phi)$ on the sphere -- \ie the light intensities captured by the spherical camera -- the whole image can be organized in a $[0,2\pi)\times[0,\pi)$ planar representation.
More precisely, the intensity value from the projected point $\mathbf{x}$ is mapped to an integer pixel position $(x,y)$ of a $w\times h$ 
\tlts{equirectangular}
image where ${x} = \lfloor\frac{\theta w}{2\pi}\rfloor$ and ${y} = \lfloor\frac{\phi h}{\pi}\rfloor$. 
The 
\tlts{equirectangular}
mapping is equivalent to the so-called latitude-longitude maps~\cite{Zhao:IJCV:2014, Gava:ICIP:2018, Kim:IJCV:2013}, differing only by translation factors on $\theta$ and $\phi$.

\begin{figure}
    \centering
    \subfloat[]{\includegraphics[height=2.5cm]{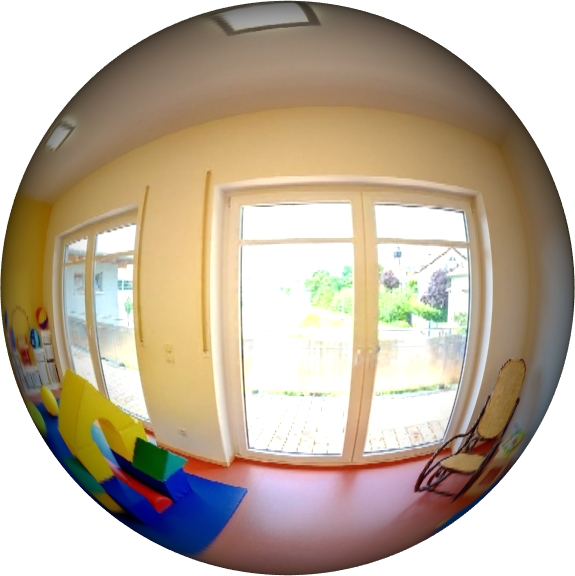}}\qquad
    \subfloat[]{\includegraphics[height=2.5cm]{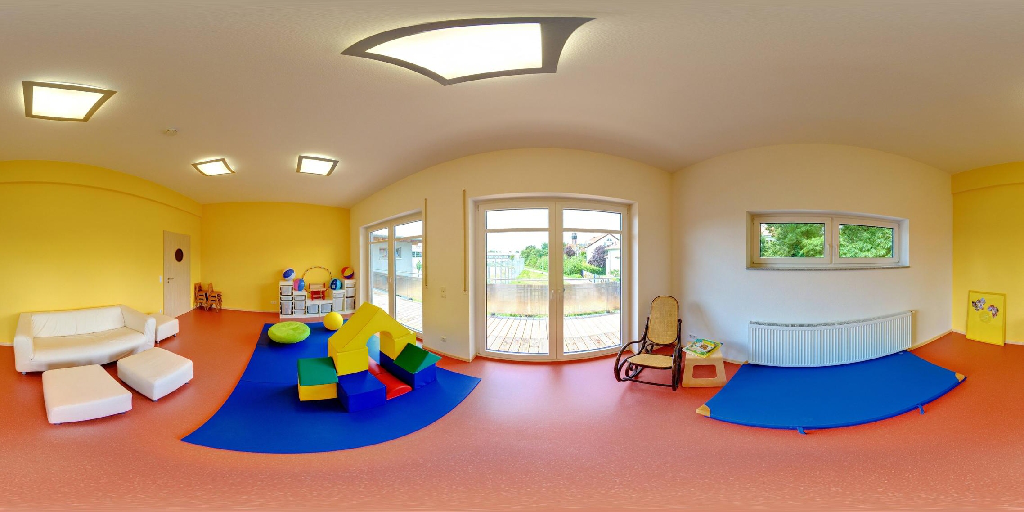}}\qquad
    \subfloat[]{\includegraphics[height=2.5cm]{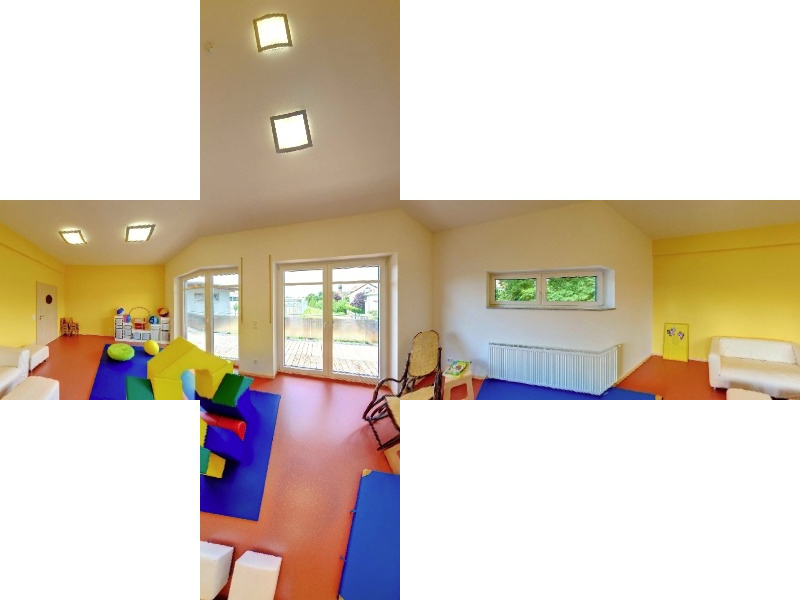}}
    \caption{
    Ominidirectional image representations: (a)
    a spherical image (projected orthographically) and its mapping to (b) 
    \tlts{equirectangular}
    and (c) 
    \tlts{cube-map}
    formats. Sample image taken from the SUN360 dataset~\cite{Xiao:CVPR:2012}.}
    \label{fig:imagerepresentation}
\end{figure}

The 
\tlts{equirectangular}
projection distorts the objects differently depending on their location on the imaged scene~\cite{Cruz-Mota:IJCV:2012}, particularly
near the poles of the sphere~\cite{Ferreira:CAG:2017}. 
These deformations result from the non-uniform sampling of the sphere~\cite{Sun:SPL:2017}.
There are other ways to represent the sphere on the plane, but all of them introduce some distortion as well~\cite{Su:NIPS:2017}.
In fact, the distortions induced by spherical projection depend on the FoV amplitude~\cite{Silveira:ICIP:2018, Zhang:ICRA:2016}.
%
A common approach for representing images with smaller deformation is to project a 90\degree vertical and horizontal FoV to each face of a cube circumscribed around the sphere. 
This format is known as cube-map projection 
\cite{Dai:VR:2019, Silveira:ICIP:2018} or sky-box \cite{Song:CVPR:2018}.
Fig.~\ref{fig:imagerepresentation} depicts the mapping of a  spherical image 
to its  
\tlts{equirectangular}
and 
\tlts{cube-map}
representations.

Emerging representations are based on successive divisions of a 3D geometric form and aim to mitigate even more the distortions from the spherical model.
The fine-grained mesh based on an icosahedron~\cite{Li:VR:2005, Sun:SPL:2017, Zhao:IJCV:2014, Guan:CVPR:2017} is by far the most popular, and it is also known as icosphere~\cite{Eder:WCVPR:2019} or tangent planes projection
\cite{Eder:CVPR:2020}. 
Octahedron-based 360\degree images are also found in the literature 
\cite{Sun:SPL:2017, Lee:TPAMI:2020}.
It is worth mentioning that many other sphere-to-plane representations can be imported from cartography~\cite{Eder:CVPR:2020}, but the resulting images can be built from an 
\tlts{equirectangular}
input image.
However, one may note that exchanging formats may induce loss of information and artifacts~\cite{Azevedo:TCSVT:2019}, mostly because of the interpolation operations needed to cope with sub-pixel transformations~\cite{Coors:ECCV:2018}.

\tlts{In the remainder of this paper, we will refer to the equirectangular projection (ERP), cube-map projection (CMP), tangent planes projection (TPP), cylindrical projection (CP), and hemispherical projection (HP) by their acronyms here introduced.}
\tlts{Other format representations, such as the SpherePHD~\cite{Defferrard:ICLR:2020} and Equirectangular to Perspective (E2P)~\cite{Yang:CVPR:2019}, will be further explained since they are tightly coupled to the layout or depth estimation methodologies where they are defined.}


\section{Monocular 3D Geometry Estimation} \label{sec:monocular}

Inferring depth from two or more views of a scene is a well-studied problem. 
The literature on \emph{single-image stereo} for 
pinhole cameras is more recent, and one of the first methods that attempted to \textit{learn} depth from color images was based on Markov Random Fields 
\cite{Saxena:NIPS:2006}. 
However, a new era started with the seminal work of Eigen \etal \cite{Eigen:NIPS:2014}, who explored deep architectures for depth inference. 
Early approaches for room layout (sparse global 3D information) extraction from pinhole images rely on geometrical cues like vanishing lines and points~\cite{kovsecka:eccv:2002}, but results started to improve significantly with the use of deep learning
\cite{mallya:iccv:2015}.

The complete FoV of spherical images allows the estimation of full depth images or layout from a single capture. Methods based on spherical images can either cast the problem as a set of planar projections or directly work on the spherical domain. For the latter set of methods, 
\tlts{deep learning} is the dominant strategy nowadays, and several modifications have been proposed to traditional architectures so they can perform better on panoramas. This section starts by reviewing \tlts{general approaches} 
that adapt/tailor neural networks to handle panoramas better.  Then, it revises \tlts{techniques} for 3D layout extraction and dense depth estimation using a single panorama. 


Table~\ref{tab:monocularmethods} summarizes the single-panorama approaches reviewed in this section, highlighting: the main goal (layout or depth); the spherical representation used as input (ERP, CMP, TPP, E2P or \textit{SpherePHD}, as described in this section, or the previous one); the core technology explored by the method (where DNN denotes a planar \emph{deep neural network} with convolutional and possibly fully-connected layers, and SDNN indicates if some network adaptation to the spherical domain was used); and density of the produced 3D representation (mostly sparse or dense).
\tlts{Except for 
the works 
\cite{Tateno:ECCV:2018} and 
\cite{Payen:ECCV:2018}, all the others listed in Table~\ref{tab:monocularmethods} assume an indoor scene capture as input.}






\begin{table}[]
\footnotesize
\caption{3D geometry estimation methods based on a single spherical image.}
\centering
\label{tab:monocularmethods}
\begin{tabulary}{\linewidth}{lccccc}
\hline
\textbf{Reference} & \textbf{Category} &  \textbf{Representation} &  \textbf{Technology} &  \textbf{Density}  \\
\hline 
Shum \etal~\cite{Shum:CVPR:1998} & Layout & ERP  & Interactive & Sparse \\
Yang and Zhang~\cite{Yang:VRCAI:2014} & Layout & TPP & Semi-interactive & Sparse \\
Jia and Li~\cite{Jia:ICRA:2015} & Layout & TPP & Optimization & Sparse \\
Fukano \etal~\cite{Fukano:ICPR:2016} & Layout & CMP & Optimization & Sparse \\
Pintore \etal~\cite{Pintore:WACV:2016} &  Layout & ERP & Optimization & Sparse \\
Xu \etal~\cite{Xu:WACV:2017} & Layout & TPP & Bayesian Optimization & Sparse \\ 
Fernandez-Labrador~\etal~\cite{Fernandez-Labrador:LRA:2020}  & Layout & ERP & CNN/SCNN & Sparse \\ Fernandez-Labrador~\etal~\cite{Fernandez-Labrador:LRA:2018}  & Layout & TPP & Geometry/CNN & Sparse \\ 
Fernandez-Labrador~\etal~\cite{Fernandez-Labrador:arXiv:2018}  & Layout & ERP & Geometry/CNN & Sparse \\ 
Zou~\etal~\cite{Zou:CVPR:2018}  & Layout & ERP & Geometry/CNN & Sparse \\ 
Yang~\etal~\cite{Yang:CVPR:2019}  & Layout & ERP/E2P & CNN/BLSTM & Sparse \\ 
Sun~\etal~\cite{Sun:CVPR:2019} & Layout & ERP & DNN & Sparse \\
Pintore~\etal~\cite{Pintore:ECCV:2020} & Layout & $2\times$ E2P & CNN/BLSTM & Dense \\
Zou~\etal~\cite{zou:ijcv:2021} & Layout & ERP/E2P & Several & Sparse \\
Zioulis~\etal~\cite{Zioulis:arxiv:2021}  & Layout  & ERP  & DNN  & Sparse  \\
Yang and Zhang~\cite{Yang:CVPR:2016} & Layout/Depth & ERP & Optimization & Dense \\ 
Yang~\textit{et al.}~\cite{Yang:CVPR:2018}  & Layout/Depth & TPP & CNN & Dense \\
Sun \etal~\cite{sun:arxiv:2020} & Layout/Depth & ERP & DNN/Attention & Dense \\
Jin \etal~\cite{jin:cvpr:2020} &  Layout/Depth & ERP & CNN/Attention & Dense \\
Wang \etal~\cite{wang:arxiv:2021} & Layout & ERP & DNN/BLSTM & Sparse \\
Silveira~\etal~\cite{Silveira:ICIP:2018}  & Depth & TPP & DNN & Semi-dense \\
Zioulis~\textit{et al.}~\cite{Zioulis:ECCV:2018}  & Depth & ERP & DNN & Dense \\
de La Garanderie~\etal~\cite{Payen:ECCV:2018} & Depth & ERP & SDNN & Dense\\
Eder~\etal~\cite{Eder:3DV:2019}  & Depth & ERP & DNN & Dense \\
Tateno~\etal~\cite{Tateno:ECCV:2018}  & Depth & ERP & DNN & Dense \\
Zioulis~\etal~\cite{Zioulis:3DV:2019}  & Depth & ERP & SDNN  & Dense  \\
Wang~\etal~\cite{Wang:CVPR:2020} & Depth & ERP/CMP & SDNN & Dense \\
Lee~\etal~\cite{Lee:TPAMI:2020} & Depth & \textit{SpherePHD} & SDNN & Dense  \\
Jiang~\etal~\cite{jiang:arxiv:2021} & Depth & ERP/CMP & DNN & Dense \\
\hline
\end{tabulary}
\end{table}

\subsection{Spherical Neural Networks} \label{subsec:spherenn}


Convolutional Neural Networks (CNNs) have been widely used in  problems such as image classification, object detection, segmentation, and single-image depth estimation~\cite{Redmon:CVPR:2016,Ronneberger:MICCAI:2015,Eigen:NIPS:2014}. \tlts{However, applying ``planar'' CNNs directly to panoramas presents some challenges.}
\tlts{There are three major issues when applying planar CNNs to ERP images. 
First, these techniques do not take into account the circularity property of the ERP images, which may lead to horizontal discontinuities \cite{Lai:VR:2019, Zioulis:3DV:2019}.
Second, standard CNNs neglect the non-uniform sampling of ERP images and apply convolutional kernels (same applicable to poolings, etc.) of fixed size all over the image~\cite{Su:NIPS:2017,Coors:ECCV:2018}.
For spherical images, particularly those in the ERP format, a larger receptive field would be desirable near the poles 
since they are oversampled.
Third, most standard CNNs are trained using planar images, that when applied to panoramas might lead to domain gap issues and produce unsatisfactory results~\cite{Tateno:ECCV:2018, Su:NIPS:2017, Silveira:ICIP:2018}.
Applying standard CNNs to CMP images is also not straightforward.
Each cube face has a smaller field of view (90$^\circ$) but still contain heavy ``radial'' distortions, unlike most perspective images. 
Furthermore, the cube faces need to be processed individually and then stitched together back to the sphere. 
Such approaches commonly lead to discrepancies in each face and may demand a postprocessing step~\cite{Silveira:ICIP:2018,Zioulis:ECCV:2018}. Although our discussion focused on ERP and CMP representations, these considerations are valid to any sphere-to-(multi)plane mapping since they either present non-uniform sampling distortions or require stitching to cover the whole sphere. }

Other researchers focused on designing spherical CNNs (SCNNs) that can better handle the distortions in 360\degree images. \tlts{Next, we describe some recent methods that try to mitigate the non-uniform sampling problem in panoramas in a generic deep-learning perspective. \emph{Not all these methods have been effectively explored in the context of 3D estimation using panoramas, but they can be potentially used in the future}}.


\tlts{A simple solution to the handle the horizontal field-of-view circularly  of spherical images in ERP format is to adapt a convolutional layer by adding circular padding, as explored in~\cite{Wang:ICRA:2018,Schubert:IV:2019}. This idea (or an alternative by applying a larger circular pad to the input panorama) was explored to obtain optical flows in~\cite{Silveira:VR:2019} and 3D layout from a single panorama  in~\cite{Sun:CVPR:2019,Zioulis:arxiv:2021}.}

Some authors design SCNNs by using 3D spherical coordinates. For example, Cohen \etal \cite{Cohen:ICLR:2018} propose spherical convolutions, with the spherical coordinates represented in a 3D manifold. To achieve rotation invariance, the convolutions are performed in the spectral domain, and pooling is easily extended.  
Esteves \etal \cite{Esteves:ECCV:2018} achieves rotational equivariance by defining convolutions in the spherical harmonic domain. 
Both approaches were tested in classification tasks.
As noted in~\cite{jiang:ICLR:2019},  rotation-invariance is not always desired, like in the toy problem of differing digits ``9'' and ``6'' in the context of object classification. Furthermore, these networks 
present a computational overhead due to special functions like the Spherical Fourier Transform.

Other approaches use an approximation or sampling of the sphere. Defferrard \etal \cite{Defferrard:ICLR:2020} uses HealPix sampling to represent the sphere and define the convolution in spectral-domain, where pooling is naturally extended. Similarly, Lee \etal \cite{lee:cvpr:2019} uses a geodesic representation to map the sphere and a triangle base as a pixel point where a proper convolution and pooling are easily defined, naming it as \textit{SpherePHD}. 
Both works report outperforming results than planar CNNs in classification, object detection, and segmentation tasks\tlts{, and later explored for depth estimation in~\cite{Lee:TPAMI:2020}. } Jiang \etal~\cite{jiang:ICLR:2019} perform convolutions using Parametrized Differential Operators
designed for unstructured grids based on differential operators of different orders, which is more accurate than  using only the immediate neighborhood as claimed by the authors. Zhao \etal~\cite{Zhao:2019:IPMI} suggest a Direct Neighbor 
convolution that performs the convolution with seven values (direct neighbors and the ``center of kernel''), duplicating values if necessary. 
Although these previous approaches are faster than networks defined directly on the sphere, employing an approximation based on icosahedrons 
produces tight constraints on the number of vertices: $n = 5 \times 2^{2l+1}+2$ vertices are generated for a given subdivision $l$.

Other strategies were not designed specifically for spherical image processing, but can be adapted do deal with the non-uniform sampling issue. First,
Liu \etal \cite{liu:nips:2018} propose a coordinate convolution layer (\textit{CoordNet})  that adds coordinate information to the standard convolution to learn the pattern between the feature map and the associated spatial location. This strategy was explored \tlts{for computing the optical flow in panoramas}~\cite{Xie:3DV:2019} \tlts{and also for layout estimation in~\cite{Zioulis:arxiv:2021}.}  
On the other hand, deformable kernels \cite{Dai:ICCV:2017} have been explored in spherical imaging, where the kernel support is adapted according to each spatial location of the ERP image.
Many authors employ this strategy in tasks involving panoramas, such as classification \cite{Coors:ECCV:2018}, depth estimation \cite{Tateno:ECCV:2018}, saliency detection 
\cite{Zhang:ECCV:2018}, and layout estimation \cite{Fernandez-Labrador:LRA:2020}. 

Graph neural networks 
can inherently handle non-uniformly sampled data, and hence can be explored for panoramas. Frossard \etal \cite{Frossard:ICCV:2017}  claim that a proper weight design is crucial for graph convolution operators to 
obtain the same graph signal response at different distortion levels. Although 
\tlts{graph networks}
are very flexible \wrt the nodes (pixels) geometric distribution, their computation time is typically higher due to operations (e.g., pooling and convolutions) that act over irregular sampled data.


\subsection{Layout Estimation} \label{subsec:layout}

The main goal of layout estimation techniques is to obtain a sparse 3D representation of an \tlts{\textit{indoor}} scene. 
In general, the rooms are modeled as cuboids (which can be a reasonable model for simpler rooms) or with a more general hypothesis \tlts{such as} 
Manhattan worlds. 
Spherical cameras are particularly interesting for this application since a single capture can be used as input.

Shum \etal~\cite{Shum:CVPR:1998} present a pioneer work for layout estimation from panoramas. They propose an interactive system in which the user draws primitives (points, lines, planes) that allows the estimation of the camera pose. Then, a set of constraints (such as known points or segments) are used to obtain the 3D representation.  
Yang and Zhang~\cite{Yang:VRCAI:2014} introduce a semi-supervised approach for layout recovery.
In the automatic part, narrower FoV views are extracted from the panorama, from which geometrical cues such as orientation maps and geometric context
are computed. Candidate box layouts are projected onto the perspective views, where geometric cues are used to find the best box candidate. 
Finally, the extracted box can be manually edited to allow more generic Manhattan layouts. 
Fukano and colleagues~\cite{Fukano:ICPR:2016} follow a similar approach: they project the panorama 
to the cube, compute line primitives for each face and combine them back in the panorama. 
These primitives are separated into potential plane boundaries or not through a global labeling approach, and plane-related segments are used to find the cuboid. 
Clearly, the need for manual interaction is a drawback of~\cite{Shum:CVPR:1998,Yang:VRCAI:2014}, and the restriction to cuboids is a limitation of~\cite{Fukano:ICPR:2016}.

One of the first works on automatic cuboid layout estimation was proposed by Yang and Zhang~\cite{Yang:CADG:2013}. They assume a full horizontal view panorama with a central horizon, which is divided into four non-overlapping sub-images that are later undistorted based on 
TPP. They explore a parameterization method based on vanishing polylines to encode all four layouts into a single full-room layout, which is predicted via parameter candidate evaluation using a linear scoring function. 
For training the weights of the scoring function, they use a structured Support Vector Machine (SVM) on a non-panoramic indoor image dataset. 
\textit{PanoContext}~\cite{Zhang:ECCV:2014} recovers a cuboid room layout, the bounding volumes and the semantics of common furniture.
The core idea is to generate hypotheses for the room layout and object bounding volumes using lower-level information (edges, segmentation, orientation, and context) and rank them using an SVM.
Although \textit{PanoContext} expands the analysis over~\cite{Yang:CADG:2013} by adding object bounding volumes, it also assumes a simple primitive for the box layout. 
\textit{Pano2CAD}~\cite{Xu:WACV:2017} extends the idea of~\cite{Zhang:ECCV:2014} by allowing Manhattan layout models, enhancing the 3D object detection, and also adding context priors to relate pieces of objects (furniture) to walls. 
\tlts{As in \textit{PanoContext}~\cite{Zhang:ECCV:2014}}, \textit{Pano2CAD} focuses on images of living rooms and bedrooms, and the computational cost is high.


Geometric primitives are also explored by Jia and Li~\cite{Jia:ICRA:2015}, who adapt the Canny edge detector to the spherical domain for finding line segments,  corners and vanishing points.
They use heuristic rules based on a combination of these primitives for generating a score function that leads to the desired 3D layout when maximized. Yang and Zhang~\cite{Yang:CVPR:2016} also explore line primitives, combining them with spherical superpixels (based on~\cite{felzenszwalb:ijcv:2004})
for inferring layout and depth. 
They assume planar depth values within the superpixels and impose constraints based on neighborhood and context (ground or wall) using a constrained linear least-squares 
formulation. Their output is not precisely a room layout, but a dense depth map built with Manhattan world assumptions, in which pixels present semantic labels (walls, ceiling, floor). Clearly, the results of~\cite{Jia:ICRA:2015,Yang:CVPR:2016} are strongly dependent on the computation of the primitives, which might fail in low contrast/texture regions. Pintore and colleagues~\cite{Pintore:WACV:2016} also explore geometrical primitives but relaxed the Manhattan world assumption. Their method starts by labeling the panorama as walls, floor, and roof using~\cite{Cabral:CVPR:2014}. Then, they propose a projection function to map image edges onto the ground floor (assuming known camera height), and use a Hough transform-like voting scheme to obtain the room shape. On the one hand, this approach is more generic for not assuming Manhattan worlds. On the other, it implicitly assumes that the panorama is gravity-aligned,  which might not be accurate in practical situations~\cite{bergmann:sibgrapi:2021}.

The availability of larger annotated datasets, starting with the PanoContext dataset in 2014~\cite{Zhang:ECCV:2014}, leveraged the development of 
\tlts{deep learning}
approaches for layout estimation. However, applying CNNs directly to the ERP is not ideal due to both the space-varying distortions introduced by this projection near the poles and the loss of spatial correlation between the left and right extremities of the panorama~\cite{Cohen:ICLR:2018}. 
\tlts{As such, some deep learning
methods for layout estimation explore or propose some kind of data or network adaptation to better explore the geometry of spherical images, as discussed next.}


To the best of our knowledge, the first 
\tlts{deep learning}
approaches for layout estimation date from
2018 and combine geometrical cues with deep networks. Fernandez-Labrador \etal~\cite{Fernandez-Labrador:LRA:2018} first extract vanishing points to upright align the panorama, and split it into 60 overlapping narrow-FoV perspective images.
Then, they apply a pre-trained CNN~\cite{mallya:iccv:2015} to each projection, extracting local edge maps that are stitched back to the panorama.
Finally, they perform edge pruning, corner extraction, and classification to generate a Manhattan layout, noting that normal maps extracted from the perspective projections using \cite{Eigen:ICCV:2015} are used for refining the layout.
\textit{LayoutNet}~\cite{Zou:CVPR:2018} also explores geometrical cues and deep networks for layout estimation using a single (perspective or ERP) image. The authors firstly gravity-align the panorama, and then feed it to a two-branched encoder-decoder CNN with the same architecture: the first branch provides wall-wall, ceiling-wall, and wall-floor boundary maps, whereas the second one predicts a corner map and receives skip-connections from the former.  The CNN output feeds a 3D layout regressor that outputs the cuboid layout.  
Finally, \textit{LayoutNet} uses geometric reasoning for refining the layout respecting the Manhattan world assumption.
\textit{PanoRoom}~\cite{Fernandez-Labrador:arXiv:2018} builds on top of~\cite{Fernandez-Labrador:LRA:2018}, but working directly on the ERP image.
An encoder-decoder CNN with skip connections based on ResNet-50 predicts both corner and edge probability maps on separate output channels of a single branch. 
Geometric reasoning based on vanishing lines 
(as in~\cite{Fernandez-Labrador:LRA:2018})  estimates a Manhattan-constrained layout, showing superior quantitative results when compared to~\cite{Zhang:ECCV:2014,Fernandez-Labrador:LRA:2018,Zou:CVPR:2018}.

Other works propose end-to-end 
\tlts{deep learning}
approaches that do not require additional geometric reasoning. \textit{DuLa-Net}~\cite{Yang:CVPR:2019} explores and encoder-decoder architecture with two-branches, each one attaining a different panorama representation:
one branch is fed with an ERP image and outputs both a floor-ceiling probability map and a layout height estimation; the other receives a perspective ceiling-view, computed by an Equirectangular to Perspective (E2P) algorithm, and outputs a floor plan probability map. The final 3D layout 
is computed by first transforming the panorama branch output using E2P, and separating 
floor and ceiling probability maps, 
which are combined with the ceiling-branch output map to create a fused floor plan probability map. The fused map and the estimated height allow the creation of the final 3D layout. According to the authors, 
\textit{DuLa-Net} outperforms both \textit{PanoContext} and \textit{LayoutNet}. 
\textit{HorizonNet}~\cite{Sun:CVPR:2019} encodes a panoramic layout using three vectors with the same dimension as the image width (in ERP format), which implicitly handles the distortions of spherical images along the rows. For each column, their representation provides the vertical coordinates of ceiling-wall and floor-wall boundaries,
and the probability of wall-wall boundaries. 
As~\cite{Fernandez-Labrador:LRA:2018}, \textit{HorizonNet} employs a ResNet-50 backbone and explores column-wise recurrent information using a bidirectional long short-term memory (\tlts{BLSTM}). Finally, Manhattan world constraints are added to the network output for generating a closed layout. The follow-up study \textit{HoHoNet}~\cite{sun:arxiv:2020} also explored the idea of encoding layout (and depth) information along a column-wise vector, but as a latent representation extracted with a feature pyramid as backbone and combined by an attention layer~\cite{vaswani:nips:2017}. 
The latent vector is then unwrapped to the 2D spatial domain using the inverse discrete cosine transform, 
enabling either layout or depth estimation. 
For layout estimation, they used the same strategy as~\cite{Sun:CVPR:2019}, reporting 
\tlts{state-of-the-art} results. 
\textit{LED$^2$-Net}~\cite{wang:arxiv:2021} also builds on top of~\cite{Sun:CVPR:2019} in terms of the network architecture. 
However, they encode the room layout as a set of equiangulary (in terms of longitude) spaced rays on the floorplan (and ceiling), which leads to a differentiable loss function. 
Based on this ``horizon-depth'' representation, the 3D layout is obtained by extrusion assuming a gravity-aligned camera with approximately known height (plus a post-processing step to impose Manhattan worlds, as in~\cite{Sun:CVPR:2019}). 
The authors show 
\tlts{state-of-the-art}
results on different datasets and an interesting cross-dataset validation experiment.

Moving toward the relaxation of geometric constraints of the layout, Pintore \etal~\cite{Pintore:ECCV:2020}  assume Atlanta worlds in which the room presents a horizontal floor and ceiling, but the vertical walls are not necessarily orthogonal to each other (they may even be curved). Their  
core idea is to project the panorama into planes above and below the camera (ceiling and floor projections), similarly to the E2P step used in~\cite{Yang:CVPR:2019}.  
They proposed an encoder-decoder architecture that takes floor (or ceiling) projections and regresses a binary mask segmenting the floor (or ceiling). Their network also uses a ResNet-50 backbone, followed by a reshaping step and a 
\tlts{BLSTM}
layer that captures long-range patterns of the layout. Finally, a series of convolutional layers followed by linear interpolation decode
the sequential feature map to the output binary mask.  The 2D layout is a polygon approximation of the ceiling segmentation mask, and the room height is chosen as to maximize the correlation between the ceiling and floor projections. 
Their results for cuboid-like layouts are comparable to~\cite{Yang:CVPR:2019,Sun:CVPR:2019}, but improve as the layout complexity increases. \textit{Corners for Layout} 
\cite{Fernandez-Labrador:LRA:2020} can
handle even more generic layouts by predicting pixel-wise heatmaps for both corners and edges, with the only assumption of floor-ceiling parallelism. 
These maps are regressed by two end-to-end
fully convolutional encoder-decoder architectures with the same structure,
including as an option the use of spherical convolutions (similar to \cite{Tateno:ECCV:2018}) to account for the distortions of the ERP format.
Their experimental results demonstrate that their SDNN is more robust to distortions induced by camera alignment, but standard convolutions are competitive if the input panoramas are gravity-aligned. \textit{Single-Shot Cuboids}~\cite{Zioulis:arxiv:2021} also regresses corners, but without any additional information or post-processing, at the cost of restricting the geometry to a cuboid.
This allows end-to-end training using a network without any spherical adaptation except of circular padding 
(but exploring recent advances such as anti-aliasing pooling~\cite{zhang:icml:2019}).
Their method is considerably lighter (in terms of parameters) than competitive approaches such~\cite{Sun:CVPR:2019,Fernandez-Labrador:LRA:2020}, but such simplicity comes with the cost of a cuboid-shaped layouts only.

Providing a full comparative analysis of layout extraction methods is a challenging task. 
There is a coupling between how generic the method is in terms of layout geometry (cuboid, Manhattan, Atlanta, no restriction) and the characteristics of the images in the evaluated datasets, which also relates to the availability of publicly available datasets and standardized metrics and annotations (see Section~\ref{sec:benchmark} for a discussion). Recently, Zou and collaborators~\cite{zou:ijcv:2021} provided an in-depth comparative analysis of~\cite{Zou:CVPR:2018,Yang:CVPR:2019,Sun:CVPR:2019}, discussing how these methods can be divided into similar components implemented with different strategies and what are the effects of each design decision on the results. 
They also propose improved versions of \textit{LayoutNet} and \textit{DuLa-Net} (named \textit{LayoutNet} v2 and \textit{DuLa-Net} v2). 
Their analysis indicated that the quantitative results achieved by the methods (particularly \textit{LayoutNet} v2 and \textit{HorizonNet}) may vary considerably depending on the test dataset. 

\begin{figure}
    \centering
    \subfloat[]{\includegraphics[width=.24\textwidth]{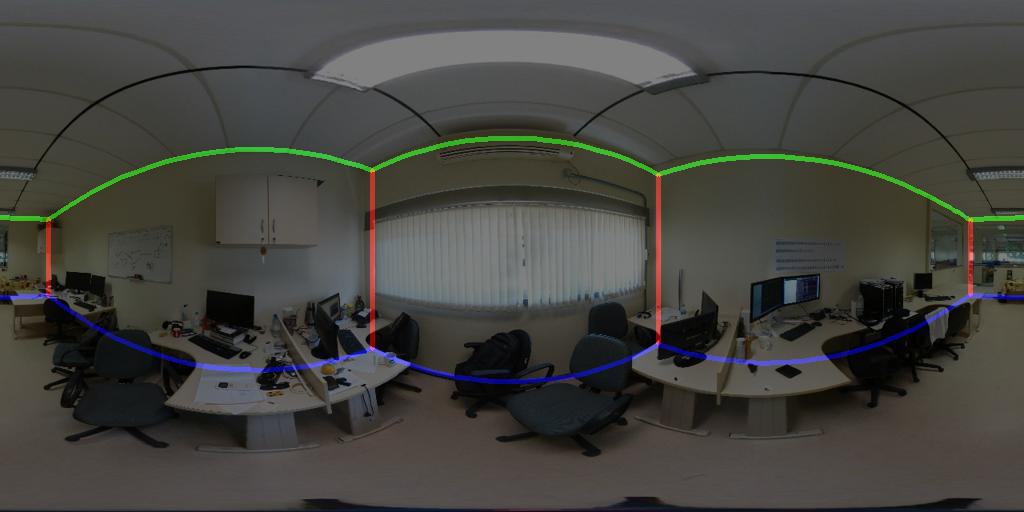}}\,
    \subfloat[]{\includegraphics[width=.24\textwidth]{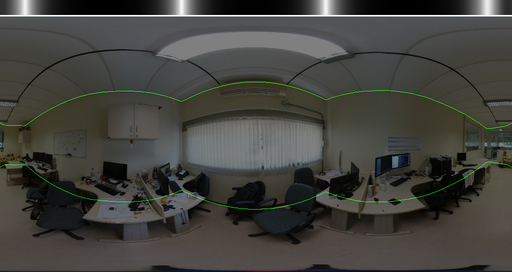}}\,
    \subfloat[]{\includegraphics[width=.24\textwidth]{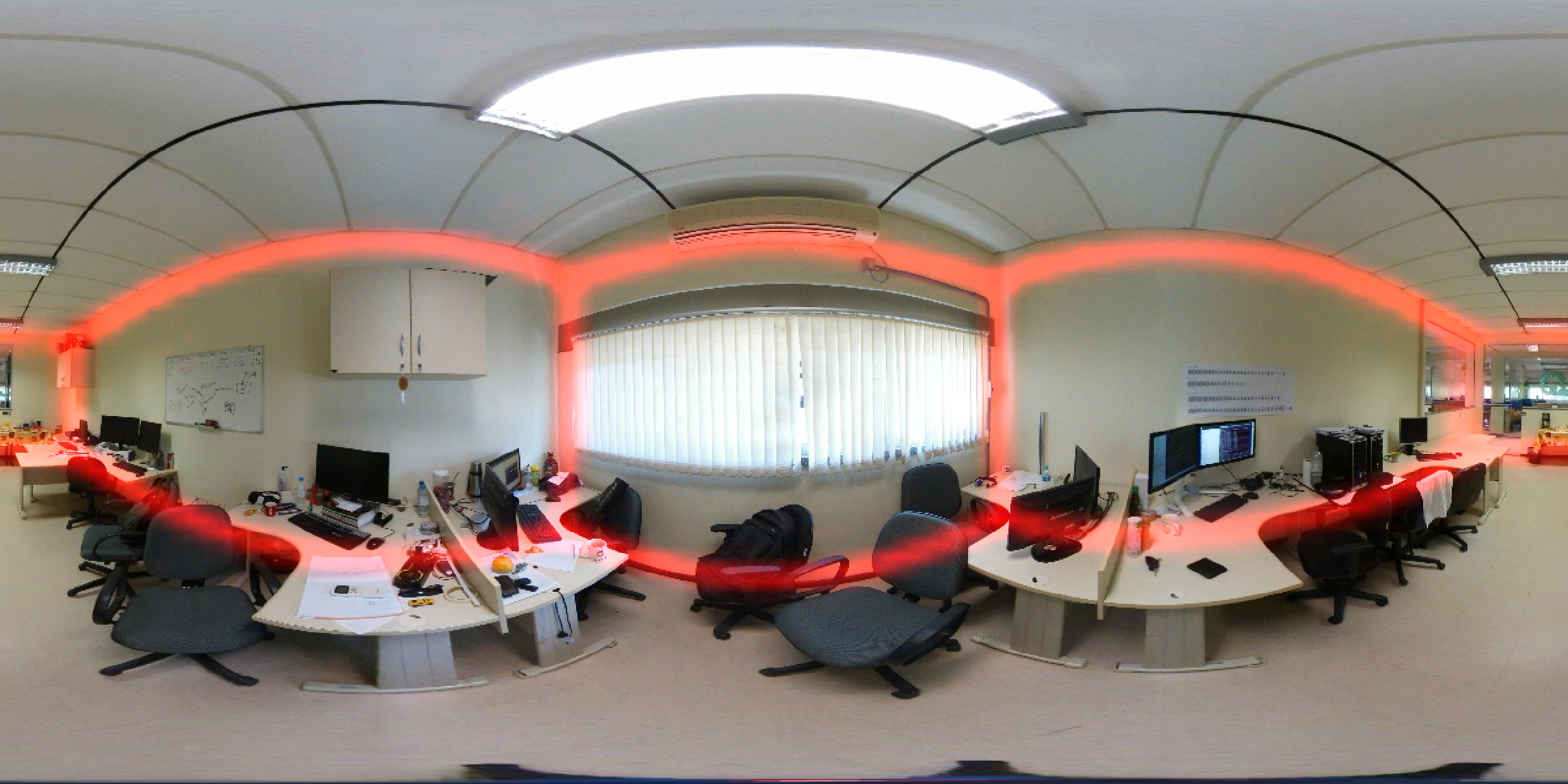}}\,
    \subfloat[]{\includegraphics[width=.24\textwidth]{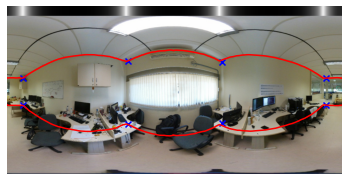}}\,
    \caption{
    Examples of layout estimation results. Color image (personal dataset) and different layout representations are overlaid for better visualization using (a) \textit{DuLa-Net}~\cite{Yang:CVPR:2019}, (b) \textit{HorizonNet}~\cite{Sun:CVPR:2019}, (c) \tlts{\textit{Corners for Layout}~    \cite{Fernandez-Labrador:LRA:2020}, and (d) \textit{HoHoNet}~\cite{sun:arxiv:2020}.} 
    Layout markers are depicted as meant by the authors in their publicly available codes.}
    \label{fig:layout_estimation}
    \vspace{-0.25cm}
\end{figure}

Fig.~\ref{fig:layout_estimation} shows the layout estimates for an unseen scenario (image from our personal dataset) obtained by  
\tlts{four state-of-the-art}
approaches~\cite{Yang:CVPR:2019,Sun:CVPR:2019,Fernandez-Labrador:LRA:2020,sun:arxiv:2020}. 
The input ERP image and the estimates are overlaid for better visualization, and each estimate is represented as meant by the authors of the original papers~\cite{Yang:CVPR:2019,Sun:CVPR:2019,Fernandez-Labrador:LRA:2020,sun:arxiv:2020}. Fig.~\ref{fig:layout_estimation}(a) distinguishes horizontal (ceiling and floor) and vertical layout edges, whereas Figs.~\ref{fig:layout_estimation}\tlts{(b), (c), and (d)} present unified edge maps. Fig.~\ref{fig:layout_estimation}(b) \tlts{and (d)} still generate a 1D probability map for the existence of wall-wall boundaries (thin grayscale image on the top).


\subsection{Depth Estimation} \label{subsec:monodepth}

One of the first works tackling the single-image depth estimation problem under the spherical optics can be found in~\cite{Silveira:ICIP:2018}, where
the authors propose to extract and process multiple overlapping narrow-FoV planar projections from the spherical image. 
More precisely, each planar image is individually fed to a (black box) depth estimation method trained with perspective images
and then the resulting depth maps are back-projected to the sphere. 
%
The minimization of depth differences and alpha-blending provides the final output.
Despite the interesting results shown in~\cite{Silveira:ICIP:2018}, the stitching artifacts arise when merging adjacent tangent planes. The work from Yang \etal~\cite{Yang:CVPR:2018} 
also represents spherical images as a collection of narrower-FoV planar images. 
The authors extract geometric and semantic cues from the images and combine them for ground plane detection and occlusions reasoning. 
Their method uses extracted lines and superpixels to impose planarity constraints on the global scene layout under the Manhattan world assumption. 
Then, the finer details associated with typical indoor objects are estimated via depth propagation.
de La Garanderie \etal~\cite{Payen:ECCV:2018} present an approach for adapting 
DNNs developed on perspective imagery to work with panoramas, focusing on object detection and depth estimation. 
Their main idea is to consider existing datasets of perspective images as patches that are projected onto the sphere using style transfer. Also, they provide a simple adaptation of traditional CNNs to deal with panoramas by circularly padding the horizontal boundaries of ERP images (as other works for layout estimation revised in Section~\ref{subsec:layout}). 
The transformed images, however, do not cover the full panorama, and it might affect tasks that require a broader spatial context.

\textit{OmniDepth}~\cite{Zioulis:ECCV:2018} is a fully convolutional encoder-decoder network that explicitly accounts for the distortions of ERP images using dilated convolutions for adapting the receptive fields depending on the latitude. \textit{OmniDepth} is trained in a completely supervised manner with ground truth depth maps. Distortions were also explicitly taken into account by Tateno \etal~\cite{Tateno:ECCV:2018}, who introduce a deformable convolutional filtering approach applicable to dense prediction tasks. Their 
filtering approach allows to train a CNN using regular convolutions and perspective imagery, and then transfer the weights to another network with the same architecture capable of deforming the convolutions for addressing the distortions present in ERP images.
The authors report improved results regarding 
traditional CNNs (applied to either ERP or CMP images), where the artifacts are prominent.
The study from Eder and colleagues~\cite{Eder:3DV:2019} introduces a plane-aware encoder-decoder network that jointly estimates depth and normal maps from indoor scenarios. 
Besides the color image, the authors also supply the network with a latitude-longitude geodesic map that, as they claim, accounts for the irregular sampling of the ERP images (recall the main idea behind \textit{CoordNet}~\cite{liu:nips:2018}). In fact, the authors provide an ablation study showing a performance gain in normal map prediction.



Zioulis and colleagues~\cite{Zioulis:3DV:2019} present a self-supervised scheme that 
minimizes depth-based photometric error instead of depth itself, similarly to~\cite{Godard:CVPR:2017} for perspective images. Given a panorama and the corresponding depth map, they used a depth image-based rendering (DIBR) approach tailored to the spherical domain to produce new synthetic views, emulating translational (vertical or horizontal) camera movement.
To deal with distortions in the ERP format, they explored a backbone based on \textit{CoordNet}~\cite{liu:nips:2018}, in which pixel coordinates are also used as input to the network. Although the reported experimental validation is limited, the authors claim that their approach can better generalize to a wider range of images.

Some authors explore branching strategies for depth estimation. Wang \etal~\cite{Wang:CVPR:2020} takes advantage of both full-FoV ERP images and the less distorted CMP representation by using a two-branch encoder-decoder network (\textit{BiFuse}), each one processing a different image representation (a spherical padding scheme was introduced to ensure continuity between adjacent faces of the CMP image in the second branch).
They use \textit{bi-projection fusion blocks} that allow a bi-directional flow of information between the two branches that yields better results according to the authors. \textit{UniFuse}~\cite{jiang:arxiv:2021} builds on top of \textit{BiFuse}, presenting two encoders, fed with ERP and CMP representations, and just one decoder. Unlike \textit{BiFuse}, however, \textit{UniFuse} unidirectionally feeds the CMP features to the ERP features at the beginning of the decoder, arguing that ``optimizing the CMP depth may cause the training to lose focus on the ERP depth''. Jin \etal~\cite{jin:cvpr:2020} 
explore a different branching strategy for jointly estimating the layout and the depth map of indoor scenes. 
Their network, which has planar convolutions, presents two stages: one that uses layout as a prior for estimating the depth, and another one that regularizes the depth map using the layout, assuming planar walls. A semantic segmentation scheme that predicts a furniture map is also used as an attention module to alleviate the influence of furniture in the regularization stage. Both~\cite{Wang:CVPR:2020} and~\cite{jin:cvpr:2020} present better results when compared to \textit{OmniDepth}~\cite{Zioulis:ECCV:2018}. Since they were published at the same conference, they do not compare with each other. 
However, the reported results in both methods for the Stanford 2D-3D-S dataset~\cite{Armeni:arXiv:2017} are very similar, whereas \textit{UniFuse} shows slightly better results. 

It is also worth mentioning the non-Euclidean spherical polyhedron representation (\textit{SpherePHD}) for panoramas presented in~\cite{lee:cvpr:2019}, and its extended version that tackled the depth inference problem in~\cite{Lee:TPAMI:2020}. The \textit{SpherePHD} structure represents a 360\degree image by triangular tessellation, in which each triangle individually has a small distortion. These triangles are then re-indexed and arranged so that conventional CNNs can be applied to the proposed representation. For depth estimation, the authors proposed an encoder-decoder scheme with residual layers using the new representation. Although they do show improved results \wrt ERP or CMP representations for this task, no comparison with other 
\tlts{state-of-the-art}
approaches were provided (in fact, depth estimation was not the main goal of the paper, but used as a potential application).

\begin{figure}
    \centering

    \includegraphics[width=0.24\textwidth]{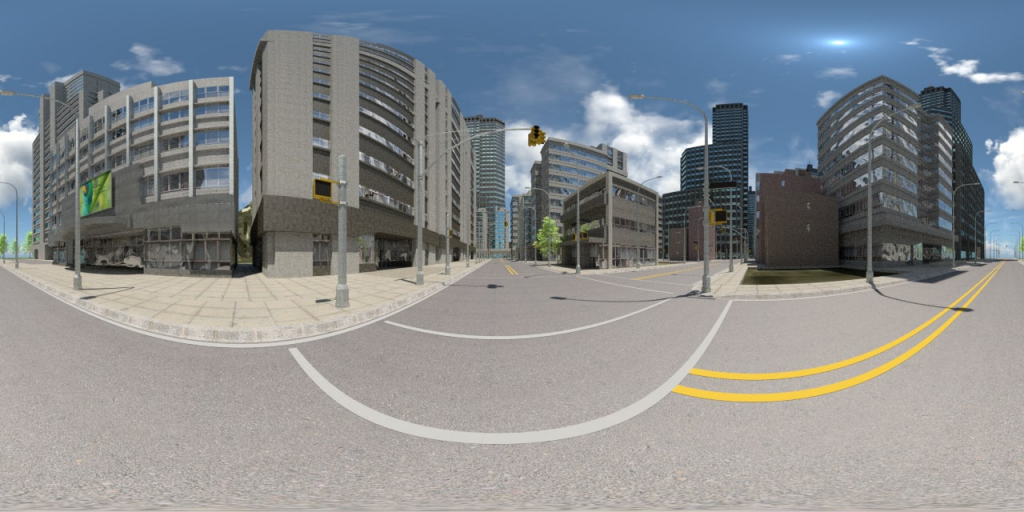}\,
    \includegraphics[width=0.24\textwidth]{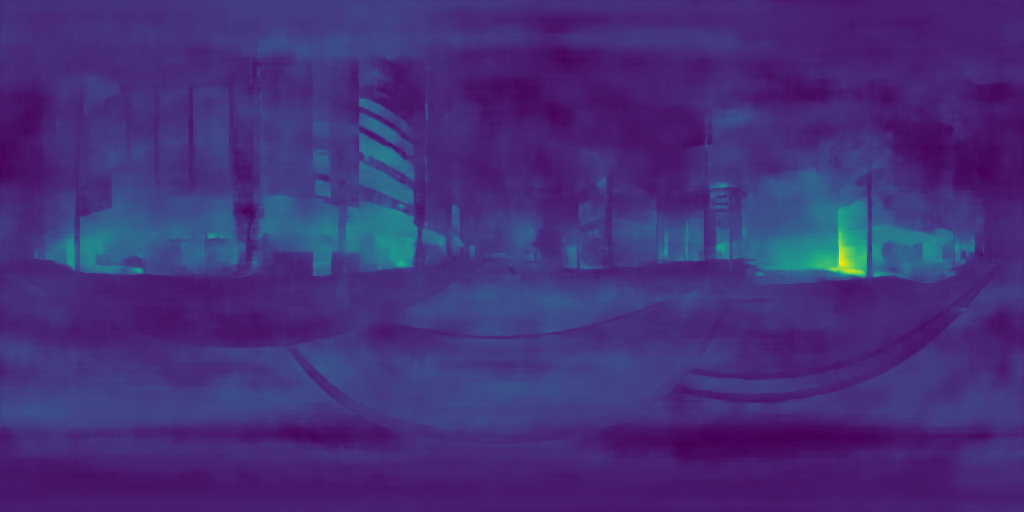}\,
    \includegraphics[width=0.24\textwidth]{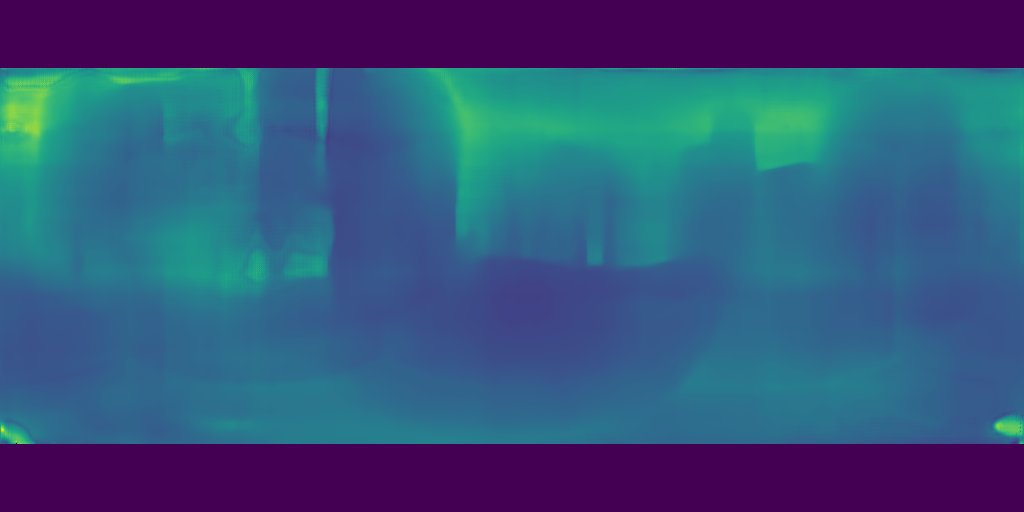}\,
    \includegraphics[width=0.24\textwidth]{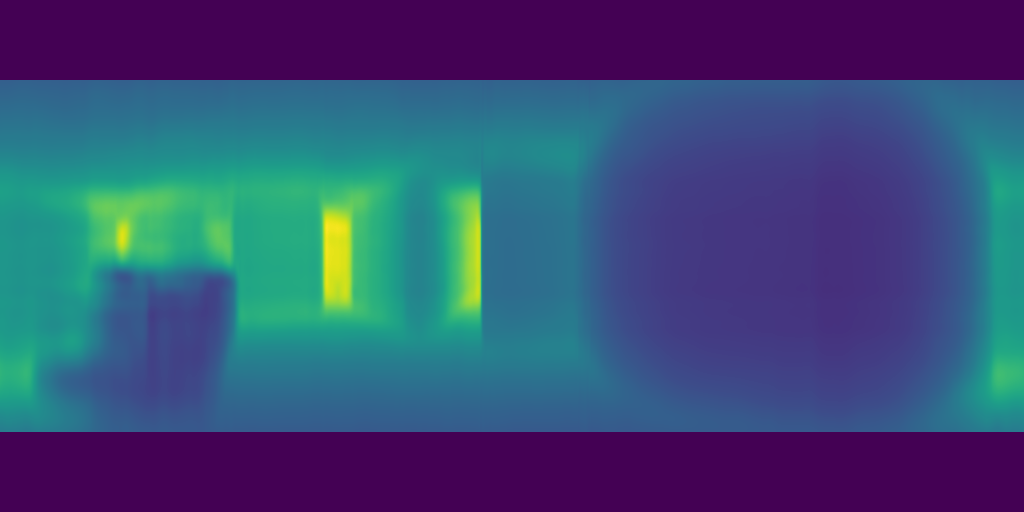}\\[.0555in]

    \includegraphics[width=0.24\textwidth]{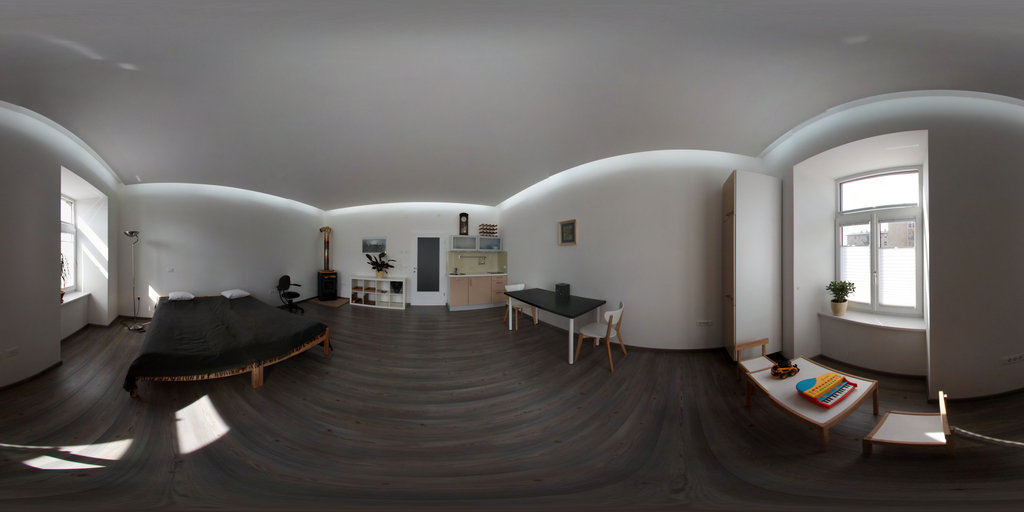}\,
        \includegraphics[width=.24\textwidth]{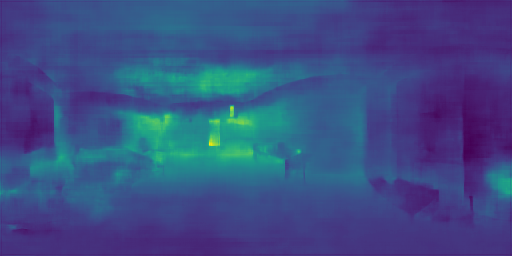}\,
        \includegraphics[width=.24\textwidth]{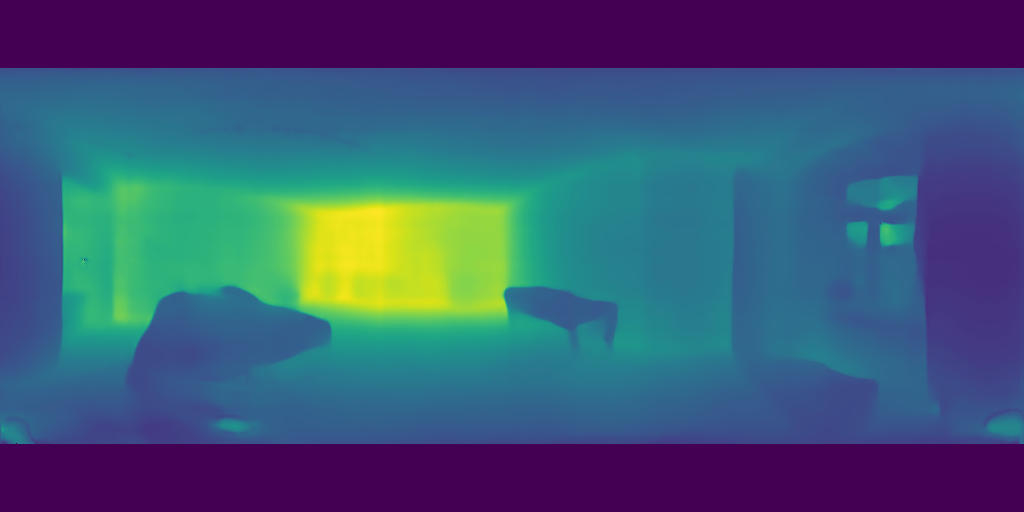}\,
        \includegraphics[width=.24\textwidth]{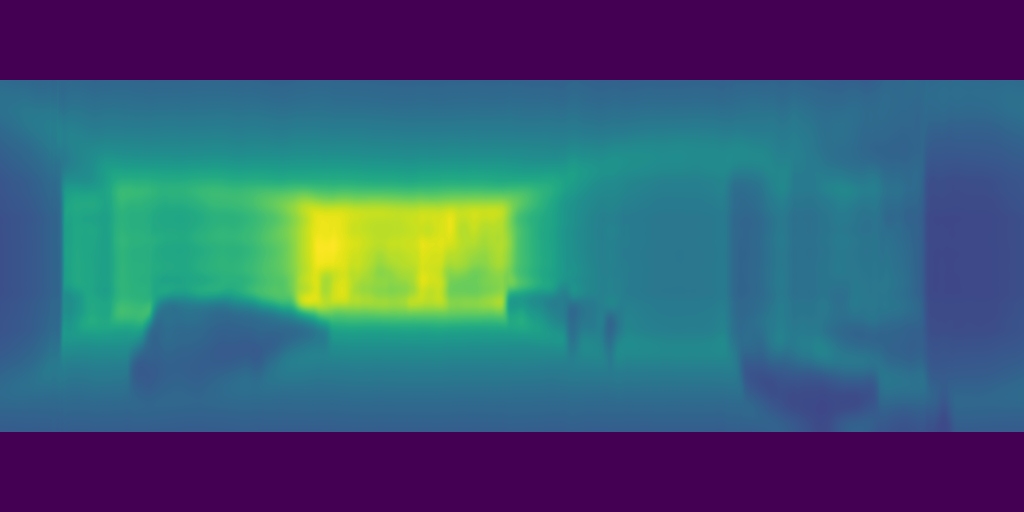}\\[.055in]
    \includegraphics[width=.24\textwidth]{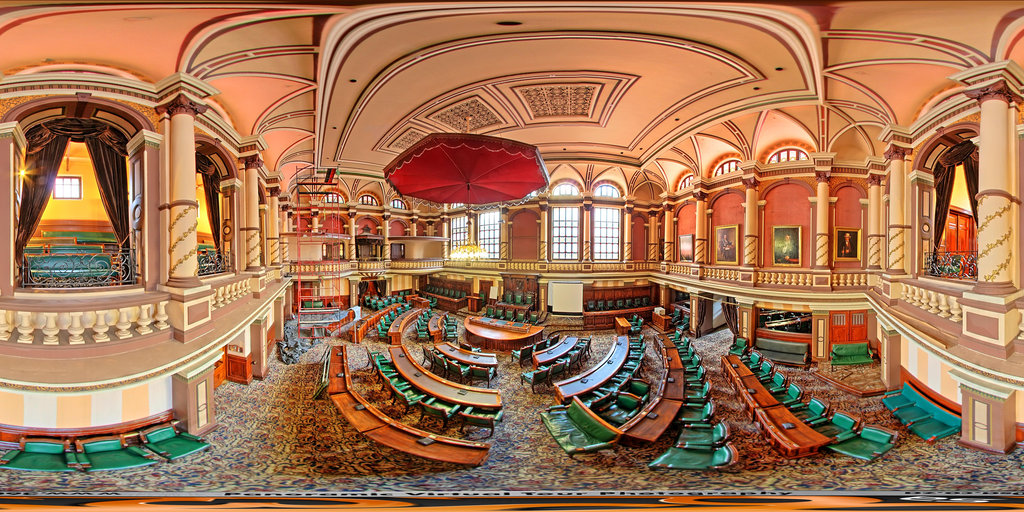}\,
        \includegraphics[width=.24\textwidth]{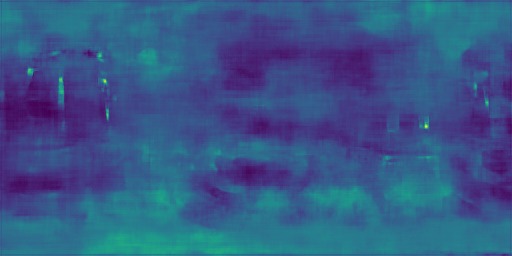}\,
        \includegraphics[width=.24\textwidth]{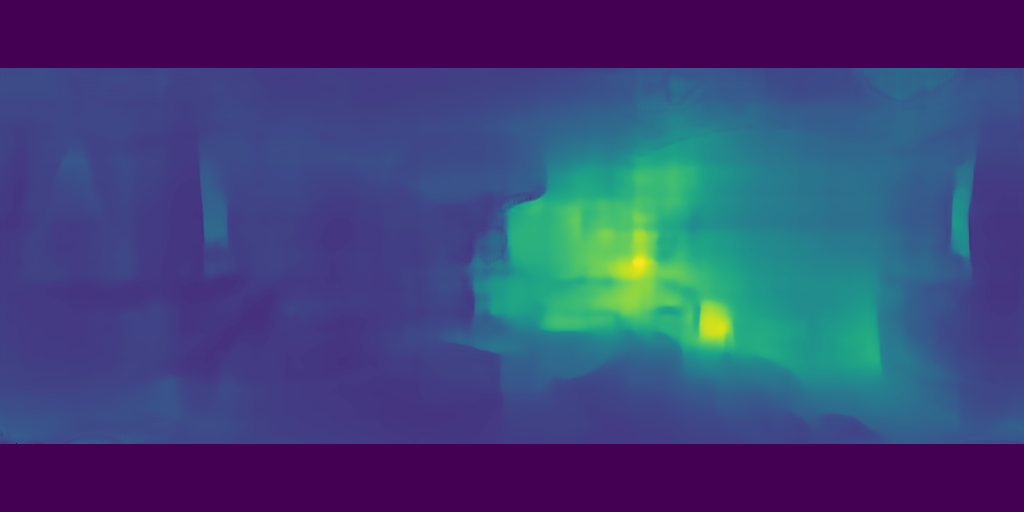}\,
        \includegraphics[width=.24\textwidth]{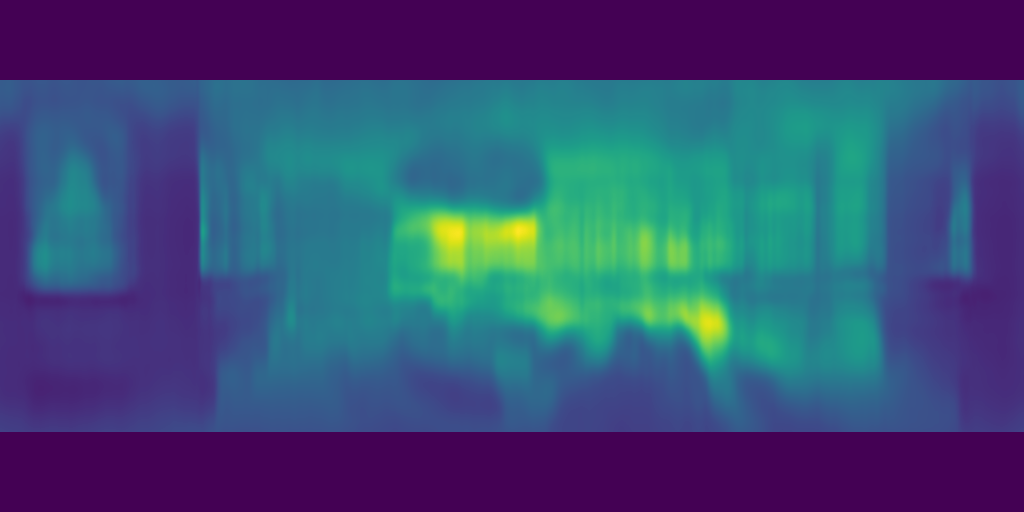}\\[.055in]
    \includegraphics[width=.24\textwidth]{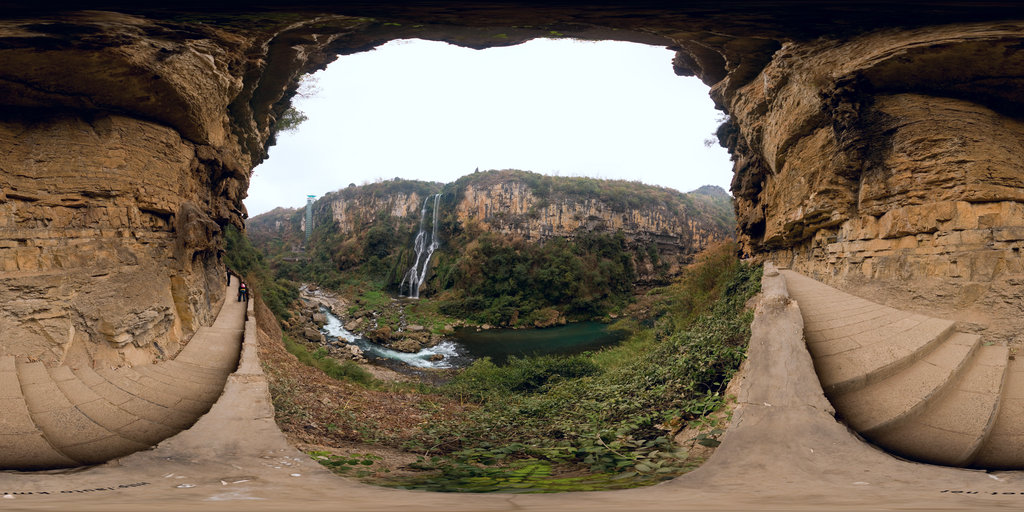}\,
        \includegraphics[width=.24\textwidth]{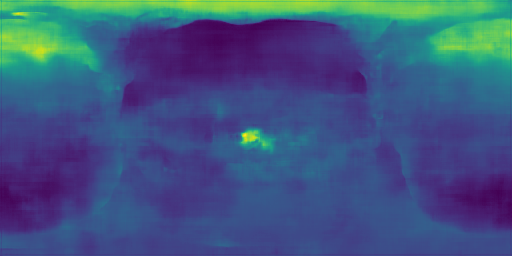}\,
        \includegraphics[width=.24\textwidth]{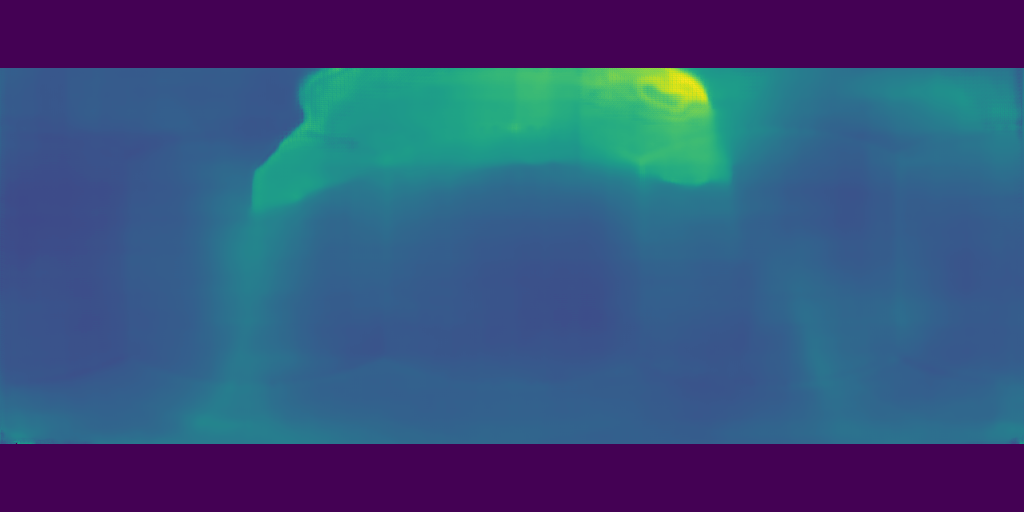}\,
        \includegraphics[width=.24\textwidth]{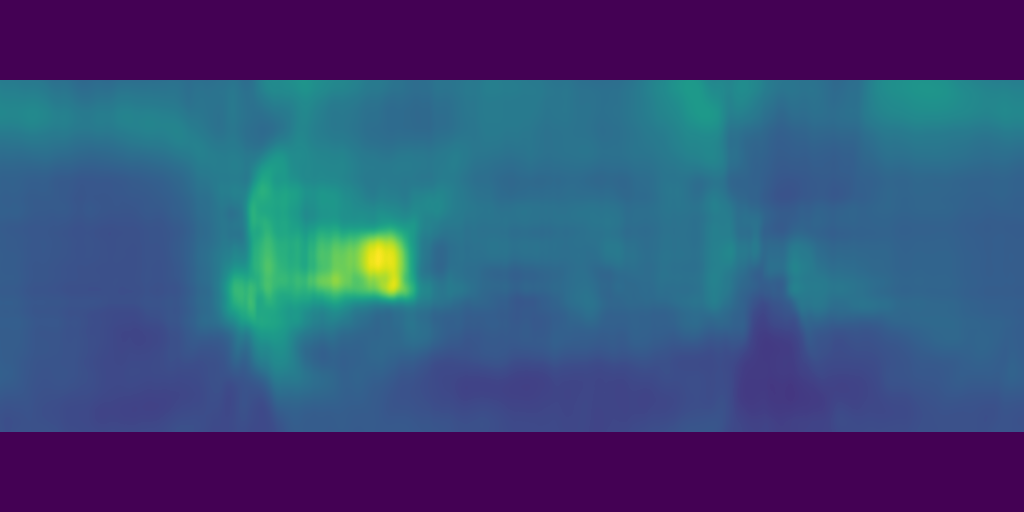}\\[.055in]
    \includegraphics[width=.24\textwidth]{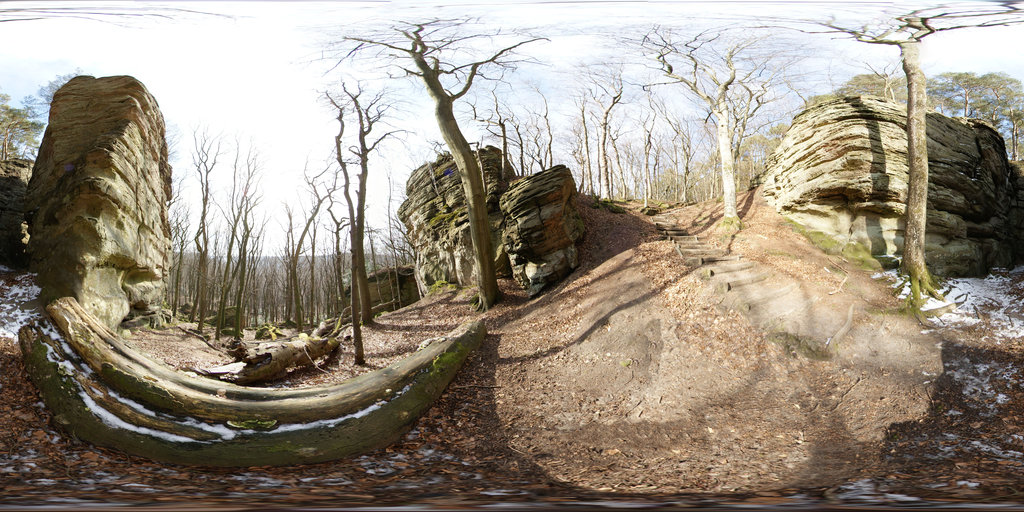}\,
        \includegraphics[width=.24\textwidth]{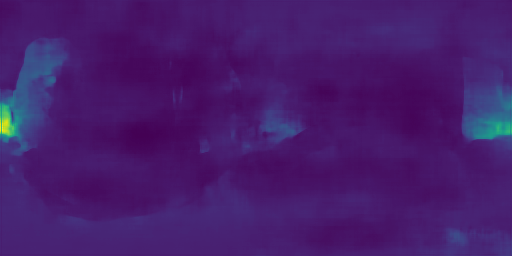}\,
        \includegraphics[width=.24\textwidth]{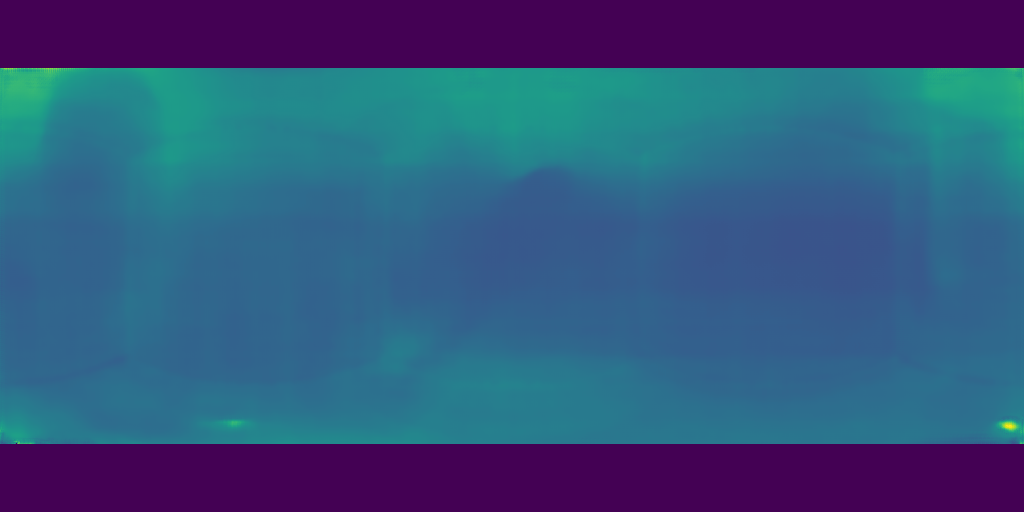}\,
        \includegraphics[width=.24\textwidth]{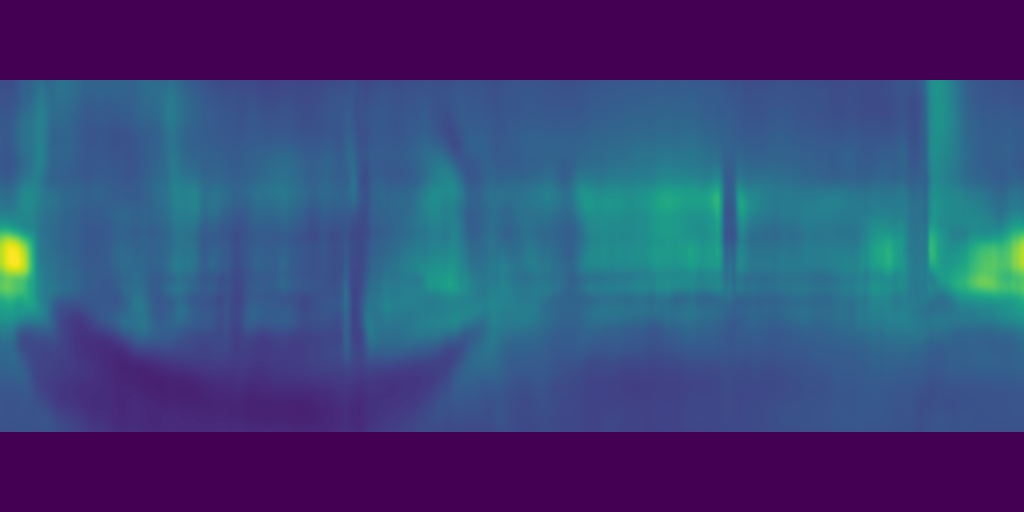}\\[.055in]
    \includegraphics[width=.24\textwidth]{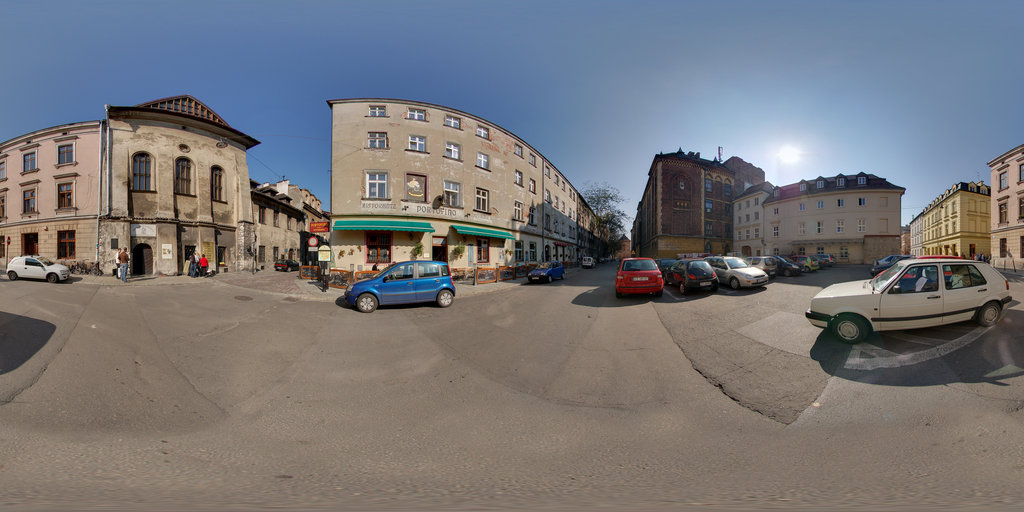}\,
        \includegraphics[width=.24\textwidth]{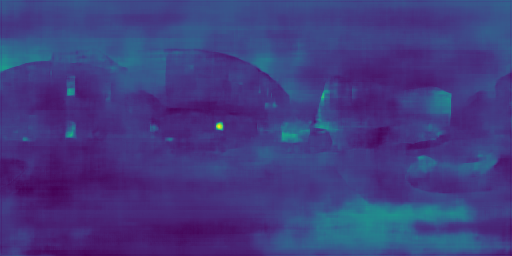}\,
        \includegraphics[width=.24\textwidth]{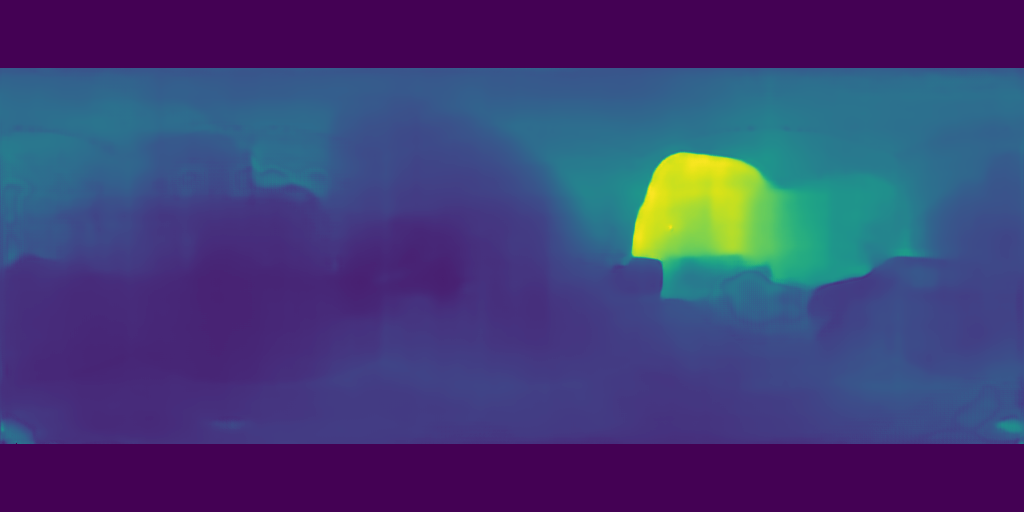}\,        \includegraphics[width=.24\textwidth]{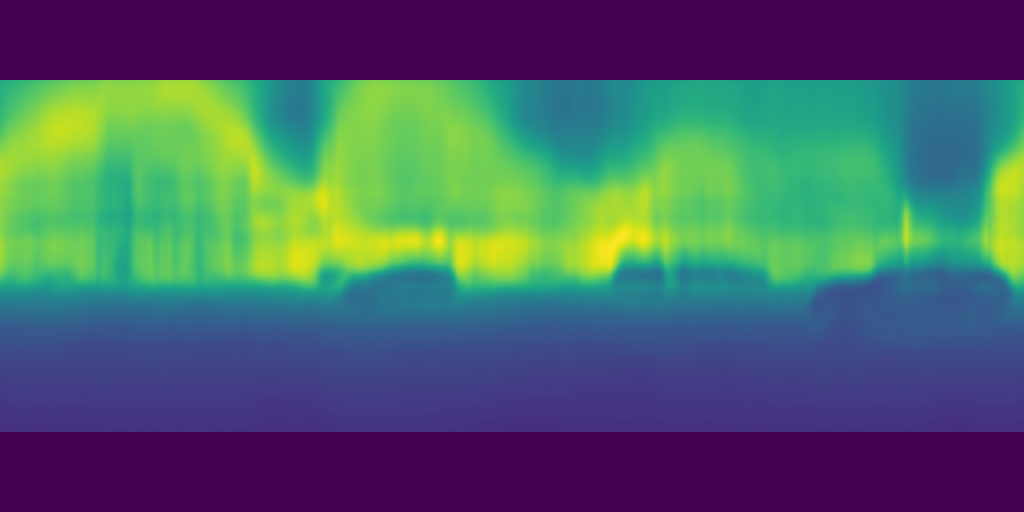}\\[-.085in]
    \subfloat[]{\includegraphics[width=0.24\textwidth]{input.png}}\,
    \subfloat[]{\includegraphics[width=0.24\textwidth]{omni_depth_resul.png}}\,
    \subfloat[]{\includegraphics[width=0.24\textwidth]{bi_fuse_result.png}}\,
    \subfloat[]{\includegraphics[width=0.24\textwidth]{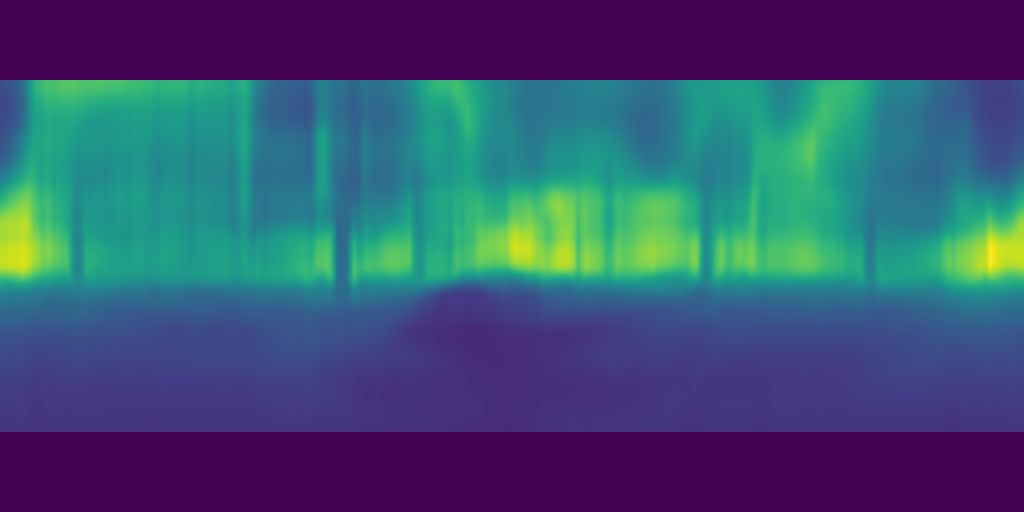}}
    \caption{
     \tlts{Examples of monocular depth estimation. 
     (a) input color panorama; (b) OmniDepth~\cite{Zioulis:ECCV:2018}, (c) BiFuse~\cite{Wang:CVPR:2020},
     and (d) HoHoNet~\cite{sun:arxiv:2020} depth estimates.
     The first and last color images are available in the 3D60~\cite{Zioulis:ECCV:2018} and Urban Canyon~\cite{Zhang:ICRA:2016} datasets, respectively. The remaining color images were obtained from the SUN360 dataset~\cite{Xiao:CVPR:2012}.
     Dark blue and yellowish colors represent closer and farther objects, respectively.
     Depth values are \emph{not} in the same scale.
     BiFuse and HoHoNet do not estimate depth at the image poles. }
    }
    \label{fig:real_depth}
\end{figure}

\tlts{
Fig.~\ref{fig:real_depth} shows depth maps produced by three state-of-the-art monocular depth estimation methods~\cite{Zioulis:ECCV:2018,Wang:CVPR:2020,sun:arxiv:2020} applied to panoramas captured in different scenarios.
The selected images illustrate common and ``unusual'' indoor environments, as well as natural and urban outdoor scenes. Both real and synthetic captures are considered. The color image in the first row is a synthetic re-rendering of a 3D model based on real captures, available in the test split of the 3D60~\cite{Zioulis:ECCV:2018} dataset. 
The color images in rows two to six are real captures taken from the SUN360 dataset~\cite{Xiao:CVPR:2012}, and the last color image is a capture from a synthetic model, available at the Urban Canyon~\cite{Zhang:ICRA:2016} dataset.
It is interesting to note that the depth maps produced by the 
approaches are 
consistent in the common indoor scenarios (first and second rows), but perform poorly in the other scene contexts. This behavior is highly related to the training datasets used in~\cite{Zioulis:ECCV:2018,Wang:CVPR:2020,sun:arxiv:2020}, which consist mainly of indoor panoramas with common decoration objects and furniture. 
Outdoor scenarios are less controlled, since the range of depth values is potentially larger, and  may present pixels with an ``infinite'' depth (often misrepresented in ground truth depth maps of synthetic outdoor scenarios).
As a result, only few works consider tackling this problem~\cite{Tateno:ECCV:2018,Payen:ECCV:2018} -- which may obligate changing the methodology design and parametrization}.


\tlts{It is also important to mention that depth maps produced by single-panorama approaches present scale ambiguity.
This problem is implicitly taken into account when all scenes present approximately the same dimensions (as in panoramas taken inside the rooms of a house, for example), but can also be explicitly handled if additional information is provided (\eg, the height of the camera as in~\cite{Pintore:WACV:2016,wang:arxiv:2021}). In terms of quantitative evaluation, some metrics handle the scale ambiguity problem by performing some kind of depth adjustment in the comparison, as discussed in Section~\ref{sec:evaluation:depth}.}

\section{Stereoscopic 3D Geometry Estimation} \label{sec:stereo}

Obtaining 3D information from a pair of images is inherent to human beings, who interpret disparities seen by both eyes as depth. 
The literature on stereo matching for perspective images is extensive, but the amount of papers that explore stereo panorama pairs is much smaller. On the one hand, there is a strong information overlap between the information captured by the two panoramas, unlike the scenario with narrow-FoV cameras. 
On the other hand, finding correspondences between two panoramas is harder due to the 
\tlts{non-affine}
distortions introduced by the spherical projection.
This section starts revising sparse 
and dense feature matching methods
tailored to the spherical domain.
Then,
it describes existing approaches for stereoscopic depth estimation using panoramas.

\subsection{Sparse Feature Extraction and Matching}  \label{subsec:sparsefeatures}


A key problem related to pose estimation and 3D reconstruction is robustly finding correspondence points across two or more views. 
For that, it is important to find interest points with a well-defined position in the image space that presents stability under image perturbations such as illumination or brightness variations. 
It is also important to obtain a feature descriptor invariant under various image transformations to perform feature matching among multiple captures of the same scene.

There are many well-known techniques for \emph{planar} keypoint finding and/or description, such as the Harris corner detector \cite{Harris:BMVA:1988}, 
Scale-Invariant Feature Transform (SIFT) \cite{Lowe:IJCV:2004}, Affine SIFT (A-SIFT)~\cite{Morel:JIS:2009}, Speeded Up Robust Features (SURF) \cite{Bay:ECCV:2006}, Features from Accelerated Segment Test (FAST) \cite{Rosten:EECV:2006}, Binary Robust Independent Elementary Features (BRIEF) \cite{calonder:pami:2011}, 
Binary Robust Invariant Scalable Keypoints (\tlts{BRISK}) \cite{Leutenegger:ICCV:2011}, Oriented FAST and Rotated BRIEF (ORB) \cite{Rublee:ICCV:2011}, 
KAZE~\cite{Alcantarilla:ECCV:2012}, Accelerated KAZE (A-KAZE)~\cite{Alcantarilla:BMVC:2013}, among others -- a recent survey on planar keypoint matching can be found in  \cite{Leng:Access:2018}.

Some works~\cite{Pathak:ICCAS:2016, Pagani:ICCV:2011, Pagani:VAST:2011} apply planar keypoint algorithms
to ERP images of panoramas, which is not ideal since they cannot handle the intrinsic distortions of the spherical camera model~\cite{Silveira:SIBGRAPI:2017}. Hence, there has been an increasing interest in keypoint extraction and matching  tailored to the spherical domain in the past decade. Cruz-Mota \etal \cite{Cruz-Mota:IJCV:2012} adapts SIFT 
to spherical coordinates
(S-SIFT), where the spherical difference of Gaussians is defined. The authors propose to compute both local spherical descriptors and local planar descriptors, so that perspective and spherical images can be matched. To avoid computing the costly spherical Fourier transform, some authors \cite{Guan:CVPR:2017, Zhao:IJCV:2014,Guan:ISVC:2013,Kitamura:ICMA:2015} approximate the sphere by a geodesic grid, where a quasi-hexagonal parameterization can be obtained, which allows the adaptation of intensity-based methods. Inspired on the ORB algorithm,
Zhao \etal~\cite{Zhao:IJCV:2014} propose 
SPHORB, which explores spherical FAST \cite{Guan:ISVC:2013} and adapts 
BRIEF to the spherical domain using the geodesic grid. In their approach, the grid is stored in distinct a set of triangles with no direct connections, which leads to distortions and special cases on the boundaries of each set.  On the other hand, Guan \etal~\cite{Guan:CVPR:2017} use a half-edge data structure to represent the geodesic grid. Then, they use the spherical FAST algorithm as a keypoint detector, and adapt BRISK to the spherical domain, which is named BRISKS. 
According to~\cite{Zhao:IJCV:2014,Guan:CVPR:2017}, these methods outperforms S-SIFT \cite{Cruz-Mota:IJCV:2012} in 
robustness and repeatability validation metrics under rotation. Further comparison of most previous methods can be found in \cite{Silveira:SIBGRAPI:2017}.

All techniques described so far adapt hand-crafted keypoint algorithms to work on the sphere. However, \tlts{solutions based on deep learning} for keypoint matching in perspective images have shown promising results (such as~\cite{DeTone:CVPR:2018}). 
\tlts{We are not aware} of 
\tlts{deep learning}-based keypoint extractor/descriptor for panoramas, but one can embed planar 
\tlts{deep learning}-based approaches through TPP projections of 360\degree imagery, as recently shown in~\cite{Eder:CVPR:2020} for planar hand-crafted descriptors.

\begin{figure}[!t]
\begin{minipage}[b]{1.0\linewidth}
  \centering
  \subfloat[]{\includegraphics[width=.24\textwidth]{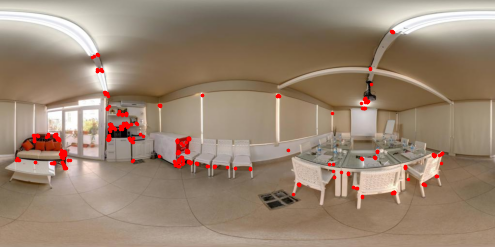}}\,
  \subfloat[]{\includegraphics[width=.24\textwidth]{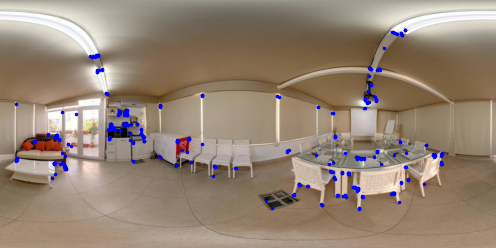}}\,
  \subfloat[]{\includegraphics[width=.24\textwidth]{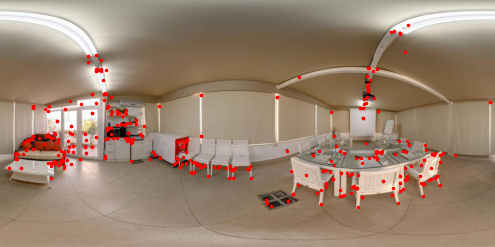}}\,
  \subfloat[]{\includegraphics[width=.24\textwidth]{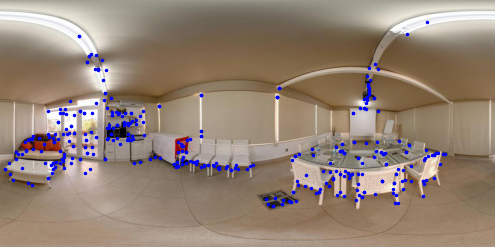}}
\end{minipage}
\caption{ Examples of keypoint detection using planar methods and \tlts{spherical} adaptations. \tlts{The input panorama, taken from the SUN360 dataset~\cite{Xiao:CVPR:2012}, is processed by  (a) ORB,  (b) SPHORB,  (c) SIFT and its (d) spherical adaptation using tangent planes~\cite{Eder:CVPR:2020}}.
} 
\label{fig:compare_keypoints}
\vspace{-0.25cm}
\end{figure}

\tlts{ For the sake of illustration, Fig.~\ref{fig:compare_keypoints} shows a comparison of planar keypoint extraction methods and different spherical adaptations.
More precisely, Figs.~\ref{fig:compare_keypoints}(a)-(b) show the keypoints detected with both ORB and SPHORB,
and Figs.~\ref{fig:compare_keypoints}(c)-(d) show keypoints obtained by planar SIFT and its spherical adaptation using tangent planes~\cite{Eder:CVPR:2020}. }
\tlts{Although it is hard to see the differences in a single image, \emph{planar} keypoint matching algorithms suffer from poor repeatability, precision, and recall metrics when applied to spherical image pairs that differ by translation and/or rotation~\cite{Guan:CVPR:2017,Zhao:IJCV:2014,Cruz-Mota:IJCV:2012}. 
It means that such standard methods cannot handle the spherical distortions present in wide- or full-FoV images such as CP or ERP. 
Keypoint matching tends to be more effective in CMP representation, but the keypoints found in each of the six cube faces of an image still need to be matched to those found in the six cube faces of another image. Some works consider padding the cube faces for handling discontinuities in correspondences tracking in CMP videos~\cite{Huang:VR:2017}.
As planar methods may not work well with 360$^\circ$ images, they can compromise other tasks such as pose and depth estimation.}

\subsection{Dense Feature Extraction and Matching} \label{subsec:densefeatures}

\begin{figure}
    \centering
    \includegraphics[height=.1325\textwidth]{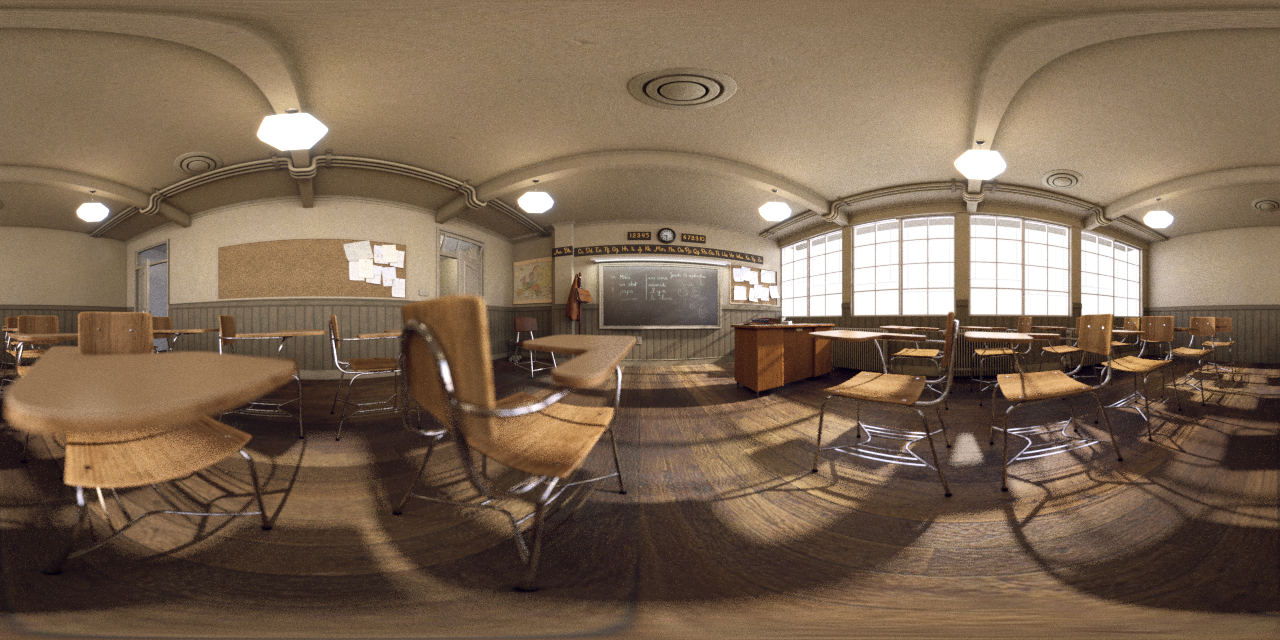}\,
    \includegraphics[height=.1325\textwidth]{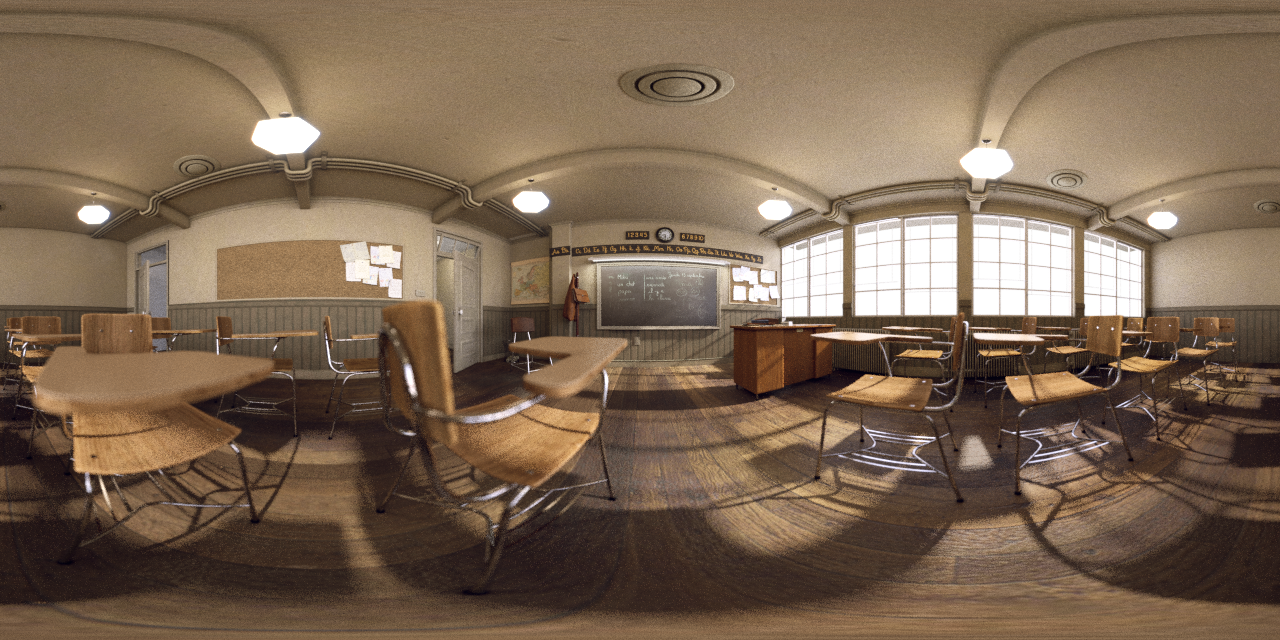}\,
    \includegraphics[height=.1325\textwidth]{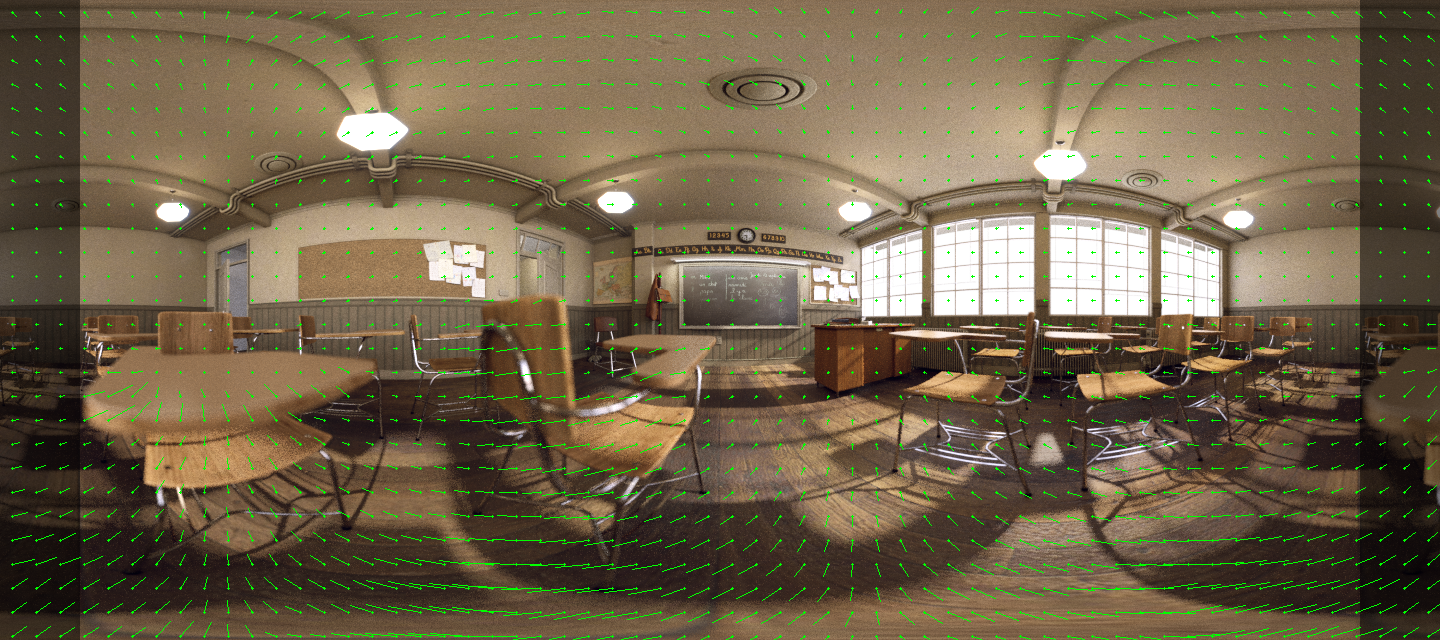}
    \caption{Input image pair and the underlying flow  estimates (shaded areas highlight the circular padding) computed by DeepFlow~\cite{Silveira:CVPR:2019}. The stereo images differ by translation along the three axes. There is no relative rotation between the captures. 
    The ``Classroom model'' is available 
    at \url{https://www.blender.org}.
    }
    \label{fig:deepflow}
    \vspace{-0.5cm}
\end{figure}

Optical flow 
captures the
apparent motion of individual pixels using 
pairs of still images or video sequences,
and often assume small motion and brightness constancy.
In particular, dense motion maps are useful for obtaining \emph{dense}
correspondences from a pair of images.
There are many 
\tlts{optical flow}
algorithms designed to work with perspective imagery (refer to~\cite{zhai:pr:2021} for a recent survey).
Similarly to sparse feature matching, some works~\cite{Silveira:VR:2019,Xu:ROBIO:2016,Pathak:ICCAS:2016} adopt traditional 
\tlts{optical flow}
approaches for 360\degree image processing, sometimes in between pre- and post-processing steps. The most straightforward workaround found in literature is to circularly pad ERP images before computing the 
\tlts{optical flow}, avoiding
longitudinal disconnection~\cite{Silveira:VR:2019,Xu:ROBIO:2016} (recall that a similar strategy was adopted in the context of 
\tlts{deep learning}
for depth or layout estimation, as mentioned in Section~\ref{sec:monocular}).
Fig.~\ref{fig:deepflow} illustrates this strategy for the popular planar
\textit{DeepFlow} algorithm~\cite{Weinzaepfel:ICCV:2013} as baseline 
\tlts{optical flow}, used in~\cite{Silveira:VR:2019} to obtain dense correspondences for 3D reconstruction.

Mochizuki \etal \cite{Mochizuki:SIMPAR:2008} adapts the classical Horn-Schunck algorithm \cite{Horn:AI:1981} to the spherical domain and 
applies it to omnidirectional image-based robot navigation. 
Inspired on \cite{Conklin:ICA:1997}, the approaches presented in~\cite{Tosic:ESPC:2005, Mochizuki:CAIP:2011} perform a multi-resolution decomposition 
of the sphere to speed-up the computation of the adapted Horn-Schunck algorithm, estimating motion on the lowest resolution and propagating it to higher resolutions. 
These approaches, however, 
tend to fail when high-frequency components of the image are noisy.
The strategies presented in~\cite{radgui:CVIU:2011,Alibouch:SIVP:2014} use a multi-channel 
approach that performs
convolution with spherical wavelets, achieving a good compromise between computational cost and robustness.
In another line of work, Bagnato \etal \cite{Bagnato:ICIP:2009}, extend the Total Variation L1-Regularized 
\cite{Zach:PR:2007}  approach for real-time 
\tlts{optical flow}.
However, instead of working on the sphere, they design a graph-based framework (where the vertices represent image pixels and edges define connections between pixels weighted by the geodesic distance) and define the differential operators over these discretization formulas. 
Kim and colleagues \cite{Kim:ICASSP:2019} normalize the flow patterns by equalizing 
\tlts{optical flow}
results of pairs of images rotated by 90 degrees on different axes, and then test it in a downstream task (rotation estimation), obtaining improved results when compared to planar methods.

Other authors focus on the adaptation of 
\tlts{deep learning}-based 
\tlts{optical flow}
algorithms originally designed for planar images to the spherical domain (recall the discussion in Section~\ref{subsec:spherenn}). 
Xie \etal \cite{Xie:3DV:2019} integrate 
DNN layers for spherical adaptation of existing 
\tlts{optical flow}
architectures \cite{Sun:CVPR:2018, Dosovitskiy:ICCV:2015,ranjan:CVPR:2016}, testing three special layer neural networks: correlation convolution \cite{Dosovitskiy:ICCV:2015}, coordinate convolutions \cite{liu:nips:2018} and deformable convolution \cite{Dai:ICCV:2017}. They report that these layers minimize the error on boundaries  on their dataset, named FlowCLEVR dataset. In the same direction,
%
Artizzu \etal \cite{artizzu:ICPR:2021} adapts the \textit{FlowNet} algorithm \cite{Dosovitskiy:ICCV:2015} to the spherical domain implementing a distortion-aware convolution kernel that reuses planar learned weights and architecture. 
Spherical data is not required for the training procedure.
Bhandari \etal \cite{bhandari:Arxiv:2020} adapts the \textit{LiteFlowNet} algorithm \cite{HUI:CVPR:2018} to the spherical domain by modifying the kernels size depending on the latitude angle of ERP images, and
a refinement process that involves a correction factor. These 
\tlts{deep learning}-based 
\tlts{optical flow} 
methods outperform their planar counterparts, but comparisons with other spherical 
\tlts{optical flow}
estimators are still scarce.



\subsection{Depth Estimation}

\begin{figure}
    \centering
    \subfloat[]{\includegraphics[height=.1325\textwidth]{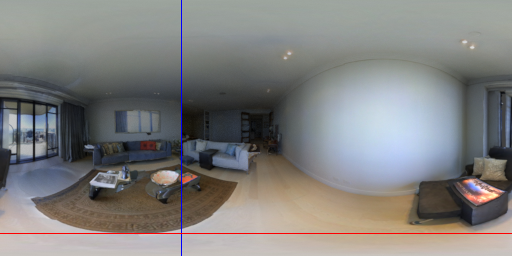}}\,
    \subfloat[]{\includegraphics[height=.1325\textwidth]{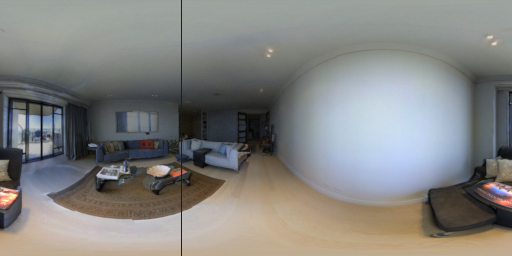}}\,
    \subfloat[]{\includegraphics[height=.1325\textwidth]{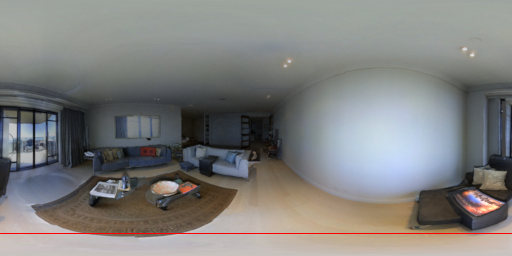}}
    \caption{Stereoscopic 360\degree image pairs. A (a) reference image  and two stereo views with (b) horizontal (left to right) and (c) vertical (down to up) baselines (0.26m in both directions). Colored guidelines are drawn for better visualization of the offsets. Source image taken from the 3D60 dataset~\cite{Zioulis:ECCV:2018}.}
    \label{fig:baselinestereo}
\end{figure}

Sparse and dense feature matching, revised in Sections~\ref{subsec:sparsefeatures} and \ref{subsec:densefeatures}, 
are often employed in stereo- and multi-view-based methods for 3D geometry recovery. Stereo image pairs are captured
under arbitrary translations
(as illustrated in Fig.~\ref{fig:deepflow}) or using fixed baselines 
-- typically horizontal or vertical (as illustrated in Fig.~\ref{fig:baselinestereo}). Furthermore,  some approaches disregard the rotation component (which might be guaranteed by the capture process using camera rigs or undone when more flexible setups are employed).
\tlts{Most works do not assume specifically indoor or outdoor image pairs as input.
Note, however, that the intermediate step of feature matching tends to be more challenging outdoors simply because some objects tend to be too far -- occupying only a few pixels.
Minimal disparities or triangulations resulting in 3D points too far from the virtual camera can be filtered in a postprocessing step. These procedures, however, are barely stated in most works. }

This section discusses 
stereo matching approaches based on spherical image pairs only, and Table~\ref{tab:stereomethods} presents an overview of the reviewed techniques. 
They vary in terms of image representation, employed technology for feature matching and depth regression, and the density of the resulting depth (or disparity or 3D) estimates. Sparse estimates are often associated with sparse feature matching, which might be densified by extrapolating the sparse field. Intrinsically dense methods, on the other hand, provide one depth (or disparity) value for each pixel of the spherical image. 

\begin{table}
\setlength{\tabcolsep}{0.25em} 
\centering
\caption{3D geometry estimation methods based on a pair of spherical images.}
\label{tab:stereomethods}
\footnotesize
\begin{tabulary}{\linewidth}{lcccc}
\hline
\textbf{Reference} & \textbf{Representation} &  \textbf{Technology} & \textbf{Matching} & \textbf{Density}  \\
\hline 
Li and Fukumori~\cite{Li:VR:2005} & HP ($\times4$) & Triangulation & Keypoints & Sparse \\
Arican and Frossard~\cite{Arican:AVSS:2007} & ERP & Energy minimization & Pixel-wise & Dense \\
Kim and Hilton~\cite{Kim:ICCV:2009} & ERP & PDE-based optimization & Block matching & Dense \\
Pathak~\etal~\cite{Pathak:ICCAS:2016, Pathak:JCMSI:2017}  &  ERP & Linear/Non-linear optimization & Keypoints/\tlts{Optical flow} & Dense \\ 
Pathak~\etal~\cite{Pathak:IST:2016, Pathak:SII:2017}  &  ERP & Linear/Non-linear optimization & Keypoints/\tlts{Optical flow} & Dense \\ 
Kim \etal~\cite{Kim:3DV:2016} & CMP & PDE-based optimization & Block matching & Semi-dense \\
Kim~\etal~\cite{Kim:3DV:2017} &  CMP& PDE-based optimization & Block matching & Semi-dense \\
Wegner \etal \cite{Wegner:ICIP:2018} & CP & Energy minimization & Block matching & Dense \\
Lai \etal \cite{Lai:VR:2019} & ERP & 
DNN
& Learned features & Dense \\
Wang~\etal~\cite{Wang:arXiv:2019} & ERP & 
SDNN
& Learned features & Dense \\
Roxas and Oishi~\cite{roxas:iral:2020} & HP ($\times1$) & Energy minimization & Pixel-wise & Dense \\
\hline
\end{tabulary}
\end{table}

Li and Fukumori~\cite{Li:VR:2005} were one of the pioneers in stereo-panorama depth estimation. They actually used  two pairs of wide-FoV/HP images as source for estimating 3D geometry of scenes, and adapted the main concepts involved in the traditional stereo matching problem to the spherical domain (in fact, they formulated most of what is reviewed in Section~\ref{ssec:sphimage} of this survey).
%
%
They obtain correspondences in spherical images using
the Lucas-Kanade method~\cite{Lucas:IJCAI:1981} and enforcing epipolar constraints. 
Finally, the authors 
project planar texture patches extracted by
a Delanuay triangulation~\cite{Shewchuk:COMGEO:2014} of the sparse point cloud for visualization purposes.

Instead of estimating 3D information based on keypoints, Kim and Hilton~\cite{Kim:ICCV:2009} introduce a stereo-based  algorithm for \emph{dense}  360\degree disparity estimation.
%
%
ERP images are created via cross-slits from rotating line-scan cameras with wide-FoV lenses. Then,
extrinsic calibration is performed via feature matching. The authors arrange the stereoscopic cameras in a top-bottom configuration so that their epipolar lines are vertically aligned (confer Fig.~\ref{fig:baselinestereo}), and introduce a hierarchical partial differential equation (PDE)-based algorithm to estimate continuous disparity maps using the mean-absolute-error/normalized-cross-correlation 
\cite{Radke:Book:2012} block matching techniques. %
Finally, 3D information is estimated via triangulation, and a mesh is built for visualization purposes.

Some authors 
make 
the image capture process more flexible.
Arican and Frossard~\cite{Arican:AVSS:2007} introduce a
method for estimating dense disparities from
stereoscopic omnidirectional images
in an uncontrolled small baseline.
For that aim, they first to rectify both images, so that the epipolar great circles coincide with the longitudes.
They formulate a global energy optimization scheme founded in graph-cuts~\cite{Kolmogorov:ECCV:2002} that searches for optimal disparities from a discrete set of possibilities.
The authors conclude their paper by presenting hints on how additional cameras can further improve the results.
Pathak and colleagues~\cite{Pathak:ICCAS:2016,Pathak:JCMSI:2017} also deal with a setup with  small baseline and arbitrary rotation. They first 
\tlts{estimate}
the 5-DoF pose between the two images using A-KAZE 
features and RANSAC-enabled 8-PA~\cite{Hartley:TPAMI:1997}, extract the rotation matrix and ``derotate'' the images.
Then, the authors compute dense correspondences
using DeepFlow and minimize the moment of the 3D magnitude-normalized flow in an iterative, non-linear manner. When the pose estimate converges, dense 3D information is computed via triangulation. In a follow-up work, Pathak \etal~\cite{Pathak:IST:2016} opt to process the two images directly on the ERP format, instead of working on the sphere as \cite{Pathak:ICCAS:2016, Pathak:JCMSI:2017}.
According to the authors,
this choice helps 
\tlts{avoid}
issues caused by numerical error. The result of the pipeline remains the same as in~\cite{Pathak:ICCAS:2016,Pathak:JCMSI:2017}.
This methodology was also published in \cite{Pathak:SII:2017}.

\tlts{Contrary to most contemporary approaches}, 
Wegner \etal \cite{Wegner:ICIP:2018} 
work with rotation-free stereo-rectified omnidirectional image pairs represented in a CP format.
The authors claim that an approximate formula can be efficiently embedded into existing planar block-matching-based methods for depth inference to be applied to panoramic images. As a proof of concept, they show qualitative results after adapting the cost function of the ISO/IEC MPEG Depth Estimation Reference Software
\cite{Wegner:ISOIEC:2015} and submitting a pair of stereoscopic images.
The disparity/depth estimates on the image poles are poor, and there are apparent seams on both lateral boundaries.

Roxas and Oishi~\cite{roxas:iral:2020} introduce a  variational approach for estimating depth from stereoscopic cameras equipped with fish-eye lenses, which does not apply any kind of undistortion or correction and can achieve real-time performance on modern GPUs. The authors propose to use the trajectory field of each pixel involving the baseline, handling multiple correspondences displacement by integrating a coarse-to-fine approach. This strategy constrains the search for matches along the epipolar curves, in a fast way, using linear interpolation. 
Although their method is applied to HP images, their formulation also applies to any central camera projection.

Recently, some works~\cite{Lai:VR:2019,Wang:arXiv:2019} started tackling the 360\degree stereo-based depth estimation problem from a learning-based perspective. 
Wang and colleagues~\cite{Wang:arXiv:2019} propose an end-to-end 
\tlts{deep learning}
model for regressing disparity from top-bottom stereo 360\degree images (depth 
is computed from the disparity maps).
The method comprises a two-branch feature extractor that concatenates the stereo pair and polar angle; an À Trous-Spatial Pyramid Pooling;  
a learnable cost volume that accounts for the non-linear projection; and
an stacked hourglass module for finally estimating the disparity map. The authors claim that using
polar angle allows decoupling image appearance from distortion, and show its importance on their ablation study. 
%
%
%
Lai \etal \cite{Lai:VR:2019} introduce an  encoder-decoder network fed with a stereo-rectified pair of ERP images with a small fixed baseline 
and outputs both depth and normal maps. The authors
argue that performing multi-task learning, \ie also estimating the normal vectors, was paramount for improving the depth estimates. 
Their main contribution lies in a loss function that encourages associating the left and right boundary information from ERP images. Unlike some geometric-based solutions~\cite{Pathak:ICCAS:2016, Pathak:JCMSI:2017, Pathak:SII:2017, Pathak:IST:2016, Arican:AVSS:2007}, the learning-based approaches described in~\cite{Lai:VR:2019,Wang:arXiv:2019} are restricted to specific capture setups.

Other methods use additional information to help 
\tlts{refine}
depth estimates based on geometric constraints. Kim \etal \cite{Kim:3DV:2016} presented an approach that first produces a depth map based on a pair of vertically-spaced panoramas, and then estimate object and material information for layout recovery with furniture (which requires additional labeld data for training). In the stereo matching, they use a CMP-like projection followed by line detection with the Hough transform to estimate the camera rotations \wrt the room coordinate system, and then gravity-align the image pair. Correspondences are found using the PDE-based method from~\cite{Kim:ICCV:2009}, and depth information is obtained by triangulation (assuming that the vertical baseline is known). A CNN is used to obtain semantic information about the components, which are combined with the depth map to obtain the furnished room layout. 
In a follow-up study, Kim and colleagues~\cite{Kim:3DV:2017} fuse color image and audio responses for 3D scene geometry recovery. They modify the block world reconstruction method from~\cite{Kim:3DV:2016} to accommodate audio and visual data. Visual planes are extracted with the help of superpixels~\cite{felzenszwalb:ijcv:2004}, while
reflection planes estimated from an omnidirectional microphone array are used during the plane fitting algorithm providing conformal double-source estimates. 

\begin{figure}
    \centering
    \includegraphics[height=.1325\textwidth]{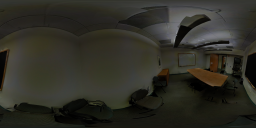}\,
    \includegraphics[height=.1325\textwidth]{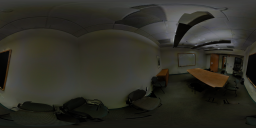}\,
    \includegraphics[height=.1325\textwidth]{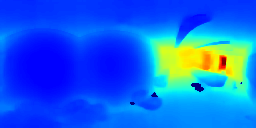}
    \caption{
    Input stereo pair and the underlying depth map estimated using the stereo approach from~\cite{Lai:VR:2019}. Source images taken from a dataset provided by the same authors in~\cite{Lai:VR:2019}.
    }
    \label{fig:stereolai2019}
    \vspace{-0.25cm}
\end{figure}

Fig.~\ref{fig:stereolai2019} depicts
a stereo 360$^\circ$ image pair with fixed horizontal baseline, and the corresponding depth map obtained with~\cite{Lai:VR:2019}. Recall that spherical images 
capture information along the epipoles -- unlike the perspective scenario. Since the epipoles indicate the relative movement direction, it is impossible to effectively estimate depth  in that particular region. 
For horizontal-aligned baselines, the epipoles appear at positions $(\theta, \phi) = (\frac{\pi}{2}, \frac{\pi}{2})$ and $(\frac{3\pi}{2}, \frac{\pi}{2})$.
Although the inferred depth map does not
present visible artifacts, the original paper~\cite{Lai:VR:2019} indicates 
missing information around the epipoles when a given 
recovered 3D scene is visualized as a point cloud.


\section{Multi-view 3D Geometry Estimation} \label{sec:multiview}

Fusing the contributions captured by many views is the core of multi-view-based 3D geometry recovery methods, being  pivotal for estimating depth from whole scenarios based on narrow-FoV pin-hole-based images.
Panoramas provide a full FoV, and although only a pair of captures is enough for geometrically obtaing depth data of the whole surrounding region (and even one capture might suffice when using learning-based approaches), using more cameras is useful for solving occlusion-related 
issues and may increase the quality of the estimates. In particular, handling occlusions and disocclusions is paramount for implementing fully immersive 6-DoF exploration for recent AR/MR/VR applications~\cite{Serrano:TVCG:2019, Silveira:VR:2019}.

This section  divides methods into three  categories according to the
required image capturing protocol.
Sections~\ref{subsec:lightfields}, \ref{subsec:mvs} and~\ref{subsec:sfm} review 3D geometry recovery techniques based on spherical light-fields (SLFs), multi-view stereo (MVS) and structure from motion (SfM), respectively. 
Table~\ref{tab:multiviewmethods} summarizes the methods reviewed
throughout this section highlighting the main category;
the input representation, following the notation used in the previous sections (we also review a method that explores a graph-based representation); the main technology behind each approach; the core idea used for feature matching; and the density of the output.
\tlts{Once again, as in the stereo case, 
most works do not make any assumptions about the scene's context, \ie indoors or outdoors.
Exceptions include works that rely on learning-based methods trained on indoor scenarios only.
}


\begin{table}[]
\setlength{\tabcolsep}{0.25em} 
\footnotesize
\centering
\caption{3D geometry estimation methods based on multiple spherical images.}
\label{tab:multiviewmethods}
\begin{tabulary}{\linewidth}{m{2.75cm}x{1.5cm}x{1.75cm}x{2.75cm}x{2.75cm}x{1.25cm}}
%
\hline
\textbf{Reference} & \textbf{Category} &  \textbf{Representation} &  \textbf{Technology} & \textbf{Matching} & \textbf{Density}  \\
\hline 
Krolla~\etal~\cite{Krolla:BMVC:2014}  & SLF & CP & Slope estimation & Direct matching & Dense \\ 
Gava~\etal~\cite{Gava:ICIP:2018}  & SLF & ERP & Energy minimization & Block matching & Dense \\

Overbeck \etal~\cite{overbeck:tog:2018} & {SLF} & {Perspective} & {Global optimization} & {Region matching} & {Dense} \\
{Bertel \etal~\cite{bertel:tvcg:2019}} & {SLF} & {Perspective} & {Camera Interpolation} & {\tlts{Optical flow}} & {Dense} \\
{Bertel \etal~\cite{bertel:tog:2020}} & {SLF} & {ERP} & {Proxy fitting} & {\tlts{Optical flow}} & {Dense} \\
Bagnato \etal~\cite{Bagnato:ICIP:2009} & MVS & Graph-based & Global optimization & \tlts{Optical flow} & Dense \\
Kim and Hilton~\cite{Kim:ICCV:2009, Kim:ECCV:2010, Kim:IJCV:2013}  & MVS & ERP & PDE-based optimization & Block matching & Dense \\ 
Kim and Hilton~\cite{Kim:IVMSPW:2013, Kim:JCVIU:2015}  & MVS & CMP & PDE-based optimization & Block matching & Semi-dense \\ 

Won~\etal~\cite{Won:ICCV:2019,won:pami:2020} & MVS &  HP ($\times4$) & CNN & Learned features & Dense \\

Torii \etal~\cite{Torii:ICCV:2009} & SfM & ERP & Non-linear optimization & Keypoint/Loop closure & Sparse \\
Micus\' {i}k and Koseck\`{a}~\cite{Micusik:CVPR:2009} & SfM & CMP - poles & Markov Random Fields & Keypoints/Superpixels & Dense \\

Barazzeti \etal~\cite{barazzetti:isprs:2017}   & SfM & ERP & (Comercial solution) & (Comercial solution) & Dense \\ 
Fangi \etal~\cite{Fangi:ISPRS:2018}   & SfM & ERP & (Comercial solution) & (Comercial solution) & Dense\\ 
Bagnato \etal~\cite{Bagnato:JMIV:2011}  & SfM & ERP & Energy minimization & \tlts{Optical flow} & Dense \\ 
Schonbein and Geiger~\cite{Schonbein:IROS:2014}  & SfM & Catadioptric ($\times4$) & Energy minimization & Keypoint/Semi-global matching & Dense\\

Caruso \etal~\cite{Caruso:IROS:2015}   & SfM & HP ($\times4$) & Non-linear optimization & Block matching & Semi-dense\\ 
Cabral and Furukawa~\cite{Cabral:CVPR:2014}  & SfM & ERP & Global optimization & Region matching  & Semi-dense\\ 
Pintore \etal~\cite{Pintore:PCCGA:2018,Pintore:CAG:2018}  & SfM & ERP & Linear/Non-linear optimization & Boundary matching & Semi-dense\\ 

Pagani and Stricker~\cite{Pagani:ICCV:2011}  & SfM  & ERP & Non-linear optimization & Keypoint/MVS  & Dense \\
Pagani~\etal~\cite{Pagani:VAST:2011}  & SfM  & ERP & Non-linear optimization & Keypoint/MVS  & Dense \\
Im~\etal~\cite{Im:ECCV:2016}  & SfM & HP ($\times2$)& Non-linear optimization & Keypoints/Sweeping & Dense \\ 
Guan and Smith~\cite{Guan:TIP:2017}  & SfM & ERP & Linear/Non-linear optimization & Keypoint & Sparse \\ 
Huang~\etal~\cite{Huang:VR:2017}  & SfM & CMP & Non-linear optimization & Keypoints/Block matching & Dense \\ 
Silveira and Jung~\cite{Silveira:VR:2019}  & SfM & ERP & Linear optimization & Keypoints/\tlts{Optical flow} & Dense\\ 

Wang \etal~\cite{Wang:ACCV:2018}  & SfM & CMP & CNN & Learned features & Dense\\ 
Won \etal~\cite{Won:ICRA:2019}  & SfM & HP ($\times4$) & CNN & Learned features & Dense \\
\hline
\end{tabulary}
\end{table}



\subsection{Spherical Light Fields (SLFs)} \label{subsec:lightfields}
 
\tlts{The light field is the representation of  the radiance as a function of position and direction, and it can be modeled as a 4D function in the free space as proposed by Levoy and Hanraran~\cite{Levoy:Hanraran:CCGIT96}. Such representation resembles the concept of epipolar volumes and planes proposed in~\cite{Bolles:etal:IJCV87}, which are produced by several closely related captures of a perspective camera along a line and generate directional structures related to depth information. Although using moving cameras is one possibility for estimating the light field, there are other possibilities such as arrays of cameras or cameras with an array of (micro)lenses~\cite{Levoy:Computer:2006}}.

\begin{figure}
    \centering
    \includegraphics[scale=.7]{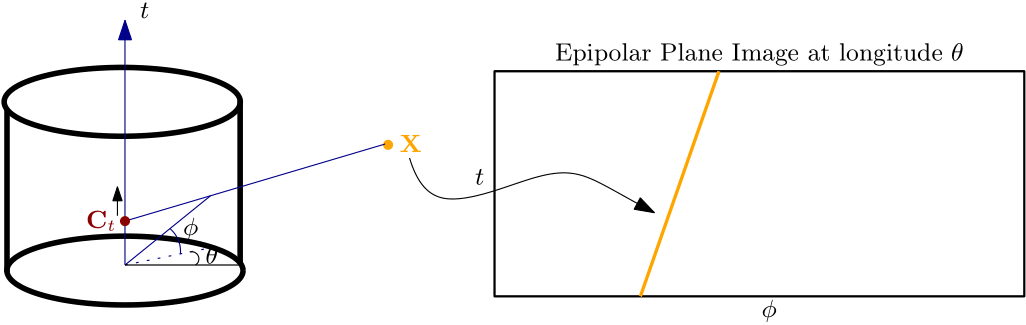}
    \caption{\tlts{Relationship between SLF, Epipolar Plane Image and depth using a CP model (adapted from~\cite{Krolla:BMVC:2014}).}}
    \label{fig:slf:krolla}
\end{figure}

\tlts{For panoramas, earlier approaches for capturing the light field consisted of using a perspective-based light field imaging device coupled with an array of spherical mirrors~\cite{Taguchi:etal:TOG2010} or rotated in a circular motion pattern~\cite{Birklbauer:etal:CGF2014}. However, the term ``spherical light field'' was coined in the seminal work from Krolla \etal \cite{Krolla:BMVC:2014}, which also presents the first attempts in indoor and outdoor scene reconstruction using full-FoV light fields. They use the Mercator projection \cite{Bentsen:MWR:1999} to produce CP images from a set of vertically displaced captures. These captures induce Epipolar Plane Images with oriented lines under static scene and Lambertian surfaces assumptions, from which depth can be estimated. Despite the elegant mathematical formulation,  the use of CP generates strong distortions around the poles, compromising the depth estimate at these regions. Fig.~\ref{fig:slf:krolla} illustrates the trajectory of a 3D point {$\mathbf{X}$} in the Epipolar Plane Image when the camera center {$\mathbf{C}_t$} moves vertically. Note that a linear pattern is expected in the  $\phi \times t$ plane, and its orientation is related to the depth of {$\mathbf{X}$}.} 

Gava \etal \cite{Gava:ICIP:2018} propose  an approach for tackling SLFs capable of estimating consistent, locally smooth depth values from a scene. Sum-of-squared-differences at the subpixel level helps building a cost volume for each pixel position in a reference view (the central one along the translation axis). A variational approach modeled with data and smoothness terms processes the cost volume and extracts the underlying 3D surface from the scene. Their method  extracts sharp depth discontinuities and  tackles outdoor scenes, but might fail in textureless areas and occlusions.

\tlts{Sitzmann and colleagues~\cite{Sitzmann:arxiv2021} present a 
light field 
representation called Light Field Networks. Similarly to~\cite{Krolla:BMVC:2014}, they find an analytical relationship between the directional structures of the Epipolar Plane Images and 
the
scene depth. However, they explore Pl\"{u}cker coordinates to encode the light field instead of a CP representation, which mitigates the limitations around the poles. Despite the promising theoretical results, only simple 3D scenes were explored in~\cite{Sitzmann:arxiv2021}. }

\tlts{Some approaches explore the concept of SLFs but focusing on other applications such as view synthesis. Although they do not explicitly explore the light field for inferring depth, they make use of multiple camera (or multiple capture) setups to obtain depth information using SfM or MVS approaches. For example, Overbeck \etal~\cite{overbeck:tog:2018} acquire SLF datasets with two camera rigs: one of them has 16 GoPro Hero4 cameras arranged in a vertical arc placed on a rotating platform (which prioritizes speed), and the other explores a spinning pivoting platform with two Sony a6500 fisheye mirrorless cameras (which prioritize quality). They extract depth information by building a 3D cost-volume and exploring a fast bilateral filtering approach~\cite{Barron:etal:CVPR2015}, which is used to build a mesh and guide the view interpolation process.  
Bertel and colleagues~\cite{bertel:tog:2020} used off-the-shelf 360\degree cameras for the capturing SLFs (typically less than a hundred). Similarly to~\cite{overbeck:tog:2018}, depth is extracted directly from the captured image set and not the SLF by using a
visual
simultaneous localization and mapping
approach~\cite{Sumikura:etal:ICM2019}}.



\subsection{Multi-view Stereo (MVS)} \label{subsec:mvs}

The core idea behind MVS methods is to explore three or more \emph{calibrated} cameras for estimating 3D information from a scene. The camera poses can be known \textit{a priori} using a fixed, pre-calibrated camera rig, or obtained on-the-fly using information about the scene. 
The latter strategy\tlts{, which demands a more complex camera setup,} will be revised in the next section.

A direct way to obtain a dense depth map is to triangulate 
\tlts{optical flow}-guided correspondences  (recall Section~\ref{subsec:densefeatures}) if the camera poses are known. As an example, Bagnato \etal~\cite{Bagnato:ICIP:2009} explore this idea, but the details of the depth map might be compromised due to the regularizer used to obtain the 
\tlts{optical flow}.

Kim and Hilton present incremental solutions for 3D recovery
using multiple stereo pairs~\cite{Kim:ICCV:2009, Kim:ECCV:2010,Kim:IVMSPW:2013,Kim:IJCV:2013}. %
\tlts{In~\cite{Kim:IJCV:2013} they extend their stereo-based approach described in~\cite{Kim:ICCV:2009}, which was discussed earlier in Section~\ref{sec:stereo}. 
More precisely, a mesh 
is extracted by triangulation from each \tlts{cube-map projected} disparity map, which is registered with the others through an adaption of the iterative closest point (ICP) algorithm \cite{Besl:TPAMI:1992} supplied with surface reliability checking. 
The combined mesh presents a fair 3D structure with attenuated occlusion-caused issues and well-defined textured regions.
Preliminary versions of the work published in~\cite{Kim:IJCV:2013} can be found in \cite{Kim:ICCV:2009, Kim:ECCV:2010}.} 
%
In~\cite{Kim:JCVIU:2015}, they present a light-weight alternative to~\cite{Kim:IJCV:2013}, assuming 
Manhattan worlds (as done for 3D layout recovery).
The authors use the same camera setup as~\cite{Kim:ICCV:2009, Kim:IJCV:2013}, but the images are 
represented in CMP format after multi-view fa\c{c}ade alignment.
Initial disparities are estimated by their PDE-based approach
\cite{Kim:ICCV:2009, Kim:IJCV:2013}, and  planes are fitted and registered together with the help of SURF features. The resulting planes are refined based on reliability, visibility, and occupancy checks, and finally, the final cuboid representations are built. 
A preliminary version of \cite{Kim:JCVIU:2015} can be found in \cite{Kim:IVMSPW:2013}.

More recently, Won \etal~\cite{Won:ICCV:2019} presented an end-to-end 
\tlts{deep learning}
approach for depth map estimation from ultra-wide FoV cameras (four 220$^\circ$ FoV fisheye views). 
A 2D CNN extracts features that are projected onto spheres with varying radii (sphere sweeping), and a 3D encoder-decoder network builds a cost volume and  outputs dense depth panoramas. An extended version~\cite{won:pami:2020} further analyzes the inverse depth probability distributions in uncertain regions and proposes a regularization strategy to improve the depth estimates. The results shown in~\cite{Won:ICCV:2019,won:pami:2020} are compelling, but they do not explore other camera setups and disregard calibration camera noise.

\tlts{As a final comment, we note that a  calibrated spherical rig generates a set of six calibrated ``semi-planar'' views for each spherical image in the rig when the CMP representation is used, which might allow the use of traditional MVS approaches for the individual cube faces. However, the FoV ($90^\circ$) in each face is higher than most pinhole cameras and leads to a certain level of ``radial'' distortion~\cite{lee:cvpr:2019,Caruso:IROS:2015}, which may impact the performance of planar feature detectors~\cite{Morel:JIS:2009}. Also, small indoor scenarios and captures where the camera is too close from a wall may result in cube faces with practically no texture, so that feature matching and further depth recovery tend to fail~\cite{Hartley:Book:2003}. Moreover, spherical captures presenting rotations generate CMP images that are not aligned -- the faces in one capture might relate to a region comprising two or three faces in other captures -- which might highlight distortions. As such, it is advisable to perform image derotation~\cite{bergmann:sibgrapi:2021,Jung:VR:2019} on the sphere before generating the CMP images, or to explore camera setups with small rotations as in~\cite{Kim:ICCV:2009, Kim:ECCV:2010,Kim:IVMSPW:2013,Kim:IJCV:2013}, which assume upright image captures. Finally, per-face depth estimation tends to produce discontinuities along face edges~\cite{Silveira:ICIP:2018,Wang:CVPR:2020,Won:ICRA:2019}, and post-processing is required to obtain the full 3D model of the scene. Despite the stronger distortions present in ERP representations, we believe that they present more potential for exploring global information through deep learning approaches as in \cite{Won:ICCV:2019,won:pami:2020}, particularly if coupled with spherical network adaptations as presented in Section~\ref{subsec:spherenn}. }

\subsection{Structure from Motion (SfM)} \label{subsec:sfm}

SfM aims to estimate camera poses and the 3D structure of the scene. If camera poses are known, 3D reconstruction can be achieved using MVS, as described in Section~\ref{subsec:mvs}. SfM is often called visual simultaneous localization and mapping (V-SLAM) in the robotics jargon, with the additional constraint of online processing~\cite{Saputra:CSUR:2018}. We add that V-SLAM  usually explores temporal information (robot-mounted camera exploring an environment) and focus mostly on the camera pose (trajectory) part of the problem, whereas our goal in this survey is 3D geometry recovery.

The problem of reconstructing large 3D outdoor scenes using a single moving omnidirectional camera was tackled in~\cite{Torii:ICCV:2009,Micusik:CVPR:2009,Caruso:IROS:2015}. 
Torii \etal~\cite{Torii:ICCV:2009} and Micus\' {i}k and Koseck\`{a}~\cite{Micusik:CVPR:2009}
use planar SURF descriptors to obtain correspondences across frames, but trajectory estimation exploring loop closure (as in other approaches for V-SLAM) was the focus in~\cite{Torii:ICCV:2009},  whereas depth estimation (with the help of superpixels, under a piece-wise planar assumption)  was the main goal of~\cite{Micusik:CVPR:2009}. Caruso \etal~\cite{Caruso:IROS:2015}  explore a robust photometric error to align the frames and explore keyframe matching to refine the alignment. A semi-dense depth map is then obtained by direct search over epipolar curves. 
The main goal of~\cite{Torii:ICCV:2009,Caruso:IROS:2015} is motion/trajectory estimation, and only a visual analysis is provided for the obtained 3D data; the method from Micus\' {i}k and Koseck\`{a}~\cite{Micusik:CVPR:2009} provides a denser reconstruction, but since it is based on planar primitives, it 
is not able to capture finer scene details. Huang and colleagues~\cite{Huang:VR:2017} also explore a moving camera, but targeting indoor scenes and view synthesis applications for VR. They track features using Kanade-Lucas-Tomasi (KLT) ~\cite{Lucas:IJCAI:1981} in regularly sampled
CMP keyframes, and explore a bundle-adjustment (BA) procedure for incrementally refining both the camera pose and sparse 3D estimates. 
They further adopt   
\tlts{an}
algorithm \tlts{based on normalized-cross-correlation}
\cite{Shen:TIP:2013} that generates a dense depth map via interpolation
of the sparse 3D points of each keyframe. Finally, 
the method merges all the depth maps and creates a mesh representation suited for 3D rendering.
Despite the promising view synthesis results shown in~\cite{Huang:VR:2017}, large camera displacements across adjacent frames might compromise KLT tracking. \tlts{It is also worth mentioning the OpenVSLAM framework~\cite{Sumikura:etal:ICM2019}, which can handle several camera types including panoramas. However, since planar ORB features are used for matching, the tool does not fully explore the potential of spherical imagery.}

The works of Barazzeti \etal~\cite{barazzetti:isprs:2017} and Fangi \etal~\cite{Fangi:ISPRS:2018}  targeted larger scenes in the context of cultural heritage, but focused on the evaluation the 3D error of spherical SfM approaches using different cameras. A consumer-grade Samsung Gear 360 camera was analyzed in~\cite{barazzetti:isprs:2017}, whereas a high-resolution Panono $360^\circ$ camera 
was evaluated in~\cite{Fangi:ISPRS:2018}. Although not many technical details about the underlying SfM approaches were provided in either work (they use a comercial solution), they both reported low reconstruction errors, highlighting that non-uniform illumination was a challenge for capturing indoor panoramas.  


%

Other works focus on the estimation of denser 3D maps. Bagnato \etal~\cite{Bagnato:JMIV:2011} explore their spherical 
\tlts{optical flow}
approach~\cite{Bagnato:ICIP:2009} 
to  obtain dense correspondences, which is integrated 
with an ego-motion estimator in an alternating fashion. Unlike~\cite{Bagnato:ICIP:2009}, it does not require the camera trajectories to be known, but it  suffers from the same drawbacks of~\cite{Bagnato:ICIP:2009}: problems with large displacements (due to linearization of the flow) and loss of details for small objects. Schonbein and Geiger~\cite{Schonbein:IROS:2014} also target at dense 3D reconstruction assuming augmented Manhattan worlds. The authors
explore a moving rig of four calibrated catadioptric cameras so that MVS can be used for spatial and SfM for temporal alignments. They match planar descriptors (a combination of FAST
and BRIEF) for motion estimation, and use it to rectify the four omnidirectional stereo pairs. Depth is recovered through a spherical semi-global stereo matching algorithm,  in which planar hypotheses are  obtained by a Hough-like voting scheme applied to a superpixel representation of the (inverse) depth map and  combined in a discrete energy minimization problem.
Their approach presents fair results, but requires a pre-calibrated rig and is computationally expensive.
Pagani and Stricker~\cite{Pagani:ICCV:2011} propose a pipeline with two phases: (i) an SfM step for recovering the pose and sparse 3D representation of the scene; 
and (ii) an MVS step for estimating denser representations of the scene.  More precisely, they use ASIFT-like keypoints and an 8-PA RANSAC framework, followed by non-linear optimization, for pose recovery. A sparse point cloud is obtained by iteratively 
solving the perspective-n-point (PnP) problem via direct linear transform (DLT)~\cite{Abdel-Aziz:PERS:1971}
and applying a non-linear camera pose refinement. 
Finally, a densification process explores
the popular Patch-based Multiple View Stereo (PMVS) algorithm~\cite{Furukawa:CVPR:2007}. A similar work~\cite{Pagani:VAST:2011} also explored PMVS, but using a more systematic approach for finding anchors to guide PMVS.
Both approaches present compelling visual results, but no comparison with other techniques is
provided.

Specific properties of panorama capture devices can also be used for SfM. Im \etal \cite{Im:ECCV:2016} 
explore weakly overlapping fisheye sensors looking at opposite directions from short, narrow-baseline monocular 360\degree videos. Due to the small motion assumption, the authors 
directly minimize the reprojection error on the unit sphere 
using Harris corner features that are tracked by the
KLT algorithm~\cite{Lucas:IJCAI:1981}. 
They also propose a sphere sweeping algorithm (extending the main idea behind traditional plane sweeping) that generates a dense depth map from the video. Besides being dependent on a dual wide-FoV fisheye sensor and short slow motion videos, their method tends to produce noisy 3D reconstructions due to the lack of a regularizer.

\begin{figure}
    \centering
    \includegraphics[scale=.55]{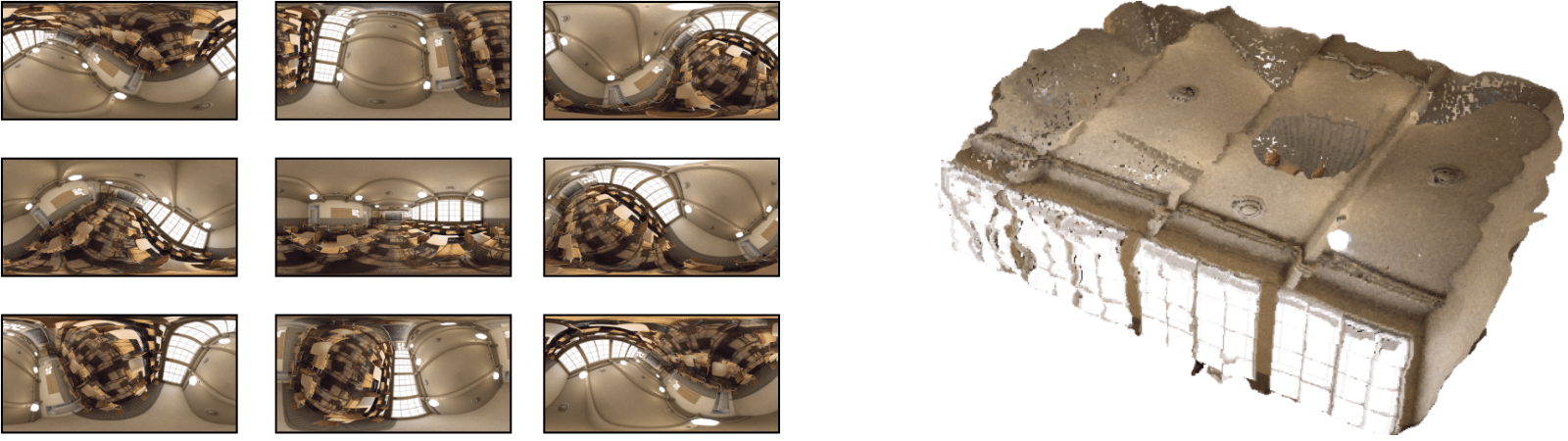}
    \caption{Multiple views and the underlying recovered 3D geometry using the SfM approach from~\cite{Silveira:VR:2019}. The~``Classroom model'' is available
    at \url{https://www.blender.org}.}
    \label{fig:mvssilveira2019}
\end{figure}

Guan and Smith~\cite{Guan:TIP:2017} address the 360\degree
video stabilization problem via SfM. They match sparse S-SIFT features across frames, and use the 8-PA supplied with RANSAC
to estimate the initial two-view camera pose. Additional camera poses are estimated by solving the PnP (renamed  to `Spherical-n-Point'', or SnP) problem via DLT.
The main contribution of~\cite{Guan:TIP:2017} is the introduction of a BA framework that is optimal for the assumed noise model for keypoint matching errors on the unit sphere, based on the von Mises-Fisher distribution \cite{Wood:CS-SC:1994}. The low pose estimation errors achieved by their method may indicate good results in 3D reconstruction, but dense  geometry recovery is not addressed in their work.
Silveira and Jung~\cite{Silveira:VR:2019} followed a similar approach for pose estimation, but aiming dense reconstructions. 
They used sparse SPHORB features to solve the two-view camera poses (via 8-PA-RANSAC) relating one reference view to all others. The rotation parameters extracted from the Essential matrices are used to align (derotate) all views (similarly to~\cite{Micusik:CVPR:2009}), and DeepFlow was used to obtain dense correspondences between the reference image and the remaining aligned views (circular horizontal padding was used). 
The full translation vectors of all views are obtained using a simplified SnP model, and a weighted multi-view calibrated reconstruction yields a dense depth map, further post-processed with a spherically-modified Domain-Transform 
filter~\cite{Gastal:TOG:2011}. As an illustration, Fig.~\ref{fig:mvssilveira2019} shows a
reconstruction result using~\cite{Silveira:VR:2019} based
on nine input panoramas.

A set of approaches focus on indoor 3D scene reconstruction 
assuming Manhattan worlds~\cite{Cabral:CVPR:2014,
Pintore:PCCGA:2018,Pintore:CAG:2018}. 
Cabral and Furukawa~\cite{Cabral:CVPR:2014}  
claim to use a traditional SfM scheme to find the camera poses (no details are provided) and MVS to obtain 3D point clouds, which are combined with an oversegmented representation~\cite{felzenszwalb:ijcv:2004} of each panorama to obtain an initial labeling as wall, floor or ceiling. 
A 2D floorplan is built using a graph-based approach, and the final 3D model is obtained by extruding the walls up to the ceiling. Pintore and colleagues~\cite{Pintore:PCCGA:2018,Pintore:CAG:2018} also explored superpixel-based classification primitives, but using Simple Linear Iterative Clustering (SLIC)~\cite{achanta:pami:2012} instead.  
While the work in~\cite{Pintore:PCCGA:2018} strongly relies on SfM for obtaining 3D information, the approach in~\cite{Pintore:CAG:2018} focuses on independent single-image layout reconstruction (recall Section~\ref{subsec:layout}), and explores multi-view alignment (using the BA approach for CMP representations presented in~\cite{kangni:ICCV:2007}) mostly for refining the per-image layouts and obtaining the multi-room reconstruction. Note that these three approaches
explore superpixels designed for planar images, which might lead to an undesidered over-segmentation around the poles (due to the non-uniform sampling of panoramas), which 
could
be replaced with superpixel algorithms tailored to the spherical domain such as~\cite{Zhao:TMM:2018, Giraud:ARXIV:2020,silveira:sigpro2021}.



\tlts{In the direction of deep learning},  Wang \etal~\cite{Wang:ACCV:2018} present a self-supervised approach for spherical SfM. 
Their model uses CMP images and treat each cube face as
planar (with a special treatment at the edges of the cube). 
Two networks (one for depth and the other for camera motion estimation) work on each individual cube face, and then are combined to incorporate photometric and pose constraints  
exploring relative (known) pose differences among the faces.
%
Won \etal~\cite{Won:ICRA:2019} also tackle both the pose and 3D scene reconstruction problems, but by considering
four fisheye cameras spaced by a wide and fixed baseline. 
Intrinsic camera calibration is a requirement for fisheye imagery (performed offline), and a BA algorithm handles the
extrinsic calibration. The authors introduce a CNN model that takes warped fisheye images and outputs a cost volume that is  filtered using a sweeping method and further aggregated by semi-global matching. Although the authors argue that the method works with full spheres, they disregard the image poles
in their evaluation. In a follow-up study~\cite{won:arxiv:2020}, they modify the CNN and add a loop-closure detection module, 
improving both pose and 3D reconstruction estimates.


\section{Benchmarking: Datasets and Evaluation Metrics} \label{sec:benchmark}

All approaches surveyed in this paper 
address the problem of inferring 3D information from a scene. 
However, the representation can be encoded as layout corners, sparse point clouds, dense point clouds,  set of planar patches, or even meshes.
Moreover, some methods present metrical depth estimates while others use a fictional distance (particularly for single-panorama methods).
As such, comparing this diversified set of approaches is not an easy task.
Next, \tlts{in Sections~\ref{subsec:datasets} and~\ref{subsec:metrics},} we compile and briefly describe relevant datasets and 
figures of merit 
used to assess all these kinds of techniques.
\tlts{Finally, we show selected state-of-the-art results considering common datasets and figures of merit for the tasks of single-image layout and depth estimation in Section~\ref{subsec:sota}.}

\subsection{Datasets} \label{subsec:datasets}

\begin{table}[]
\color{black}
\setlength{\tabcolsep}{0.25em} 
\footnotesize
\centering
\caption{\tlts{Publicly available 360\degree imagery datasets.}}
\label{tab:datasets}
\begin{tabulary}{\linewidth}{m{2.8cm}x{5cm}x{1.cm}x{2cm}x{2cm}}
\hline
\textbf{Reference} & \textbf{Dataset URL} & \textbf{Number of Images} & \textbf{Data Source} & \textbf{Annotation Type} \\
\hline
Zhang~\etal~\cite{Zhang:ECCV:2014} & \url{http://panocontext.cs.princeton.edu} & 700 & Subset of \cite{Xiao:CVPR:2012} & Layout \\
Fernandez-Labrador~\etal \cite{Fernandez-Labrador:LRA:2020} & 
\url{http://webdiis.unizar.es/~jmfacil/cfl/sun360} & 518 & Subset of \cite{Xiao:CVPR:2012} & Layout \\
Zou~\etal~\cite{Zou:CVPR:2018} & \url{https://github.com/zouchuhang/LayoutNet} & 571 & Subset of~\cite{Armeni:arXiv:2017} & Layout\\
Yang~\etal  \cite{Yang:CVPR:2019} & \url{https://github.com/SunDaDenny/DuLa-Net} & 2,573 & Subset of \cite{Xiao:CVPR:2012}/Original & Layout\\
{Pintore \etal~\cite{Pintore:ECCV:2020}} & \url{https://github.com/crs4/AtlantaNet} & {122} & Subset of \cite{Zheng:arXiv:2019,Chang:3DV:2017}/Original  & Layout \\

Armeni~\etal \cite{Armeni:arXiv:2017} & \url{http://3dsemantics.stanford.edu} & 1,413 & Original & Depth, normals, semantics, and camera poses \\
Chang~\etal~\cite{Chang:3DV:2017} & \url{https://niessner.github.io/Matterport/} & 10,800 & Original & Depth, semantics, and camera poses \\
Zioulis~\etal \cite{Zioulis:ECCV:2018} & \url{https://vcl3d.github.io/3D60/}& 24,933 & Re-renderings from~\cite{Chang:3DV:2017,Armeni:arXiv:2017, Song:CVPR:2016} & Depth and normals \\

Wang~\etal \cite{Wang:ACCV:2018} & \url{https://fuenwang.ml/project/360-depth/} & 25,000 & Re-renderings from~\cite{Song:CVPR:2016} & Depth and camera poses\\

Tchapmi and Huber~\cite{Tchapmi2019} & \url{https://sumochallenge.org/} & 58,631 & Re-renderings from \cite{Song:CVPR:2016} & Depth, semantics, and camera poses\\
Zheng~\etal \cite{Zheng:arXiv:2019}& \url{https://structured3d-dataset.org/} & 196,515 & Original & Layout, depth, normals, albedo, instance, semantics, and camera poses\\
Lai~\etal \cite{Lai:VR:2019} & \url{https://github.com/pokonglai/ods-net} & 50,000 & Re-renderings from~\cite{Armeni:arXiv:2017}   &Depth and normals \\
Wang~\etal~\cite{Wang:arXiv:2019} & \url{https://albert100121.github.io/360SD-Net-Project-Page/} & 3,577 &Re-renderings from~\cite{Armeni:arXiv:2017} & Depth and disparity\\
Won~\textit{et al.}~\cite{won:pami:2020} & \url{http://cvlab.hanyang.ac.kr/project/omnistereo/} & 2,560 & Original (Blender Scenes) & Depth\\
Jin~\etal \cite{jin:cvpr:2020} & \url{https://svip-lab.github.io/dataset/indoor} & 1,755 & Original & Layout and depth\\
\hline
\end{tabulary}
\end{table}

An increasing number of datasets contain annotated panoramas for 3D reconstruction applications. 
\tlts{A list of 
representative
datasets,  categorized \wrt the information they provide, is 
shown in Table~\ref{tab:datasets}.} 
%
%
The next subsections briefly describe these datasets, indicating if they are mostly suited for the layout or depth estimation problems.

\subsubsection{Layout Estimation}

\setlist[description]{labelindent=0pt,leftmargin=0.30cm}
\begin{description}
\item[Dataset from~\cite{Zhang:ECCV:2014} --]
Zhang and colleagues~\cite{Zhang:ECCV:2014} select 700 ERP images of scenes including bedrooms and living rooms from the SUN360 database~\cite{Xiao:CVPR:2012}, which has about 57,000 different full-FoV color images taken in both indoor and outdoor scenarios. 
The dataset from~\cite{Zhang:ECCV:2014} provides the 3D bounding volume annotations for the scene layout and the objects therein. 
%
The layouts are 
modeled as simple cuboid shapes.
\tlts{Two halves compose the train and test data.}

\item [Dataset from~\cite{Fernandez-Labrador:LRA:2020} --]
Fernandez-Labrador~\etal \cite{Fernandez-Labrador:LRA:2020}
provide layout annotations to a
subset of the SUN360~\cite{Xiao:CVPR:2012} and Stanford 2D-3D~\cite{Armeni:arXiv:2017} datasets, yielding  518 
$1024\times 512$ ERP images and annotated layout borders and corners, provided as probability maps.
\tlts{Annotated layouts are assumed to fit the Manhattan world assumption.}
Train and test splits  are preset.

\item[Dataset from~\cite{Zou:CVPR:2018} --]
Zou~\etal~\cite{Zou:CVPR:2018} provides cuboid layout annotations for 571 images taken from the Stanford 2D-3D~\cite{Armeni:arXiv:2017} dataset. 
%
\tlts{Layout annotations are given as 2D layout corners in ERP format.}
%
Train, validation, and test
splits are provided.

\item[Realtor360 dataset --]
Yang and colleagues~\cite{Yang:CVPR:2019} provide layout annotations (4 to 12 corners) for 593 panoramas from a subset of the SUN360 database~\cite{Xiao:CVPR:2012}. 
The authors also annotate 1,980 360\degree images from a real estate database. 
\tlts{The annotated data models Manhattan world scenarios.}
Train and test splits are provided, and the dataset might be available upon request.
{\color{black}
\item[AtlantaLayout dataset --] 
Pintore \etal~\cite{Pintore:ECCV:2020} annotate
122 ERP images of variable size 
(\eg, 512$\times$1024, 2048$\times$1024, 5376$\times$2688) 
from different sources, such as  Matterport3D~\cite{Chang:3DV:2017} and Structured3D \cite{Zheng:arXiv:2019}.
The annotations are provided as 2D ERP coordinates corresponding to 3D corners of  Atlanta world scenarios. Train, validation and test splits are provided.}
\end{description}


\subsubsection{Depth/3D Estimation}

\begin{description}
\item [SUNCG --]
Song~\etal~\cite{Song:CVPR:2016} provide
$45,622$ computer-generated 3D scenes with semantic information suitable for rendering. 
%
%
The SUNCG dataset does not provide panoramic images, but it is used in related literature. 
A method for simulating real depth captures is also made available.

\item[Stanford 2D-3D --]
Armeni~\etal~\cite{Armeni:arXiv:2017} 
made available more than $1,413$ full-FoV indoor real captures from six large-scale areas, including 
offices, classrooms, and corridors. 
The dataset offers registered depth, normals, and semantic maps.
Color information is, however, missing in the poles. The camera poses, point clouds and 3D meshes are also available. Train and test splits at the area level are preset.

\item[Matterport3D --]
Chang~\etal~\cite{Chang:3DV:2017} 
provide 
$10,800$ panoramic views from
90 real building-scale scenarios. 
The images are in CMP format and cover 3.75 steradians (poles are missing).
%
The dataset also provides depth, camera poses and semantic segmentation information.
%
%
%
Train, validation, and test splits are made avaliable at the scenario level.  Manhattan layout annotations were recently added to a subset of Matterport3D by~\cite{zou:ijcv:2021}.

\item[3D60 --] 
%
Zioulis \etal
\cite{Zioulis:ECCV:2018}
made available
24,933 $512\times 256$ ERP-formatted renders from other datasets: Matterport3D~\cite{Chang:3DV:2017}, Stanford 2D-3D~\cite{Armeni:arXiv:2017}, and SunCG~\cite{Song:CVPR:2016}.
It contains aligned color, depth, and normal maps in a trinocular camera setup. 
%
%
Train and test splits are preset.

\tlts{\item[PanoSUNCG --]
Wang \etal \cite{Wang:ACCV:2018} provide about 25,000 images captured from 103 different scenarios from SUNCG~\cite{Song:CVPR:2016} in five camera paths. The dataset provides the color, depth and rendering trajectory. The authors suggest using data from 80 scenes for training and the remaining 23 for testing.}

\item[Scene Understanding and Modeling (SUMO) --]
Tchapmi and Huber~\cite{Tchapmi2019} made available 
$58,631$ computer-generated indoor CMP images 
from the SUNCG dataset~\cite{Song:CVPR:2016}. 
Each cube face has $1024\times 1024$ pixels. 
Camera poses, depth, and
2D/3D semantic information are also provided.

\item[Structured3D --]
Zheng~\etal \cite{Zheng:arXiv:2019} 
provide
$196,515$ frames from $3,500$ synthetic, photo-realistic house designs. 
A scene has $1024\times 512$ 
ERP images at three different light conditions and camera poses.
%
Instance annotation, semantic, albedo, normal, layout, and depth maps 
are available.
%
The dataset also enables
object detection, scene understanding,
and view synthesis.

\item[Dataset from~\cite{Lai:VR:2019} --]
Lai and colleagues~\cite{Lai:VR:2019} make available a re-rendered subset of the Stanford 2D-3D~\cite{Armeni:arXiv:2017} dataset along with ground truth for depth and normal maps. 
Namely, the authors provide about $50,000$ $256\times128$ stereoscopic ERP images separated by a
horizontal baseline of  $6.5$cm. 
Depth values are given in meters. 
Train and test splits are
provided.

\item[Dataset from~\cite{Wang:arXiv:2019} --]
Wang~\etal~\cite{Wang:arXiv:2019}
make available a re-rendered subset of
the Stanford 2D-3D~\cite{Armeni:arXiv:2017} and Matterport3D~\cite{Chang:3DV:2017} datasets.
%
It contains $3,577$ $1024\times 512$ stereoscopic ERP image pairs
separated by a vertical baseline of $20$cm.
Depth and disparity maps are also made available.
%
Train, validation, and test splits are provided.

\item[Datasets from~\cite{won:pami:2020} --]
Won~\textit{et al.}~\cite{won:pami:2020} 
provide three wide-FoV synthetic video-sequences rendered with Blender~\cite{Blender:misc:2020}.
There are three scenarios: one outdoors and one indoors.
A total of $2,560$ frames are made available. 
Depth information is provided.
Train and test splits at video-sequence level are preset.

\item[Shanghaitech-Kujiale dataset --]
Jin~\etal \cite{jin:cvpr:2020} 
provide a 
synthetic dataset containing 
two $1024\times 512$ ERP 
images
for 1,775 rooms -- with and without furniture.
The dataset also has depth maps and room layout (corners and edges) annotations. 
Train and test splits are provided.
\end{description}

\tlts{In general, datasets with preset train, validation, and test splits allow a fair and reproducible comparison among different techniques. For the layout case, simpler annotations (such as cuboids) can be used for (pre-)training and assessing more general approaches (such as those dealing with Manhattan or Atlanta worlds~\cite{Zou:CVPR:2018,Pintore:ECCV:2020}), but without exploring their full potential. On the other hand, methods that assume more constrained layouts cannot be trained or tested using datasets with generic annotations without a proper conversion.  Note that datasets proposed exclusively for layout estimation still have a limited number of images. 
\tlts{Many datasets are used for monocular depth benchmarking, such as the Matterport3D, Stanford 2D-3D, and the 3D60 -- which unites the two former plus re-renderings from the SUNCG.}
\tlts{Only few datasets consider captures from multiple views to be used for stereo- or multi-view-based depth estimation, such as 3D60, PanoSUNCG, and the datasets from \cite{Lai:VR:2019, Wang:arXiv:2019, won:pami:2020}.}
More generic datasets, such as the SUMO, Structured3D, and Shanghaitech-Kujiale datasets have much more data, are applicable to both monocular layout and depth estimation, but contain only synthetic scenarios.}

\subsection{Evaluation Metrics} \label{subsec:metrics}


A plethora of figures of merit can be used to assess the quality of a 3D scene geometry reconstruction method. 
Next, we point out the most popular metrics found in the literature, distinguishing approaches suitable for layout from those for depth/3D geometry estimates.

\subsubsection{Layout Estimation}

Layout information can be represented by corners or edges, which can be considered as (possibly binarized) probability maps lying on the 2D or 3D space.
Classical metrics like precision, recall, F1-score, and accuracy can be applied to layout estimates -- represented either as corners or edges -- when formatted in an ERP image~\cite{Fukano:ICPR:2016, Fernandez-Labrador:LRA:2020, Fernandez-Labrador:arXiv:2018}.
Additionally, the intersection over union (IoU) can be applied to both 
ERP 
images (2D IoU)~\cite{Pintore:ECCV:2020, Yang:CVPR:2019, Fernandez-Labrador:LRA:2020, sun:arxiv:2020} or 
data
(back)projected 
to the 3D space (3D IoU)~\cite{Pintore:ECCV:2020, Zheng:arXiv:2019, Zou:CVPR:2018, Yang:CVPR:2019, Fernandez-Labrador:LRA:2020, Sun:CVPR:2019, Fernandez-Labrador:arXiv:2018, sun:arxiv:2020}.
Angular corner error and object/planar surfaces length/area disagreement are also used for assessing scaled layout estimates~\cite{Xu:WACV:2017, Pintore:WACV:2016}.

More specific  metrics are also found in the literature.
%
The corner error (CE) computes the normalized L2 distance between the predicted and ground truth corners~\cite{Pintore:ECCV:2020, Zheng:arXiv:2019, Zou:CVPR:2018, Fernandez-Labrador:LRA:2020, Sun:CVPR:2019, Fernandez-Labrador:arXiv:2018}; and the pixel error (PE) -- or pixel classification error (PCE) -- measures the pixel-wise error between predicted and ground truth surface (plane) classes of the cuboid representation~\cite{Pintore:ECCV:2020, Yang:VRCAI:2014, Zheng:arXiv:2019, Zou:CVPR:2018, Fernandez-Labrador:LRA:2020, Sun:CVPR:2019, Fernandez-Labrador:arXiv:2018}.
%
The equally oriented pixel ratio (EOP)~\cite{Fernandez-Labrador:LRA:2018, Jia:ICRA:2015}
computes the percentage of pixels in a predicted label map representing the walls/floor/ceiling orientations that match the ground truth (a simplified normal map-like with three classes only). 


\subsubsection{Depth/3D Estimation}
\label{sec:evaluation:depth}


Most methods evaluate
depth or 3D estimates after scale correction/alignment (\eg, via mean~\cite{Yang:CVPR:2018} or median normalization~\cite{Zioulis:ECCV:2018, Eder:3DV:2019}) \wrt the ground truth annotations. 
In the monocular depth estimation case, 
 absolute depth values are difficult to obtain. Hence, many authors follow the guidelines proposed by Eigen \etal~\cite{Eigen:NIPS:2014, Eigen:ICCV:2015}, 
 introduced  for perspective cameras, which include the accuracy (Acc / $\delta$) at a preset threshold (acceptable error)~\cite{Lai:VR:2019, Kim:JCVIU:2015, Im:ECCV:2016, Zioulis:ECCV:2018, roxas:iral:2020, sun:arxiv:2020, Payen:ECCV:2018, Eder:3DV:2019, Lee:TPAMI:2020, Wang:CVPR:2020, Won:ICRA:2019, Won:ICCV:2019}, where $\delta$ is a suitable error metric;
the mean absolute relative error (Abs Rel / Rel) -- or mean relative error (MRE) --~\cite{sun:arxiv:2020, Wang:CVPR:2020, Lai:VR:2019, Zioulis:ECCV:2018, Payen:ECCV:2018, Eder:3DV:2019, Tateno:ECCV:2018, Lee:TPAMI:2020, Silveira:VR:2019};
the mean square relative error (Sq Rel)~\cite{Zioulis:ECCV:2018, Payen:ECCV:2018, Eder:3DV:2019, Lee:TPAMI:2020};
the linear root mean square error (RMS / RMSE)~\cite{Wang:arXiv:2019, Zioulis:3DV:2019, Kim:JCVIU:2015, Lai:VR:2019, Kim:IJCV:2013, Won:ICCV:2019, Zioulis:ECCV:2018, sun:arxiv:2020, Payen:ECCV:2018, Tateno:ECCV:2018, Wang:CVPR:2020, Won:ICRA:2019}, 
the logarithmic root mean square error (RMSLE / RMSLog)~\cite{Zioulis:3DV:2019, Zioulis:ECCV:2018, sun:arxiv:2020, Payen:ECCV:2018, Eder:3DV:2019, Tateno:ECCV:2018, Wang:CVPR:2020},
and the log, scale-invariant mean square error (SIMSE / RMSE log, scale-invariant MSE)~\cite{Silveira:ICIP:2018, Lee:TPAMI:2020}.

Other common metrics include 
the L2~\cite{Yang:CVPR:2016} and 
cosine distances~\cite{Yang:CVPR:2016, Yang:CVPR:2018},
the mean square error (MSE)~\cite{Arican:AVSS:2007}, and the mean absolute error (MAE)~\cite{Krolla:BMVC:2014,Wang:arXiv:2019, Kim:JCVIU:2015, Kim:IJCV:2013, Gava:ICIP:2018, Won:ICCV:2019, Kim:ICCV:2009, roxas:iral:2020, sun:arxiv:2020, Wang:CVPR:2020, Won:ICRA:2019} between estimates and ground truth annotations. Similarly to the accuracy, the completeness (Com) indicates the percentage of the estimates that cover the ground-truth~\cite{Kim:JCVIU:2015, roxas:iral:2020}. 
Although most metrics 
work on 2D disparity/depth maps, some are considered for assessing 3D representations such as point clouds and meshes~\cite{Kim:JCVIU:2015, Kim:IJCV:2013, Kim:ICCV:2009}. 
It is also worth mentioning that some methods also disregard part of the obtained estimates (missing renders, poles, etc.) when computing these metrics, such as in~\cite{Zioulis:ECCV:2018, Silveira:ICIP:2018, Wang:arXiv:2019, Won:ICCV:2019, won:pami:2020, Silveira:VR:2019}.

\subsection{Selected State-of-the-Art Results} \label{subsec:sota}

\tlts{
This section compiles state-of-the-art results for the problems of layout and depth estimation using a single spherical image. Throughout this manuscript, the reader can observe a growing trend among researchers and the industry for tackling these ill-posed problems via deep learning, whereas the number of recent stereo- and multi-view-based approaches is not very abundant. 
}

\tlts{Table~\ref{tab:sotalayout}, adapted from~\cite{Sun:CVPR:2019} and~\cite{Zioulis:arxiv:2021}, presents state-of-the-art results for layout estimation under the Manhattan world assumption, which is the most common constraint. 
Listed results consider the PanoContext~\cite{Chang:3DV:2017} and Stanford 2D-3D~\cite{Armeni:arXiv:2017} datasets and the CE, PE, and 3D IoU figures of merit. 
To the best of our knowledge, DuLa-Net v2~\cite{zou:ijcv:2021} and Single-Shot Cuboids~\cite{Zioulis:arxiv:2021} attain the best results up to now.}


\begin{table}[]
\color{black}
\setlength{\tabcolsep}{0.25em} 
\centering
\caption{\tlts{State-of-the-art Results in Monocular Layout Estimation.}}
\label{tab:sotalayout}
\begin{tabular}{ccccc}
\hline
\multirow{2}{*}{Dataset} & \multirow{2}{*}{Method} & \multicolumn{3}{c}{Evaluation Metric}                    \\
                         &                         & CE (\%) & PE (\%) & 3D IoU (\%)  \\
\hline
\multirow{11}{*}{PanoContext}
& PanoContext~\cite{Zhang:ECCV:2014}   &  1.60   &  4.55   &  67.23 \\   
& LayoutNet~\cite{Zou:CVPR:2018}   &  1.06   &  3.34   &  74.48 \\      
& DuLa-Net~\cite{Yang:CVPR:2019}   &  N/A   &  N/A   &  77.42 \\      
& Corners for Layout~\cite{Sun:CVPR:2019}  &  0.79   &  2.49   &  78.79 \\    
& HorizonNet~\cite{Sun:CVPR:2019}   &  0.76   &  2.20   &  82.17 \\   
& HorizonNet*~\cite{Sun:CVPR:2019}   &  0.76   &  2.22   &  81.30 \\   
& LayoutNet v2*~\cite{zou:ijcv:2021}   &  0.63   &  1.79   &  \textbf{85.02} \\   
& DuLa-Net v2*~\cite{zou:ijcv:2021}   &  0.82   &  2.54   &  83.41 \\   
& Single-Shot Cuboids~\cite{Zioulis:arxiv:2021}   &  \textbf{0.53}   &  \textbf{1.65}   &  84.89 \\   
& AtlantaNet~\cite{Pintore:ECCV:2020} & N/A & N/A & 78.76\\
& LED$^2$-Net~\cite{wang:arxiv:2021} & N/A & N/A & 82.75\\
\hline
\multirow{9}{*}{Stanford 2D-3D} 
& LayoutNet~\cite{Zou:CVPR:2018}   &  1.04   &  2.70   &  76.33 \\      
& DuLa-Net~\cite{Yang:CVPR:2019}   &  N/A   &  N/A   &  79.36 \\      
& HorizonNet~\cite{Sun:CVPR:2019}   &  0.71   &  2.39   &  79.79 \\   
& HorizonNet*~\cite{Sun:CVPR:2019}   &  0.78   &  2.65   &  80.44 \\   
& LayoutNet v2*~\cite{zou:ijcv:2021}   &  0.71   &  2.04   &  84.17 \\   
& DuLa-Net v2*~\cite{zou:ijcv:2021}   &  0.66   &  2.43   &  \textbf{86.45} \\   
& Single-Shot Cuboids~\cite{Zioulis:arxiv:2021}   &  \textbf{0.56}   &  \textbf{1.83}   &  85.16 \\ 
& AtlantaNet~\cite{Pintore:ECCV:2020} & 0.70 & 2.25 & 82.43\\
& LED$^2$-Net~\cite{wang:arxiv:2021} & N/A & N/A & 83.77\\
\hline
\end{tabular}\\
* Results considering the ResNet-34 architecture~\cite{zou:ijcv:2021}.
\end{table}

\tlts{Table~\ref{tab:sotamonodepth}, adapted from~\cite{sun:arxiv:2020} and~\cite{jiang:arxiv:2021}, shows state-of-the-art results in single-image depth estimation. 
The results are organized by dataset:  Matterport3D~\cite{Chang:3DV:2017}, Stanford 2D-3D~\cite{Armeni:arXiv:2017}, 3D60~\cite{Zioulis:ECCV:2018}, and PanoSUNCG~\cite{Wang:ACCV:2018}. 
Each dataset listed is used for training and testing, using the suggested preset splits.
The estimates are assessed using the MRE, MAE, RMSE, RMSLE and accuracy metrics. 
Up to this point, to the best of our knowledge, UniFuse~\cite{jiang:arxiv:2021} and HoHoNet~\cite{sun:arxiv:2020} achieve the best results in most relevant datasets.}

\begin{table}[]
\color{black}
\setlength{\tabcolsep}{0.25em} 
\centering
\caption{\tlts{State-of-the-art Results in Monocular Depth Estimation.}}
\label{tab:sotamonodepth}
\begin{tabular}{ccccccccc}
\hline
\multirow{2}{*}{Dataset} & \multirow{2}{*}{Method} & \multicolumn{7}{c}{Evaluation Metric}                    \\
                         &                         & MRE & MAE & RMSE & RMSLE & $\delta<1.25$ & $\delta<1.25^2$ & $\delta<1.25^3$ \\
\hline
\multirow{4}{*}{Matterport3D}
&     OmniDepth*~\cite{Zioulis:ECCV:2018}                    &  0.2901   &  0.4838   &  0.7643    &   0.1450         &   0.6830     &    0.8794    & 0.9429\\   
&     BiFuse~\cite{Wang:CVPR:2020}                    &  0.2048   &  0.3470   &  0.6259    &   0.1134         &   0.8452     &  0.9319      & 0.9632\\   
& UniFuse~\cite{jiang:arxiv:2021}   &  \textbf{0.1063}   &  \textbf{0.2814}   &  \textbf{0.4941}    &   \textbf{0.0701}    &   \textbf{0.8897}     &   \textbf{0.9623}   & \textbf{0.9831} \\
&     HoHoNet~\cite{sun:arxiv:2020}        &   0.1488  & 0.2862    &  0.5138    &   0.0871         &  0.8786      &   0.9519     & 0.9771\\  
\hline
\multirow{5}{*}{Stanford 2D-3D} 
&     OmniDepth*~\cite{Zioulis:ECCV:2018}                    &  0.1996   &   0.3743  &   0.6152   &    0.1212        &0.6877        &   0.8891     & 0.9578\\   
&     BiFuse~\cite{Wang:CVPR:2020}                    &  0.1209   &  0.2343   &    0.4142  &     0.0787       &  0.8660      &    0.9580    &0.9860 \\   
& UniFuse~\cite{jiang:arxiv:2021}   &  0.1114  &  0.2082   &  \textbf{0.3691}    &   0.0721    &   0.8711     &   0.9664   & 0.9882 \\   
&     HoHoNet~\cite{sun:arxiv:2020}        &   \textbf{0.1014}  &  \textbf{0.2027}   &   0.3834   & \textbf{0.0668}           &  \textbf{0.9054}      &  0.9693      & \textbf{0.9886}\\   
& Jin \etal~\cite{jin:cvpr:2020}   &  0.1180   &  N/A   &  0.4210    &   N/A    &   0.8510     &   \textbf{0.9720}   & 0.9860 \\ 
\hline
\multirow{3}{*}{3D60} 
&OmniDepth~\cite{Zioulis:ECCV:2018}   &  0.0931   &  0.1706 & 0.3171   &  0.0725    &   0.9092    &   0.9702     & 0.9851 \\ 
&BiFuse~\cite{Wang:CVPR:2020}   &  0.0615   &  0.1143   &  0.2440    &   0.0428    &   0.9699     &   0.9927   & 0.9969 \\ 
&UniFuse~\cite{jiang:arxiv:2021}   &  \textbf{0.0466}   &  \textbf{0.0996}   &  \textbf{0.1968}    &   \textbf{0.0315}    &   \textbf{0.9835}     &   \textbf{0.9965}   & \textbf{0.9987} \\ 
\hline

\multirow{3}{*}{PanoSUNCG} 
&OmniDepth~\cite{Zioulis:ECCV:2018}   &  0.1143   &  0.1624 & 0.3710   &  0.0882    &   0.8705    &   0.9365     & 0.9650 \\ 
&BiFuse~\cite{Wang:CVPR:2020}   &  0.0592   &  0.0789   &  \textbf{0.2596}    &   0.0443&   0.9590     &   0.9838   & 0.9907 \\ 
&UniFuse~\cite{jiang:arxiv:2021}   &  \textbf{0.0485}   &  \textbf{0.0765}   &  0.2802    &   \textbf{0.0416}    &   \textbf{0.9655}     &   \textbf{0.9846}   & \textbf{0.9912} \\ 
\hline
\end{tabular}\\
* OmniDepth is trained with batch normalization for better convergence.
\end{table}

\tlts{We conclude this section by reminding the reader that qualitative results, including some of the methods shown in Tables \ref{tab:sotalayout} and \ref{tab:sotamonodepth}, can be seen in Figs.~\ref{fig:layout_estimation}
and
\ref{fig:real_depth}.} 

\section{Conclusions and Future Trends} \label{sec:conclusions}

Throughout the manuscript, 
we identified approaches 
targeting at  applications based on
room layouts (3D corners or edges of indoor scenes) or explicit depth estimates, which might be either sparse or dense. 
%
The revised techniques face
this fundamental problem relying on a single, two, or multiple 360\degree captures.
The latter camera setup further divides into light fields -- in which the original formulation assumes a very specific camera setup with small baselines in between the cameras and static scenes --, multi-view stereo -- which alleviates such constraints but often require manual camera calibration --, and structure from motion -- that infers camera poses and 3D geometry from the image sequences (including but not limiting video footage).
As for many other 
\tlts{visual computing areas,}
the scientific community is focusing on modeling the 3D scene geometry recovery problem as a learning problem, achieving impressive results.

As 
the reader can see 
throughout this paper, a current trend is to explore a single spherical image for estimating either the layout (of indoor scenes) or explicit depth information.
In most scenarios, this camera setup is certainly the most attractive, since there is no need to pre-calibrate cameras and/or take multiple captures of the environment, and even panoramas captured in the past can be used for layout/depth estimation. 
%
\tlts{3D layout estimation methods}
often assume different geometric constraints for modeling the scene: cuboid shape, Manhattan, and Atlanta worlds, in increasing order of layout flexibility. Only few approaches for layout recovery rely on unconstrained scene models. 
However, one may note
that the vast majority of indoor rooms do follow simpler geometrical rules (such as cuboid and Manhattan), and consequently most datasets used to train layout estimation models contain mostly these types of rooms. 
Hence, more complex datasets are required to further explore the potential of more generic techniques (such as Atlanta or even unconstrained scenarios).

Single-image-based depth estimation techniques provide richer information (typically a dense map with the depth value at each pixel), enabling the discrimination of other objects (such as furniture in the case of indoor scenes or the overall scene geometry in outdoor captures). 
However, as mentioned throughout the manuscript, full (geometrically-consistent) 3D representations of the scenes, such as for immersive navigation in AR/MR/VR applications, require complementary information from 
additional supporting views for addressing occlusions and disocclusions that appear when the (virtual) camera position changes, and also to obtain \emph{truly} geometrical 3D information. 
Therefore, robust multi-view approaches should still be the focus of attention of several research groups worldwide.  
%
Hybrid solutions may
combine multiple single-view depth estimates from the same scene in a multi-view scene completion framework.

The increasing popularity of learning-based solutions raises the need for large annotated datasets (or 3D synthetic data that can be rendered using virtual spherical cameras). In fact, the number of publicly available datasets has been increasing in the past years (see Table~\ref{tab:datasets} for a summary). However, they still contain a limited variability (e.g. mostly indoor environments or indoor scenes with a certain type of room geometry).
Fig.
\ref{fig:real_depth} 
illustrates
the well-known data-dependence issue, but in the context of depth estimation\tlts{. Note that} 
results for \tlts{``unusual'' indoor and} outdoor scenes are considerably worse than those for indoor scenes. 
\tlts{In fact, only two revised works mentioned the possibility of estimating depth from a single image taken in outdoor scenarios.}
A possible trend to mitigate the lack of annotated data is to rely on knowledge distillation, where inter-domain transfer learning techniques can be applied for taking advantage of consolidated perspective datasets in problems with scarce spherical data. 

A related issue concerns the definition of standardized figures of merit for quantitative assessment, which would allow a fair comparison 
among different approaches. 
There is still no consensus in the literature about how to establish a benchmark evaluation setup that encompasses all competing approaches, but the release of publicly
available code by the authors -- which fortunately has been an increasingly popular procedure -- is a good sign. 
In particular, we believe that both the problems of dataset availability and definition of standardized quantitative assessment metrics is even more challenging for the stereo- and multi-view-based scenarios, since they required more that one capture for 
the scenarios, and the 
ground-truth annotations must refer to a reference capture. 
However, the existence of standardized datasets and benchmarks for stereo-related problems using perspective images such as Middlebury~\cite{Scharstein:SMBV:2001} and KITTI~\cite{menze2015object} can inspire the creation of spherical counterparts to deal with panoramas.

\tlts{One aspect that is transversal to all topics tackled in this paper is the chosen representation for the spherical image. On the one hand, the CMP representation provides six planar views of the scene that resemble the outputs of perspective cameras, which is a reasonable choice if we want to explore the vast arsenal of tools that is already available for perspective images. We believe that such representation is suitable for applications such as pose estimation in stereo or multi-view setups since methods for keypoint detection, description, and matching devised (or trained) for perspective images could be used with some adaptations -- Section~\ref{subsec:mvs} provides a few insights in the context of MVS. Since CMP faces still present wider FoVs than a typical pinhole camera, an alternative to alleviate the distortions in the spherical-to-planar mappings can be the use of local tangent plane representations, such as the approach presented by Eder and colleagues~\cite{Eder:CVPR:2020}. In fact, the adaptation of the planar SIFT descriptor using local tangent planes was shown in~\cite{Eder:CVPR:2020} with promising results. On the other hand, we believe that the ERP is a more suitable choice for methods that explore deep neural networks, which is usually the case for single-image layout or depth estimation. In these applications, the receptive field of convolutional layers is implicitly
enlarged with the pooling layers, allowing the network to learn progressively more global information. Since the ERP provides all the information in a single image, it allows the exploration of all image content in the network. In fact, recent methods that have shown state-of-the-art results such as~\cite{Sun:CVPR:2019,sun:arxiv:2020} propose non-local propagation of information over the horizontal direction of the ERP image using recurrent layers.  Performing analogous data propagation schemes in CMP representations require adequate stitching of the cube faces, and our literature review shows that single-image layout or depth estimation methods use mostly CMP in conjunction with ERP~\cite{Wang:CVPR:2020,jiang:arxiv:2021}.
} 

We could also perceive an increasing interest in developing deep neural networks that take into account the inherent deformations of spherical images regardless the representation, as shown in Section~\ref{subsec:spherenn}. However, these models are still considerably slower  than traditional planar deep neural networks, and the use of spherically-adapted networks is still not a consensus at least in layout/depth estimation applications.
Not only end-to-end approaches for 3D geometry recovery are expected to emerge, but also relevant geometrically correct learning-based tools for dealing with sparse and/or dense matching, plane-aware image oversegmentation, pose estimation, etc., should be developed in the near future.

\tlts{Last but not least, we understand that there are still some challenges to be overcome in terms of technical development and applications. Single-panorama layout estimation of a room with simplified geometry presents quite mature solutions and may allow virtual tours in real estate applications, but recovering more complex geometry and fined 3D structures (e.g., furniture) is still ongoing research. Contextual information may preclude several applications that depend on learning-based methods, such as for single-image depth inference. Allowing the estimation of a wide range of depth values for outdoor scenarios, for instance, may impact the parametrization of the networks and demand unavailable data.
Estimating depth in a single frame fashion from video footage may generate flickering due to the lack of temporal coherence. 
Single and multi-view approaches can be coupled to allow full 6-DoF immersive navigation.
Applications such as robot navigation and infrastructure inspection need to build a scene map (mapping) and track the camera motion (localization) jointly, also requiring tighter runtime requirements. 
Most revised methods focus mainly on either mapping or localization, depending on the main application. Constructing accurate world representations in an SfM/VSLAM pipeline often recalls non-linear optimization schemes that are unfeasible for dense estimates, being also more demanding computationally.}


%


\section*{Acknowledgments}
We acknowledge the financial support from the Fundação de Amparo à Pesquisa do Estado do Rio Grande do Sul (FAPERGS - Brazil), the Conselho Nacional de Desenvolvimento Científico e Tecnológico (CNPq - Brazil), and the Coordenação de Aperfeiçoamento de Pessoal de Nível Superior (CAPES - Brazil).




\bibliographystyle{unsrt}  
\bibliography{main}

\end{document}